\documentclass[twoside,11pt]{article}

\usepackage{jmlr2e}

\usepackage[subpreambles=false,mode=buildnew]{standalone}

\usepackage{import}

\usepackage[T1]{fontenc}
\usepackage[utf8]{inputenc}

\usepackage[english]{babel}

\usepackage{soul}

\usepackage{mathtools}

\usepackage{mathrsfs}
\usepackage{amsfonts}

\usepackage{amsmath}

\usepackage{subfig}

\usepackage{hhline}

\usepackage{array}

\usepackage{booktabs}

\usepackage[toc,page]{appendix}

\usepackage{pgfplots,tikz-3dplot}

\usepackage{colortbl}

\usepackage[ruled,vlined]{algorithm2e}

\DeclarePairedDelimiter\abs{\lvert}{\rvert}
\DeclarePairedDelimiter\norm{\lVert}{\rVert}
\DeclarePairedDelimiter\dprod{\langle}{\rangle}

\DeclareMathOperator*{\argmin}{arg\,min}

\makeatletter
\let\oldabs\abs
\def\abs{\@ifstar{\oldabs}{\oldabs*}}
\let\oldnorm\norm
\def\norm{\@ifstar{\oldnorm}{\oldnorm*}}
\let\olddprod\dprod
\def\dprod{\@ifstar{\olddprod}{\olddprod*}}
\makeatother

\newcolumntype{C}[1]{>{\centering\arraybackslash}m{#1}}

\usetikzlibrary{decorations.text,calc,arrows.meta, backgrounds, intersections}
\tdplotsetmaincoords{70}{90}

\tikzset{image/.style={above right, inner sep=0pt, outer sep=0pt}}

\pgfplotsset{compat=1.15}
\usepgfplotslibrary{groupplots}

\pgfplotsset{select coords between index/.style 2 args={
                        x filter/.code={
                                        \ifnum\coordindex<#1\def\pgfmathresult{}\fi
                                        \ifnum\coordindex>#2\def\pgfmathresult{}\fi
                                }
                }}

\makeatletter
\def\relativepath{\import@path}
\makeatother

\usepackage[outdir=./epstopdf/]{epstopdf}

\usepackage{lastpage}
\jmlrheading{23}{2022}{1-\pageref{LastPage}}{12/20; Revised
7/22}{10/22}{20-1405}{Bernardo Fichera, Aude Billard}

\ShortHeadings{\title{Linearization and Identification of Multiple-Attractor DS}}{Fichera, Billard}
\firstpageno{1}

\begin{document}

\title{Linearization and Identification of Multiple-Attractor Dynamical Systems through Laplacian Eigenmaps}

\author{\name Bernardo Fichera \email bernardo.fichera@epfl.ch \\
    \addr Learning Algorithms and Systems Laboratory (LASA)\\
    École polytechnique fédérale de Lausanne\\
    Lausanne, Switzerland
    \AND
    \name Aude Billard \email aude.billard@epfl.ch \\
    \addr Learning Algorithms and Systems Laboratory (LASA)\\
    École polytechnique fédérale de Lausanne\\
    Lausanne, Switzerland
}

\editor{Corinna Cortes}

\maketitle

\begin{abstract}
    Dynamical Systems (DS) are fundamental to the modeling and understanding time evolving phenomena, and have application in physics, biology and control. As determining an analytical description of the dynamics is often difficult, data-driven approaches are preferred for identifying and controlling nonlinear DS with multiple equilibrium points. Identification of such DS has been treated largely as a supervised learning problem. Instead, we focus on an unsupervised learning scenario where we know neither the number nor the type of dynamics. We propose a Graph-based spectral clustering method that takes advantage of a velocity-augmented kernel to connect data points belonging to the same dynamics, while preserving the natural temporal evolution. We study the eigenvectors and eigenvalues of the Graph Laplacian and show that they form a set of orthogonal embedding spaces, one for each sub-dynamics. We prove that there always exist a set of 2-dimensional embedding spaces in which the sub-dynamics are linear and n-dimensional embedding spaces where they are quasi-linear. We compare the clustering performance of our algorithm to Kernel K-Means, Spectral Clustering and Gaussian Mixtures and show that, even when these algorithms are provided with the correct number of sub-dynamics, they fail to cluster them correctly. We learn a diffeomorphism from the Laplacian embedding space to the original space and show that the Laplacian embedding leads to good reconstruction accuracy and a faster training time through an exponential decaying loss compared to the state-of-the-art diffeomorphism-based approaches.
\end{abstract}



\section{Introduction}
\label{sec:introduction}
Relying on the mathematical framework of differential equations, \emph{Dynamical Systems} (DS) describe how a system evolves temporally and spatially. Their application span various fields, such as physics, biology, engineering, and economics. In recent years, DS have been successfully applied to model and control robots (e.g. \cite{heinzmann2003quantitative,corteville2007human}).

Finding an analytical description of the dynamics is, however, often difficult. This led researchers to turn to data-driven identification of the dynamics using machine learning algorithms. Data is usually provided by an expert, within the generic framework of learning from demonstration (\cite{billard2016learning}). Data can also be gathered from trial and error in a reinforcement learning framework (\cite{wabersich2018safe}).

From a machine learning perspective, approximating a DS can be framed as a regression problem. One estimates a non-linear function $\mathbf{f}: \mathbb{R}^d \rightarrow \mathbb{R}^d$, mapping the  d-dimensional input state $\mathbf{x}(t) \in \mathbb{R}^d$ to its time-derivative $\dot{\mathbf{x}}(t) \in \mathbb{R}^d$, such that:
\begin{equation}
  \dot{\mathbf{x}}(t) = \mathbf{f}(\mathbf{x}(t)).
\end{equation}
Training data consists of a set of trajectories, as examples of path integrals of the DS. These trajectories cover a limited portion of the state space. To ensure that the learned DS stably generalizes over regions not covered by the data represents one of the most challenging and active research area. One option to tackle this problem is to embed constraints explicitly in the algorithm so that the learned dynamics offers similar guarantees as those generated from control theory. One desirable property is stability at an equilibrium point, or \emph{attractor}. If $\mathbf{x}^*$ is the attractor, we wish to guarantee that
\begin{equation}
  \lim_{t \rightarrow \infty} \mathbf{x}(t) = \mathbf{x}^*, \quad
  \mathbf{f}(\mathbf{x}^*) = 0.
\end{equation}

\subsection{Learning Stable Dynamical Systems}
In the Stable Estimator of Dynamical Systems (SEDS) proposed by \cite{mohammadkhansari-zadeh_learning_2014}, the density estimation with Gaussian Mixture Models is reformulated to enforce constraints on the parameters of the Gaussian Mixtures, by imposing conditions derived via Lyapunov's second method for stability. The DS is then estimated through Gaussian Mixture Regression (GMR). This approach was, however, limiting. Constraints from a quadratic Lyapunov function were conservative and led to a poor approximation of the flow when the dynamics was highly nonlinear.

Accuracy and stability turned out to be conflicting objectives whilst constraints were derived from a quadratic Lyapunov function, and this prevented modeling dynamics that are non-monotonic, namely temporarily moving away from the attractor. To address this issue, other works used more complex expression for the Lyapunov function (e.g. \cite{mirrazavi2018compliant,figueroa_physicallyconsistent_2018}). An alternative to using Lyapunov stability is Contraction Theory (CT) introduced by \cite{lohmiller_contraction_1998}.
Stability under CT is less restrictive, as it follows a differential perspective and enforces solely that the flow contracts locally. Further, it does not require prior knowledge of the location of the attractor. \cite{ravichandar_learning_2017} used CT to reformulate the SEDS constraints. Similarly \cite{sindhwani_learning_2018} constraints a Support Vector regression problem with CT constraints. Although CT represents a promising approach, current works adopting it are limited to local stability guarantees making the approximated vector field unsuitable for regions with sparse or no data.

A third approach to tackle the problem of increasing accuracy while preserving stability is based on the idea of using latent representation to ease the stabilization of the DS. That is achieved via diffeomorphic mapping. One of the first attempts was offered in \cite{neumann_learning_2015}, which uses a diffeomorphic map to send a complex Lyapunov function, learned from the sampled trajectories, to a space where it appears quadratic. In this space, SEDS is applied, and the original vector field is recovered via the inverse diffeomorphic map. The two-step learning approach requires fitting both a Lyapunov function (or a diffeomorphism) and a DS from training data, increasing the number of tunable parameters and the overall learning time for non-convex optimization problems. \cite{perrin_fast_2016} follow a similar approach but apply the diffeomorphism directly to the DS using one single sampled trajectory. The diffeomorphism learning process is purely geometrical, namely only original and target points' position are considered. Reconstruction of the proper velocity profile is achieved via re-scaling the learned DS. The recent work proposed by \cite{rana_euclideanizing_2020} follows a similar approach but introduces a formulation of the optimization framework that includes dynamics information (e.g., velocity) within the process, providing for a one-step learning algorithm.

All the methods reviewed above focus on single-attractor DS, and, except for works using stability constraints based on CT-derived conditions, they require prior knowledge of the location of the attractor. Assuming single dynamics and single attractor DS considerably constrains the applicability of these methods to learn uni-modal dynamics. Embedding multiple dynamics in a single control law based on DS increases the complexity of the dynamics that we can model. Next, we review  recent efforts that have been dedicated to learning DS with multiple attractors.

\subsection{Learning Dynamical Systems with Multiple Attractors}
Learning DS with multiple-attractor can be achieved by explicitly partitioning the space to separate each dynamics and their respective attractors, as shown by \cite{shukla_augmentedsvm_2012}. This work requires knowing how many sub-dynamics exist in the system and the attractors' locations. Such knowledge may not always be easy to obtain.

Hence, if, for learning single-attractor DS, the main requirement is to know the location of the attractor, when learning multiple-attractor DS, correct labelling and classification of data points to distinguish the attractors and their associated dynamics is necessary. In realistic scenarios, proper segmentation and labeling might not be available either because data are generated by naive users or because the dataset is constructed by sampling demonstrations of complicated physical tasks (e.g., cleaning up a messy office). One cannot expect lay users to properly identify the number of sub-dynamics with the relative attractors location. Asking users to divide the task into sub-segment will be very cumbersome and prevent proper skill display and transfer; finally, even for expert users, it might often be difficult to classify the dynamics, especially when these require data that are not easy to interpret, such as force and haptic information.

In order to overcome this limitation, one option is to offer automatic segmentation and identification of the dynamics. Such an approach was followed by \cite{medina_learning_2017}, using Hidden Markov Models to extract automatically the attractors' positions of multiple-attractor DS. The algorithm assumes that such multiple-attractor DS can be thought of as a long sequence of multiple sub-dynamics to be performed one after the other. Such approach implicitly assumes time labeling of the points and can handle only multiple-attractor DS considered as a long sequence of dynamics. \cite{manschitz_mixture_2018} presented another interesting application of multiple-attractor DS. The DS-based control law is modeled as a weighted combination of linear systems, each of which is stable with respect to an attractor. The location of the attractors along with the parameters of each linear sub-dynamics is automatically derived via optimization problem. However the minimum number of attractors necessary to solve a specific task is a user-defined hyper-parameter that has to be known a priori.

To avoid providing prior information on both the number of dynamics and attractors,  we see a need for a fully \emph{unsupervised learning} approach to the identification of multiple-attractor DS. In this light, we offer an algorithm whose goal is to: (a) cluster the sub-dynamics within a certain data set, (b) find underlying interesting structure of the data that would allow to locate the attractors and ease the task of learning stable vector field. To do so, we seek to exploit structure discovery techniques to identify the number of sub-dynamics automatically.

\subsection{Manifold Learning for Latent Embedding Spaces of Dynamical Systems}
\emph{Manifold learning} techniques, such as \emph{Laplacian Eigenmaps} (\cite{belkin_laplacian_2003}) and \emph{Isometric Mapping} (\cite{tenenbaum_global_2000}), are particularly relevant to our problem. These techniques can generate Euclidean spaces recovering the intrinsic geometry of a certain manifold (e.g., \cite{tenenbaum_global_2000}). As in Kernel PCA (Principal Component Analysis) (\cite{scholkopf_learning_2002}), these methods are based on eigenvalue decomposition of a matrix. The matrix embeds information about the feature of the data, as encapsulated by the kernel. Depending on the kernel used, different features can be extracted by the manifold.

We find applications of manifold learning techniques for DS in biology for analyzing emergent dynamics behaviors in data measured from bioelectric signals (\cite{erem2016extensions}), or for person identification from ECG (\cite{sulam2017dynamical}); in control theory for data-driven time series analysis (\cite{shnitzer_diffusion_2017}); in finance for describing the characteristics of financial system (\cite{huang_learning_2018}); in chemistry for stochastic model of cellular chemotaxis (\cite{dsilva2018parsimonious}).

In the works reviewed above, the primary goal is to use manifold learning to reduce the dimensionality of the data and simplify the estimation of the DS. The scope of our work is different. We do not aim at reducing the dimensionality but rather at finding an embedding that can simplify the control of DS by linearizing a non-linear dynamics that would otherwise be difficult to stabilize.

The success of manifold learning depends heavily on the choice of kernel. We define a velocity-augmented kernel to extract the temporal evolution of data points. We generate the desired graph structure with this kernel and compute the associated Laplacian. We study the embedding spaces generated by the eigendecomposition of such a graph-based Laplacian matrix. We show that an analysis of the eigenvalues enables us to determine the number of underlying dynamics and to identify a set of eigenvectors that forms an embedding space in which the DS is linearized.

We validate our technique for sub-dynamics clustering and stable equilibria localization of multiple-attractor DS using theoretical and noisy instances of DS. We compare our algorithm to Kernel K-Means, Spectral Clustering, and Gaussian Mixtures. We showcase that even when these algorithms are provided with the correct number of sub-dynamics, they fail to cluster them correctly.

\section{Problem Formulation}
\label{sec:problem}
Let $\mathbf{x} \in \mathbb{R}^d$ be the state of a DS. Its temporal evolution is governed by the non-linear function $\mathbf{f}(\mathbf{x}(t))$:
\begin{equation}
    \mathbf{f}:\mathbb{R}^d \rightarrow \mathbb{R}^d, \quad \dot{\mathbf{x}}(t) = \mathbf{f}(\mathbf{x}(t)).
\end{equation}
\label{eq:DS_definition}
Assume that $\mathbf{f}$ is composed of $Q$ sub-dynamics, each asymptotically stable at one equilibrium point, the attractor.

\begin{figure}[ht]
    \centering
    \scalebox{.9}{\subimport{../figures/}{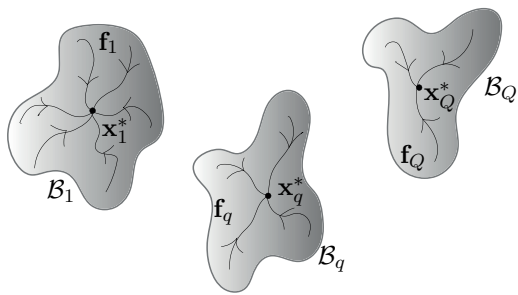}}
    \caption{\footnotesize Schematic representation of a multiple-attractor DS. Each sub-dynamics $\mathbf{f}_q$ takes within a sub-space $\mathcal{B}_q$ and converges towards the attractor $\mathbf{x}_q^*$.}
    \label{fig:multids_example}
\end{figure}
Let the $q$-th sub-dynamics be
\begin{equation}
    \dot{\mathbf{x}}(t) = \mathbf{f}_q(\mathbf{x}(t)) \quad \forall q \in [1,Q] \subset \mathbb{N}.
    \label{eq:subDS_definition}
\end{equation}
Each \emph{sub-dynamics}, $\mathbf{f}_q(\mathbf{x})$, is Lyapunov stable and there exists $\delta > 0$ such that if $\norm{\mathbf{x}(0) - \mathbf{x}^*_q} < \delta$, then
\begin{equation}
    \scalebox{1}{$
            \lim_{t \rightarrow \infty} \norm{\mathbf{x}(t) - \mathbf{x}^*_q}
        $}=0,
    \label{eqn:ds_stable}
\end{equation}
where $\mathbf{x}^*_q$ is the attractor of the sub-dynamics $\mathbf{f}_q(\mathbf{x})$, Fig.~\ref{fig:multids_example}.

We know neither the number of sub-dynamics $Q$, the shape of the dynamics $\mathbf{f}_q$, nor the number and location of the attractors. All we have at our disposal is a finite set of observations, samples of trajectories from the DS. These are constituted by \emph{unlabeled} position-velocity pairs $\mathcal{V} = \lbrace \nu_i = (\mathbf{x}_i,\dot{\mathbf{x}}_i), i=1,\dots,M\rbrace$ sampled from $\mathbf{f}(\mathbf{x})$ at a constant frequency, $f_{sampling}$. Among these trajectories, a subset belongs to some of the sub-dynamics, but we know neither to which dynamics they belong nor how many trajectories belong to the same dynamics. All sampled trajectories end at the attractor of the associated sub-dynamics.

This paper presents a method by which we can a)  determine the number $Q$ of sub-dynamics present in the dataset, b) identify to which sub-dynamics each data point belongs and the attractor of the corresponding sub-dynamics and c) project each dynamics into a separate subspace in which the dynamics is linear. The method is based on a representation of the data through a graph and exploits properties from the eigendecomposition of the graph-based Laplacian matrix, as we present next.

\section{Graph Embedding and Linearization of Dynamical Systems}
\label{sec:laplacian}
\begin{figure}[ht]
    \centering
    \scalebox{.8}{\begin{tikzpicture}
    \node[circle, draw, label=right:{$\nu_1^k$}] at (0,0) (v1) {};
    \node[circle, draw] at (0.2,1) (v2) {};
    \node[circle, draw, label=left:{$\nu_{p_k}^k$}] at (1,2) (v3) {};

    \node[circle, draw] at (4,2) (v4) {};
    \node[circle, draw] at (3,1.5) (v5) {};
    \node[circle, draw] at (2,2) (v6) {};

    \node[circle, draw] at (-.5,3) (v7) {};
    \node[circle, draw] at (0.5,3.5) (v8) {};
    \node[circle, draw] at (1.5,3) (v9) {};

    \draw (v1) -- (v2) -- (v3);
    \draw (v4) -- (v5) -- (v6);
    \draw (v7) -- (v8) -- (v9);

    \draw (v3) -- (v6) -- (v9) -- (v3);
\end{tikzpicture}}
    \caption{\footnotesize Each sub-dynamics is embedded in a graph where each trajectory forms a path connected by a set of termination nodes that form a cyclic path around the attractor. $\nu_1^k$ $\nu_{p_k}^k$ are the first and last of the $k$-th path graphs, respectively. $p_k$ is the number of nodes with the $k$-th path graph.}
    \label{fig:graph_structure}
\end{figure}
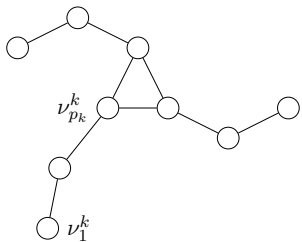

Consider a symmetric weighted graph $G(\mathcal{V},\mathcal{E})$. The nodes $\nu_i \in \mathcal{V}$ represent the data points $\nu_i = \lbrace \mathbf{x}_i,\dot{\mathbf{x}}_i \rbrace$, $i=\lbrace 1,\dots,M \rbrace$ sampled from our DS, where $M = \sum_{q=1}^Q p_q$ with $p_q$ being the number of data points sampled from each of the sub-dynamics and $Q$ being the number of sub-dynamics. $\mathcal{E}$ is the set of edges $e_{ij} \in \mathcal{E}$ connecting nodes $\nu_i$ and $\nu_j$. For a generic DS containing $Q$ sub-dynamics, the graph G is the result of the union of $Q$ sub-graphs $\mathcal{F}$ such that \( G = \mathcal{F}_1 \cup \mathcal{F}_2 \cup \dots \cup \mathcal{F}_Q \) and \( \mathcal{F}_1 \cap \mathcal{F}_2 \cap \dots \cap \mathcal{F}_Q = \emptyset \). Each sub-graph $\mathcal{F}$ is given by the composition of $K$ \emph{path graphs}, equal to the number of sampled trajectories. We order the nodes (vertices) of $\mathcal{F}$ monotonically as $\lbrace \nu^k_1, \nu^k_2, \dots, \nu^k_{p_k} \rbrace$, $k=1,\dots,K$, such that each $\nu^k_{p_k}$ is the last node of the $k$-th path graph and $p_k > 1$. The edges are $e^k_{ij} = \lbrace \nu^k_i, \nu^k_{i+1} \rbrace$. The K nodes $\lbrace \nu^1_{p_1},\dots, \nu^K_{p_K} \rbrace$ form a \emph{cyclic graph} (or simple circuit). Each node belonging to the cyclic path has degree $3$. All other nodes along each path graph have degree 1 or 2. The nodes are numbered as
\begin{equation*}
    \{\nu_1^1,\dotsc\nu_{p_1}^1,\nu_1^2,\dotsc,\nu_{p_2}^2,\dotsc, \nu_1^K,\dotsc,\nu_{p_K}^K \} \sim\{1,2, \dotsc , M\}
\end{equation*}
With reference to Fig.~\ref{fig:graph_structure}, $\lbrace \nu_1^k, \dots, \nu_{p_k}^k \rbrace$ represents the nodes of the $k$-th path graph where $p_k$ is the number of samples within the $k$-th path graph.

The Graph Laplacian matrix, $L(G) = D(G) - A(G)$, where $A(G)$ is the adjacency matrix
\begin{equation}
    A_{ij} =
    \begin{cases}
        1, & \text{if } e_{ij} \in  \mathcal{E} \\
        0, & \text{otherwise}
    \end{cases}
\end{equation}
and $D(G)$ is a diagonal matrix composed of the sum of the rows of $A(G)$
\begin{equation}
    D_{ii} = \sum_j A_{ij}.
\end{equation}
$L(G)$ is positive semi-definite and, hence, admits an eigendecomposition. The eigenvalues of the $L(G)$ are non-negative and we have at least one eigenvalue equal to zero. The multiplicity of the eigenvalue zero allows to determine the number of sub-dynamics embedded in the graph.

\begin{proposition}
    \label{prop:multiplicity_zero}
    The eigendecomposition of the Laplacian generates a set of $M$ positive eigenvalues \( \Lambda = \lbrace 0 = \lambda_0^1=\lambda_0^2=\dots=\lambda_0^Q < \lambda_1 \le \dots \le \lambda_{M-Q} \rbrace \). The multiplicity of the eigenvalue $0$ is equal to the number of sub-dynamics $Q$.
\end{proposition}

\begin{proof}
    \footnotesize
    The multiplicity of the eigenvalue zero of the Laplacian matrix is equal to the number of connected components, hence, by construction of $G$ is equal to the number of sub-dynamics $Q$.
\end{proof}

As the graph $G$ is composed of a set of disconnected components, the eigendecomposition of $L(G)$ can be performed by blocks and we can associate a subset of eigenvectors to each of the $Q$ blocks. We show in the next subsection that each set of eigenvectors of $L(G)$ generates $Q$ separate sub-spaces in which each sub-dynamics is linearized.

Observe first that, in each of the graph connected component, the nodes are connected in such a way that each trajectory forms a path in the graph. With $K$ trajectories for each sub-dynamics, we have $K$ paths. Each trajectory is connected to another trajectory through the last node of each path graph.

\subsection{Determining the Eigenvectors that Generate a Linear Embedding}
We are now ready to present the main results of this paper, namely that a well-chosen set of eigenvectors of the graph Laplacian generates a linear embedding of the sub-dynamics.

We start by showing that, for each path in the graph, the corresponding entries in the eigenvectors follow a recursive law.
\begin{lemma}
    \label{lem:recursion}
    Consider $K$ different path graphs that form a graph $G$ with one connected component. Let $\mathbf{u}$ be an eigenvector of the Graph Laplacian $L(G)$. The elements of $\mathbf{u}$ entail $K$ sets of scalars $\lbrace u^k_{1},\dots,u^k_{p_k} \rbrace$, corresponding to the nodes within the $k$-th path graph. Each set follows the recursive relation:
    \begin{align}
        u^k_1 & \in \mathbb{R}, \notag                                                              \\
        u^k_2 & = (1 - \lambda) u^k_1, \notag                                                       \\
        u^k_n & = (2 - \lambda) u^k_{n-1} - u^k_{n-2}, \quad \operatorname{for } n = 3, \dots, p_k.
        \label{eqn:recursion}
    \end{align}
\end{lemma}

\begin{proof}
    \footnotesize
    See Appendix~\ref{app:proof_lem2}.
\end{proof}

\begin{lemma}
    \label{lem:chebyshev_poly}
    In each of the $K$ sets of elements in $\mathbf{u}$ denoted as $\lbrace u^k_{1},\dots,u^k_{p_k} \rbrace$, the recursive relation in Eq.~\ref{eqn:recursion} corresponds to a combination of Chebyshev polynomials $T$ and $V$, of first and second kind, given by:
    \begin{equation}
        u^k_{n} = u^k_{1}\left[ T_n(\lambda) - \frac{\lambda}{2}V_{n-1}(\lambda) \right] \quad \text{for } n \ge 1.
        \label{eqn:chebyshev_poly1}
    \end{equation}
    Let $ \theta \coloneqq \cos^{-1}{(1-\frac{\lambda}{2})}$. Eq.~\ref{eqn:chebyshev_poly1} can be expressed as the following combination of trigonometric functions:
    \begin{equation}
        u^k_{n}(\lambda, \theta(\lambda)) = u^k_1 \left[ \cos ( (n-1) \theta ) - \frac{\lambda}{2} \frac{\sin((n-1) \theta)}{\sin(\theta)} \right].
        \label{eqn:chebyshev_poly2}
    \end{equation}
\end{lemma}

\begin{proof}
    \footnotesize
    See Appendix~\ref{app:proof_lem3}.
\end{proof}

Next, we show that, when we select a specific set of eigenvectors, the coordinates of the path graphs in these eigenvectors either grow or decrease monotonically. This a key property to prove the linearity of the embedding.

\begin{proposition}
    \label{prop:monotonicity}
    If each of the path graphs have at least 3 nodes, $p_k \geq 3$ $\forall k$, the coordinates of a path graph $\lbrace u^k_{1}, \dots, u^k_{p_k} \rbrace$ in the eigenvectors $\mathbf{u}$ with corresponding eigenvalues $0 < \lambda \leq 2\left[ 1 - \text{cos}\left(\frac{\pi}{p_k-\frac{1}{2}} \right) \right]$ , either increase or decrease monotonically within each path graph, s.t.:
    \begin{equation}
        u_n^k \ge (\le) u_{n+1}^k \quad \text{for } n = 1, \dots, p_k, \quad \text{and }  k \in \{1,\dots, K\}
    \end{equation}
\end{proposition}

\begin{proof}
    \footnotesize
    See Appendix~\ref{app:proof_prop4}.
\end{proof}

Next, we prove that, when all the path graphs have the same length, i.e $p_k=N$, $\forall k$, there always exist, at least, one pair of eigenvectors with properties of Prop.~\ref{prop:monotonicity}, by showing that there exists an eigenvalue $\lambda \leq 2\left[ 1 - \text{cos}\left(\frac{\pi}{p_k-\frac{1}{2}} \right) \right] $ with algebraic multiplicity $2$.

The proof is done in two steps: first, we show that the spectrum of $L(G)$ presents repeated eigenvalues; second, we show that there is at least one repeated eigenvalue upper bounded by a value inferior to the monotonicity constraint introduced in Prop.~\ref{prop:monotonicity}.

Following \cite{gupta2022learning}, observe that the graph Laplacian, analyzed in this work, has the following structure:
\begin{equation}
    L(G) = 2 I - J(G),
\end{equation}
where $J$ has a block circulant matrix structure of the following type
\begin{equation}
    J(G) = \text{circ} (B_0, B_1, \underbrace{\mathbf{0}^{(N \times N)}}_{K-3 \text{ times}}, B_1),
    \label{eqn:matrix_j}
\end{equation}
with $K$ the number of path graphs and $N$ the number of nodes in each path graph. The matrices $B_0 \in \mathbb{R}^{N \times N}$ is a tri-diagonal matrix of the form
\begin{equation}
    B_0 = \begin{bmatrix}
        1 & 1      &        &        &    \\
        1 & 0      & \ddots &        &    \\
          & \ddots & \ddots & \ddots &    \\
          &        & \ddots & 0      & 1  \\
          &        &        & 1      & -1
    \end{bmatrix}
\end{equation}
and $B_1 \in \mathbb{R}^{N \times N}$ is defined as
\begin{equation}
    B_1 = \begin{bmatrix}
        \mathbf{0}^{(N-1 \times N-1)} & 0 \\
        0                             & 1
    \end{bmatrix}.
\end{equation}
We repeat  results from \cite{gupta2022learning} in Proposition \ref{prop:multiplicity_eig} and \ref{prop:eig_bound}:
\begin{proposition}
    \label{prop:multiplicity_eig}
    If the number of paths $K$ is even, $\frac{K}{2}-1$ eigenvalues of $L(G)$ have multiplicity $2$ and for $K$ odd, $\frac{K-1}{2}$ eigenvalues of $L(G)$ have multiplicity $2$.
\end{proposition}

\begin{proof}
    \footnotesize
    See Appendix \ref{app:proof_prop5}.
\end{proof}

\begin{proposition}
    \label{prop:eig_bound}
    The second smallest eigenvalue of the Graph Laplacian $L(G)$ with algebraic multiplicity $2$ is denoted by $\lambda_{min}(L)$ and is bounded above and below as follows:
    \begin{equation}
        1-2\cos\left( \frac{\pi}{N}\right) < \lambda_{min}(L) \leq 2 \left( 1-\cos\left( \frac{\pi}{N-\frac{1}{2}}\right) \right)
        \label{eqn:lambda_bound}
    \end{equation}
    where $N$ is the number of nodes in each of the $K$ paths.
\end{proposition}

\begin{proof}
    \footnotesize
    See Appendix \ref{app:proof_prop6}.
\end{proof}

We have shown that there exist at least a pair of eigenvectors with same eigenvalues in which the corresponding entries of our path graph grow or decrease monotonically.

Next, we show that the DS is linear in the  embedding formed by these two eigenvectors.
\begin{corollary}
    For each eigenvector $u$ with associated eigenvalue $0<\lambda<1$, the rate of change $r(u_n)=u_n-u_{n-1}$ along the entries $\lbrace u_1, \dots, u_{p_k} \rbrace$, for each path $k=1, \dots, K$ decreases monotonically according to:
    \begin{equation}
        \dot{r}(u)= - \lambda u.
        \label{Eq:rate_change}
    \end{equation}
    \label{corrolary:rate-change}
\end{corollary}

\begin{proof}
    \footnotesize
    $r(u_{n+1})=u_{n}-u_{n-1}-\lambda u_{n} = r(u_n) -\lambda u_n.$
\end{proof}

\begin{theorem}
    \label{thm:linearity}
    If the graph Laplacian $L(G)$ admits a set of eigenvectors $\mathbf{u}$, with same eigenvalue, the $K$ paths of the graph expressed in the domain spanned by $\mathbf{u}$ form $K$ lines.
\end{theorem}

\begin{proof}
    \footnotesize
    Consider a group of three consecutive nodes $\lbrace \nu_{n-1},\nu_n,\nu_{n+1} \rbrace$ of one of the paths of the graph. The coordinate of these points when projected in the space spanned by two eigenvectors $u^z$ and $u^k$ are $\lbrace u_{n-1}^{z,k}, u_n^{z,k}, u_{n+1}^{z,k} \rbrace$. We show that the three points are on the same line, as illustrated in  Fig.\ref{fig:slope_constant}.
    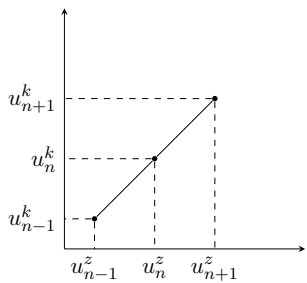
\begin{figure}[ht]
        \centering
        \scalebox{.8}{\begin{tikzpicture}
    \node[circle, fill, inner sep=0.9] at (0.5,0.5) (a) {};
    \node[circle, fill, inner sep=0.9] at (1.5,1.5) (b) {};
    \node[circle, fill, inner sep=0.9] at (2.5,2.5) (c) {};

    \draw[-stealth] (0,0) -- (4,0);
    \draw[-stealth] (0,0) -- (0,4);

    \draw[dashed] (a) -- (0.5,0) node[below] {$u^z_{n-1}$};
    \draw[dashed] (b) -- (1.5,0) node[below] {$u^z_{n}$};
    \draw[dashed] (c) -- (2.5,0) node[below] {$u^z_{n+1}$};

    \draw[dashed] (a) -- (0,0.5) node[left] {$u^k_{n-1}$};
    \draw[dashed] (b) -- (0,1.5) node[left] {$u^k_{n}$};
    \draw[dashed] (c) -- (0,2.5) node[left] {$u^k_{n+1}$};

    \draw[] (a) -- (b) -- (c);
\end{tikzpicture}}
        \caption{\footnotesize  Coordinates of three points on two eigenvectors with constant rate of change, resulting in a linear path.}
        \label{fig:slope_constant}
    \end{figure}
    We prove that these coordinates grow at the same rate for all the eigenvectors that have equal eigenvalues $\lambda<1$. Hence the spacing across each group of coordinates on each axis is the same, as illustrated in Fig.\ref{fig:slope_constant}.

    In the following, we drop the upper index referring to the specific eigenvector. For any eigenvector with  eigenvalue $0<\lambda<1$, using Eq.~\ref{eqn:ui_to}, we have
    \begin{equation}
        u_{n+1} = \gamma u_{n} - u_{n-1},
        \label{eqn:recurr_simple}
    \end{equation}
    with $\gamma = 2-\lambda, \gamma >1$. Furthermore, for the first two points on the path graph, we have, from Eq.~\ref{eqn:u1_to}: $u_2 = \delta  u_1$, with $\delta=1-\lambda, \delta >0$. Combining these two equations, the third point can be expressed as a function of the first coordinate $u_1$ only: $u_3 = \gamma (\delta  u_1) - u_1 = (\gamma \delta-1)  u_1$. Similarly the fourth point can be expressed solely as a function of the first coordinate $u_1$. We have: $u_4 = \gamma u_3 - u_2= \gamma (\gamma \delta -1)  u_1 -  \delta  u_1) = (\gamma (\gamma \delta -1) - \delta)  u_1$.
    The same reasoning holds for all points and hence by induction, we have:
    \begin{equation}
        u_n = P_{n}(\lambda) u_1.
        \label{eqn:poly_linear}
    \end{equation}
    $P_n(\lambda)$ is a polynomial of order $n-2$  that depends only on the eigenvalue $\lambda$. If the eigenvectors considered have same eigenvalue $\lambda$, the coordinates along each axis grow with the same series of polynomial $P_{n}(\lambda)$. Hence, all points are located on a line. For a pair of eigenvectors $z,k$, the line has slope $\frac{u^k_1}{u^z_1}$.
\end{proof}

Note that, while Thm.~\ref{thm:linearity} holds for all the eigenvectors in the spectrum of $L(G)$, the original ordering of the nodes along each path graph is preserved  only when  $\lambda < 2\left[ 1 - \text{cos}\left( \frac{\pi}{p_k-\frac{1}{2}} \right) \right]$. For the eigenvectors with larger $\lambda$, the coordinates of the nodes change sign within the path graph. Hence, while the path may still appear linear in the embedding, the order of the nodes will not be preserved anymore.

\subsection{Creating a Linear Embedding}
We summarize how the above theoretical results allow us to find embedding spaces so that we can separate each of the $Q$ sub-dynamics and that they are linear in each embedding.

First, let us recall that, since the matrix $L(G)$ can be decomposed by block with each block representing data points that belong to each sub-dynamics $\mathbf{f}_q$, we can generate a set of $Q$ separate embedding spaces by using only the subset of eigenvectors associated to each sub-block. By orthogonality of the eigenvectors, all other sub-dynamics $\mathbf{f}_k$, $k \neq q$ will project to the origin in the embedding space of the $q$-th dynamics. We now need to determine the existence of such embedding spaces.

Given a K-paths graph, from Prop.~\ref{prop:multiplicity_eig}, for $N\geq 3$ and $K \geq 3$, there is at least one pair of eigenvector with eigenvalue $0<\lambda\leq 2\left[1 - \text{cos}\left( \frac{\pi}{N-\frac{1}{2}} \right) \right]$. It follows that if the $Q$ dynamics of the graph have at least $N\geq 3$ nodes and $K \geq 3$, we have $2Q$ eigenvectors with properties of Prop.~\ref{prop:monotonicity} and hence $Q$ embedding spaces.

\begin{figure}[ht]
    \centering
    \subfloat[]{
        \scalebox{.6}{\begin{tikzpicture}
    \begin{axis}[
            xlabel=$x$,
            ylabel=$y$,
            height = 5cm,
            scatter/classes={%
                    1={mark=*,black,fill=red,scale=0.7},%
                    2={mark=*,black,fill=cyan,scale=0.7}},
            every node near coord/.append style={font=\tiny},
        ]
        \addplot[only marks,mark=*,nodes near coords={\labelz},
            visualization depends on={value \thisrowno{0}\as\labelz}, mark size=1] table [x=x, y=y]{\relativepath ../figures/data/toy_3traj.csv}
        node[pos=0, inner sep=0, outer sep=0] (n1) {}
        node[pos=0.0667, inner sep=0, outer sep=0] (n2) {}
        node[pos=0.1333, inner sep=0, outer sep=0] (n3) {}
        node[pos=0.2000, inner sep=0, outer sep=0] (n4) {}
        node[pos=0.2667, inner sep=0, outer sep=0] (n5) {}
        node[pos=0.3333, inner sep=0, outer sep=0] (n6) {}
        node[pos=0.4000, inner sep=0, outer sep=0] (n7) {}
        node[pos=0.4667, inner sep=0, outer sep=0] (n8) {}
        node[pos=0.5533, inner sep=0, outer sep=0] (n9) {}
        node[pos=0.6200, inner sep=0, outer sep=0] (n10) {}
        node[pos=0.6967, inner sep=0, outer sep=0] (n11) {}
        node[pos=0.7533, inner sep=0, outer sep=0] (n12) {}
        node[pos=0.8500, inner sep=0, outer sep=0] (n13) {}
        node[pos=0.9067, inner sep=0, outer sep=0] (n14) {}
        node[pos=1.0000, inner sep=0, outer sep=0] (n15) {}
        ;

        \draw (n1) -- (n2) -- (n3) -- (n4) -- (n5);
        \draw (n6) -- (n7) -- (n8) -- (n9) -- (n10);
        \draw (n11) -- (n12) -- (n13) -- (n14) -- (n15);

        \draw (n5) -- (n10) -- (n15) -- (n5);
    \end{axis}
\end{tikzpicture}}
        \label{subfig:toy3_data}
    }
    \subfloat[]{
        \raisebox{-0.75ex}{\scalebox{.6}{\begin{tikzpicture}
    \begin{axis}[
            xlabel={axes $\#$},
            ylabel={$\lambda$},
            xtick = {1,...,9},
            y tick label style={
                    /pgf/number format/.cd,
                    fixed,
                    fixed zerofill,
                    precision=1,
                    /tikz/.cd
                },
            height = 5cm,
        ]
        \addplot table [x=label, y=lambda, select coords between index={0}{8}]{\relativepath ../figures/data/toy_3traj.csv};
        \draw[dashed] (axis cs:1,0.25) -- (axis cs:6,0.25) node[right]{Spectral gap};
    \end{axis}
\end{tikzpicture}}}
        \label{subfig:toy3_eigval}
    }
    \subfloat[]{
        \raisebox{0.25ex}{\scalebox{.6}{\begin{tikzpicture}
    \begin{groupplot}[
            group style = {group size = 1 by 2},
            height = 3cm,
            width = 6cm,
            ytick=\empty,
            xtick = {1,...,25},
            xticklabel style = {font=\tiny},
        ]
        \nextgroupplot[ylabel=$\mathbf{u}_i^2$] 
        \addplot[scatter] table [x=label, y=e2]{\relativepath ../figures/data/toy_3traj.csv};
        \nextgroupplot[ylabel=$\mathbf{u}_i^3$]
        \addplot[scatter] table [x=label, y=e3]{\relativepath ../figures/data/toy_3traj.csv};
    \end{groupplot}
\end{tikzpicture}}}
        \label{subfig:toy3_eigvec}
    }
    \subfloat[]{
        \raisebox{-0.25ex}{\scalebox{.6}{\begin{tikzpicture}
    \begin{axis}[
            xlabel=$u_2$,
            ylabel=$u_3$,
            height = 5cm,
            scatter/classes={%
                    1={mark=*,black,fill=red,scale=0.7},%
                    2={mark=*,black,fill=cyan,scale=0.7}},
            every node near coord/.append style={font=\tiny},
        ]
        \addplot[only marks,mark=*,nodes near coords={\labelz},
            visualization depends on={value \thisrowno{0}\as\labelz}, mark size=1] table [x=e2, y=e3]{\relativepath ../figures/data/toy_3traj.csv}
        node[pos=0, inner sep=0, outer sep=0] (n1) {}
        node[pos=0.0667, inner sep=0, outer sep=0] (n2) {}
        node[pos=0.1333, inner sep=0, outer sep=0] (n3) {}
        node[pos=0.2000, inner sep=0, outer sep=0] (n4) {}
        node[pos=0.2667, inner sep=0, outer sep=0] (n5) {}
        node[pos=0.3333, inner sep=0, outer sep=0] (n6) {}
        node[pos=0.4000, inner sep=0, outer sep=0] (n7) {}
        node[pos=0.4667, inner sep=0, outer sep=0] (n8) {}
        node[pos=0.5533, inner sep=0, outer sep=0] (n9) {}
        node[pos=0.6200, inner sep=0, outer sep=0] (n10) {}
        node[pos=0.6967, inner sep=0, outer sep=0] (n11) {}
        node[pos=0.7533, inner sep=0, outer sep=0] (n12) {}
        node[pos=0.8500, inner sep=0, outer sep=0] (n13) {}
        node[pos=0.9067, inner sep=0, outer sep=0] (n14) {}
        node[pos=1.0000, inner sep=0, outer sep=0] (n15) {}
        ;

        \draw (n1) -- (n2) -- (n3) -- (n4) -- (n5);
        \draw (n6) -- (n7) -- (n8) -- (n9) -- (n10);
        \draw (n11) -- (n12) -- (n13) -- (n14) -- (n15);

        \draw (n5) -- (n10) -- (n15) -- (n5);
    \end{axis}
\end{tikzpicture}}}
        \label{subfig:toy3_embedding2d}
    }
    \caption{\footnotesize (a) Toy graph composed by 3 path graphs connected each other with a cycle graph passing through nodes $\lbrace 5,10,15 \rbrace$; (b) Spectrum of the first 9 eigenvalues of the Laplacian applied to graph in Fig.~\ref{subfig:toy3_data}; (c) Entries of the first 2 components $\lbrace \mathbf{u}^2,\mathbf{u}^3 \rbrace$; (d) 2D embedding space reconstructed using components $\lbrace \mathbf{u}^2,\mathbf{u}^3 \rbrace$.}
    \label{fig:ex_toy3}
\end{figure}
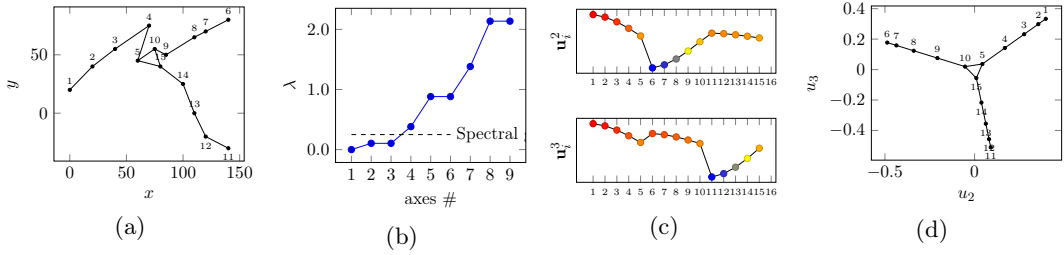

The properties stated in the previous propositions are illustrated in Fig.~\ref{subfig:toy3_data}.
We see a three-path graph connected by a cycle path across nodes $\lbrace 5,10,15 \rbrace$.The entries of the second and third eigenvectors (we exclude the first eigenvector, as it corresponds to the eigenvalue zero) are displayed in Fig.~\ref{subfig:toy3_eigvec}. To ease the interpretation, we enumerate the data points in increasing order as we move from the start to the end of each path. We plot the entries of the eigenvectors against the point label. Observe that, within each path graph, the corresponding entries on the eigenvectors are either strictly increasing or decreasing. Further, observe that the spectrum of the Laplacian, see Fig.~\ref{subfig:toy3_eigval}, entails a pair of eigenvalues with algebraic multiplicity equal to $2$. The embedding space reconstructed using the eigenvectors $\lbrace \mathbf{u}^2, \mathbf{u}^3 \rbrace$ corresponding to these two eigenvalues is shown in Fig.~\ref{subfig:toy3_embedding2d}. Observe that the structure now entails 3 paths all of which are linear. The original non-linear structure is hence linear in the embedding space.

In practice, we observe that several pairs of eigenvectors provide these embedding spaces. We also observe that the third and above eigenvalues are numerically close to the pair of smallest eigenvalues, and hence allow quasi-linear embedding spaces larger than 2 dimensions and up to $K-1$. This follows by observing that the polynomial $P_n(\lambda)$ in Eq.~\ref{eqn:poly_linear} is continuous and continuously differentiable up to $N-1$.

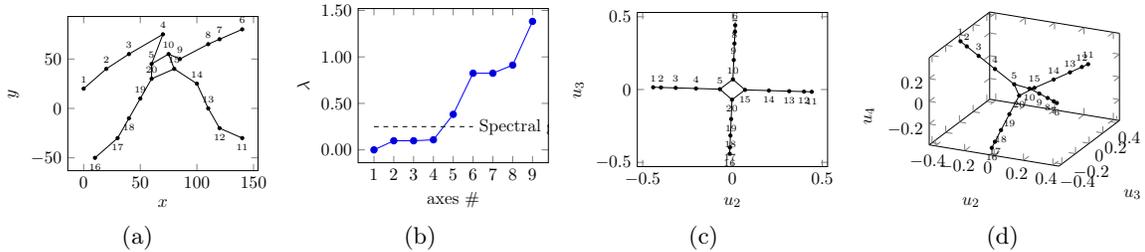
\begin{figure}[ht]
    \centering
    \subfloat[]{
        \scalebox{.6}{\begin{tikzpicture}
    \begin{axis}[
            xlabel=$x$,
            ylabel=$y$,
            height = 5cm,
            scatter/classes={%
                    1={mark=*,black,fill=red,scale=0.7},%
                    2={mark=*,black,fill=cyan,scale=0.7}},
            every node near coord/.append style={font=\tiny},
        ]
        \addplot[only marks,mark=*,nodes near coords={\labelz},
            visualization depends on={value \thisrowno{0}\as\labelz}, mark size=1] table [x=x, y=y]{\relativepath ../figures/data/toy_4traj.csv}
        node[pos=0     , inner sep=0, outer sep=0] (n1) {}
        node[pos=0.0500, inner sep=0, outer sep=0] (n2) {}
        node[pos=0.1000, inner sep=0, outer sep=0] (n3) {}
        node[pos=0.1500, inner sep=0, outer sep=0] (n4) {}
        node[pos=0.2000, inner sep=0, outer sep=0] (n5) {}
        node[pos=0.2500, inner sep=0, outer sep=0] (n6) {}
        node[pos=0.3000, inner sep=0, outer sep=0] (n7) {}
        node[pos=0.3500, inner sep=0, outer sep=0] (n8) {}
        node[pos=0.4000, inner sep=0, outer sep=0] (n9) {}
        node[pos=0.4500, inner sep=0, outer sep=0] (n10) {}
        node[pos=0.5000, inner sep=0, outer sep=0] (n11) {}
        node[pos=0.6000, inner sep=0, outer sep=0] (n12) {}
        node[pos=0.6500, inner sep=0, outer sep=0] (n13) {}
        node[pos=0.7000, inner sep=0, outer sep=0] (n14) {}
        node[pos=0.7500, inner sep=0, outer sep=0] (n15) {}
        node[pos=0.8000, inner sep=0, outer sep=0 ] (n16) {}
        node[pos=0.8500, inner sep=0, outer sep=0 ] (n17) {}
        node[pos=0.9000, inner sep=0, outer sep=0 ] (n18) {}
        node[pos=0.9500, inner sep=0, outer sep=0 ] (n19) {}
        node[pos=1.0000, inner sep=0, outer sep=0] (n20) {}
        ;

        \draw (n1) -- (n2) -- (n3) -- (n4) -- (n5);
        \draw (n6) -- (n7) -- (n8) -- (n9) -- (n10);
        \draw (n11) -- (n12) -- (n13) -- (n14) -- (n15);
        \draw (n16) -- (n17) -- (n18) -- (n19) -- (n20);

        \draw (n5) -- (n10) -- (n15) -- (n20) -- (n5);
    \end{axis}
\end{tikzpicture}}
        \label{subfig:toy4_data}
    }
    \subfloat[]{
        \scalebox{.6}{\begin{tikzpicture}
    \begin{axis}[
            xlabel={axes $\#$},
            ylabel={$\lambda$},
            xtick = {1,...,9},
            y tick label style={
                    /pgf/number format/.cd,
                    fixed,
                    fixed zerofill,
                    precision=2,
                    /tikz/.cd
                },
            height = 5cm
        ]
        \addplot table [x=label, y=lambda, select coords between index={0}{8}]{\relativepath ../figures/data/toy_4traj.csv};
        \draw[dashed] (axis cs:1,0.25) -- (axis cs:6,0.25) node[right]{Spectral gap};
    \end{axis}
\end{tikzpicture}}
        \label{subfig:toy4_eigval}
    }
    \subfloat[]{
        \scalebox{.6}{\begin{tikzpicture}
    \begin{axis}[
            xlabel=$u_2$,
            ylabel=$u_3$,
            height = 5cm,
            scatter/classes={%
                    1={mark=*,black,fill=red,scale=0.7},%
                    2={mark=*,black,fill=cyan,scale=0.7}},
            every node near coord/.append style={font=\tiny},
        ]
        \addplot[only marks,mark=*,nodes near coords={\labelz},
            visualization depends on={value \thisrowno{0}\as\labelz}, mark size=1] table [x=e2, y=e3]{\relativepath ../figures/data/toy_4traj.csv}
        node[pos=0     , inner sep=0, outer sep=0] (n1) {}
        node[pos=0.0500, inner sep=0, outer sep=0] (n2) {}
        node[pos=0.1000, inner sep=0, outer sep=0] (n3) {}
        node[pos=0.1500, inner sep=0, outer sep=0] (n4) {}
        node[pos=0.2000, inner sep=0, outer sep=0] (n5) {}
        node[pos=0.2500, inner sep=0, outer sep=0] (n6) {}
        node[pos=0.3000, inner sep=0, outer sep=0] (n7) {}
        node[pos=0.3500, inner sep=0, outer sep=0] (n8) {}
        node[pos=0.4000, inner sep=0, outer sep=0] (n9) {}
        node[pos=0.4500, inner sep=0, outer sep=0] (n10) {}
        node[pos=0.5000, inner sep=0, outer sep=0] (n11) {}
        node[pos=0.6000, inner sep=0, outer sep=0] (n12) {}
        node[pos=0.6500, inner sep=0, outer sep=0] (n13) {}
        node[pos=0.7000, inner sep=0, outer sep=0] (n14) {}
        node[pos=0.7500, inner sep=0, outer sep=0] (n15) {}
        node[pos=0.8000, inner sep=0, outer sep=0 ] (n16) {}
        node[pos=0.8500, inner sep=0, outer sep=0 ] (n17) {}
        node[pos=0.9000, inner sep=0, outer sep=0 ] (n18) {}
        node[pos=0.9500, inner sep=0, outer sep=0 ] (n19) {}
        node[pos=1.0000, inner sep=0, outer sep=0] (n20) {}
        ;

        \draw (n1) -- (n2) -- (n3) -- (n4) -- (n5);
        \draw (n6) -- (n7) -- (n8) -- (n9) -- (n10);
        \draw (n11) -- (n12) -- (n13) -- (n14) -- (n15);
        \draw (n16) -- (n17) -- (n18) -- (n19) -- (n20);

        \draw (n5) -- (n10) -- (n15) -- (n20) -- (n5);
    \end{axis}
\end{tikzpicture}}
        \label{subfig:toy4_embedding2d}
    }
    \subfloat[]{
        \scalebox{.6}{\begin{tikzpicture}
    \begin{axis}[
            xlabel=$u_2$,
            ylabel=$u_3$,
            zlabel=$u_4$,
            height = 5cm,
            scatter/classes={%
                    1={mark=*,black,fill=red,scale=0.7},%
                    2={mark=*,black,fill=cyan,scale=0.7}},
            every node near coord/.append style={font=\tiny},
        ]
        \addplot3[only marks,mark=*,nodes near coords={\labelz},
            visualization depends on={value \thisrowno{0}\as\labelz}, mark size=1] table [x=e2, y=e3, z=e4]{\relativepath ../figures/data/toy_4traj.csv}
        node[pos=0     , inner sep=0, outer sep=0] (n1) {}
        node[pos=0.0500, inner sep=0, outer sep=0] (n2) {}
        node[pos=0.1000, inner sep=0, outer sep=0] (n3) {}
        node[pos=0.1500, inner sep=0, outer sep=0] (n4) {}
        node[pos=0.2000, inner sep=0, outer sep=0] (n5) {}
        node[pos=0.2500, inner sep=0, outer sep=0] (n6) {}
        node[pos=0.3000, inner sep=0, outer sep=0] (n7) {}
        node[pos=0.3500, inner sep=0, outer sep=0] (n8) {}
        node[pos=0.4000, inner sep=0, outer sep=0] (n9) {}
        node[pos=0.4500, inner sep=0, outer sep=0] (n10) {}
        node[pos=0.5000, inner sep=0, outer sep=0] (n11) {}
        node[pos=0.6000, inner sep=0, outer sep=0] (n12) {}
        node[pos=0.6500, inner sep=0, outer sep=0] (n13) {}
        node[pos=0.7000, inner sep=0, outer sep=0] (n14) {}
        node[pos=0.7500, inner sep=0, outer sep=0] (n15) {}
        node[pos=0.8000, inner sep=0, outer sep=0 ] (n16) {}
        node[pos=0.8500, inner sep=0, outer sep=0 ] (n17) {}
        node[pos=0.9000, inner sep=0, outer sep=0 ] (n18) {}
        node[pos=0.9500, inner sep=0, outer sep=0 ] (n19) {}
        node[pos=1.0000, inner sep=0, outer sep=0] (n20) {}
        ;

        \draw (n1) -- (n2) -- (n3) -- (n4) -- (n5);
        \draw (n6) -- (n7) -- (n8) -- (n9) -- (n10);
        \draw (n11) -- (n12) -- (n13) -- (n14) -- (n15);
        \draw (n16) -- (n17) -- (n18) -- (n19) -- (n20);

        \draw (n5) -- (n10) -- (n15) -- (n20) -- (n5);
    \end{axis}
\end{tikzpicture}}
        \label{subfig:toy4_embedding3d}
    }
    \caption{\footnotesize (a) Toy graph composed by 4 path graphs connected each other with a cycle graph passing through nodes $\lbrace 5,10,15,20 \rbrace$; (b) Spectrum of the first 9 eigenvalues of the Laplacian applied to graph in Fig.~\ref{subfig:toy4_data}; (c) 2D embedding space reconstructed using components $\lbrace \mathbf{u}^1,\mathbf{u}^2 \rbrace$; (d) 3D embedding space reconstructed using components $\lbrace \mathbf{u}^1,\mathbf{u}^2,\mathbf{u}^3 \rbrace$.}
    \label{fig:ex_toy4}
\end{figure}
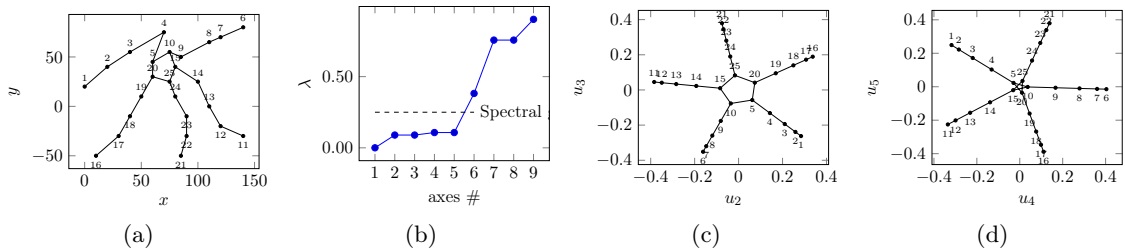
\begin{figure}[ht]
    \subfloat[]{
        \scalebox{.6}{\begin{tikzpicture}
    \begin{axis}[
            xlabel=$x$,
            ylabel=$y$,
            height = 5cm,
            scatter/classes={%
                    1={mark=*,black,fill=red,scale=0.7},%
                    2={mark=*,black,fill=cyan,scale=0.7}},
            every node near coord/.append style={font=\tiny},
        ]
        \addplot[only marks,mark=*,nodes near coords={\labelz},
            visualization depends on={value \thisrowno{0}\as\labelz}, mark size=1] table [x=x, y=y]{\relativepath ../figures/data/toy_5traj.csv}
        node[pos=0,    inner sep=0, outer sep=0] (n1) {}
        node[pos=0.04, inner sep=0, outer sep=0] (n2) {}
        node[pos=0.08, inner sep=0, outer sep=0] (n3) {}
        node[pos=0.12, inner sep=0, outer sep=0] (n4) {}
        node[pos=0.16, inner sep=0, outer sep=0] (n5) {}
        node[pos=0.20, inner sep=0, outer sep=0] (n6) {}
        node[pos=0.24, inner sep=0, outer sep=0] (n7) {}
        node[pos=0.28, inner sep=0, outer sep=0] (n8) {}
        node[pos=0.32, inner sep=0, outer sep=0] (n9) {}
        node[pos=0.36, inner sep=0, outer sep=0] (n10) {}
        node[pos=0.40, inner sep=0, outer sep=0] (n11) {}
        node[pos=0.44, inner sep=0, outer sep=0] (n12) {}
        node[pos=0.48, inner sep=0, outer sep=0] (n13) {}
        node[pos=0.56, inner sep=0, outer sep=0] (n14) {}
        node[pos=0.60, inner sep=0, outer sep=0] (n15) {}
        node[pos=0.64, inner sep=0, outer sep=0 ] (n16) {}
        node[pos=0.68, inner sep=0, outer sep=0 ] (n17) {}
        node[pos=0.72, inner sep=0, outer sep=0 ] (n18) {}
        node[pos=0.76, inner sep=0, outer sep=0 ] (n19) {}
        node[pos=0.80, inner sep=0, outer sep=0] (n20) {}
        node[pos=0.84, inner sep=0, outer sep=0] (n21) {}
        node[pos=0.88, inner sep=0, outer sep=0] (n22) {}
        node[pos=0.92, inner sep=0, outer sep=0] (n23) {}
        node[pos=0.96, inner sep=0, outer sep=0] (n24) {}
        node[pos=1.00, inner sep=0, outer sep=0] (n25) {}
        ;

        \draw (n1) -- (n2) -- (n3) -- (n4) -- (n5);
        \draw (n6) -- (n7) -- (n8) -- (n9) -- (n10);
        \draw (n11) -- (n12) -- (n13) -- (n14) -- (n15);
        \draw (n16) -- (n17) -- (n18) -- (n19) -- (n20);
        \draw (n21) -- (n22) -- (n23) -- (n24) -- (n25);

        \draw (n5) -- (n10) -- (n15) -- (n25) -- (n20) -- (n5);
    \end{axis}
\end{tikzpicture}}
        \label{subfig:toy5_data}
    }
    \subfloat[]{
        \scalebox{.6}{\begin{tikzpicture}
    \begin{axis}[
            xlabel={axes $\#$},
            ylabel={$\lambda$},
            xtick = {1,...,9},
            y tick label style={
                    /pgf/number format/.cd,
                    fixed,
                    fixed zerofill,
                    precision=2,
                    /tikz/.cd
                },
            height = 5cm
        ]
        \addplot table [x=label, y=lambda, select coords between index={0}{8}]{\relativepath ../figures/data/toy_5traj.csv};
        \draw[dashed] (axis cs:1,0.25) -- (axis cs:6,0.25) node[right]{Spectral gap};
    \end{axis}
\end{tikzpicture}}
        \label{subfig:toy5_eigval}
    }
    \subfloat[]{
        \scalebox{.6}{\begin{tikzpicture}
    \begin{axis}[
            xlabel=$u_2$,
            ylabel=$u_3$,
            height = 5cm,
            scatter/classes={%
                    1={mark=*,black,fill=red,scale=0.7},%
                    2={mark=*,black,fill=cyan,scale=0.7}},
            every node near coord/.append style={font=\tiny},
        ]
        \addplot[only marks,mark=*,nodes near coords={\labelz},
            visualization depends on={value \thisrowno{0}\as\labelz}, mark size=1] table [x=e2, y=e3]{\relativepath ../figures/data/toy_5traj.csv}
        node[pos=0,    inner sep=0, outer sep=0] (n1) {}
        node[pos=0.04, inner sep=0, outer sep=0] (n2) {}
        node[pos=0.08, inner sep=0, outer sep=0] (n3) {}
        node[pos=0.12, inner sep=0, outer sep=0] (n4) {}
        node[pos=0.16, inner sep=0, outer sep=0] (n5) {}
        node[pos=0.20, inner sep=0, outer sep=0] (n6) {}
        node[pos=0.24, inner sep=0, outer sep=0] (n7) {}
        node[pos=0.28, inner sep=0, outer sep=0] (n8) {}
        node[pos=0.32, inner sep=0, outer sep=0] (n9) {}
        node[pos=0.36, inner sep=0, outer sep=0] (n10) {}
        node[pos=0.40, inner sep=0, outer sep=0] (n11) {}
        node[pos=0.44, inner sep=0, outer sep=0] (n12) {}
        node[pos=0.48, inner sep=0, outer sep=0] (n13) {}
        node[pos=0.56, inner sep=0, outer sep=0] (n14) {}
        node[pos=0.60, inner sep=0, outer sep=0] (n15) {}
        node[pos=0.64, inner sep=0, outer sep=0 ] (n16) {}
        node[pos=0.68, inner sep=0, outer sep=0 ] (n17) {}
        node[pos=0.72, inner sep=0, outer sep=0 ] (n18) {}
        node[pos=0.76, inner sep=0, outer sep=0 ] (n19) {}
        node[pos=0.80, inner sep=0, outer sep=0] (n20) {}
        node[pos=0.84, inner sep=0, outer sep=0] (n21) {}
        node[pos=0.88, inner sep=0, outer sep=0] (n22) {}
        node[pos=0.92, inner sep=0, outer sep=0] (n23) {}
        node[pos=0.96, inner sep=0, outer sep=0] (n24) {}
        node[pos=1.00, inner sep=0, outer sep=0] (n25) {}
        ;

        \draw (n1) -- (n2) -- (n3) -- (n4) -- (n5);
        \draw (n6) -- (n7) -- (n8) -- (n9) -- (n10);
        \draw (n11) -- (n12) -- (n13) -- (n14) -- (n15);
        \draw (n16) -- (n17) -- (n18) -- (n19) -- (n20);
        \draw (n21) -- (n22) -- (n23) -- (n24) -- (n25);

        \draw (n5) -- (n10) -- (n15) -- (n25) -- (n20) -- (n5);
    \end{axis}
\end{tikzpicture}}
        \label{subfig:toy5_embedding2d_1}
    }
    \subfloat[]{
        \scalebox{.6}{\begin{tikzpicture}
    \begin{axis}[
            xlabel=$u_4$,
            ylabel=$u_5$,
            height = 5cm,
            scatter/classes={%
                    1={mark=*,black,fill=red,scale=0.7},%
                    2={mark=*,black,fill=cyan,scale=0.7}},
            every node near coord/.append style={font=\tiny},
        ]
        \addplot[only marks,mark=*,nodes near coords={\labelz},
            visualization depends on={value \thisrowno{0}\as\labelz}, mark size=1] table [x=e4, y=e5]{\relativepath ../figures/data/toy_5traj.csv}
        node[pos=0,    inner sep=0, outer sep=0] (n1) {}
        node[pos=0.04, inner sep=0, outer sep=0] (n2) {}
        node[pos=0.08, inner sep=0, outer sep=0] (n3) {}
        node[pos=0.12, inner sep=0, outer sep=0] (n4) {}
        node[pos=0.16, inner sep=0, outer sep=0] (n5) {}
        node[pos=0.20, inner sep=0, outer sep=0] (n6) {}
        node[pos=0.24, inner sep=0, outer sep=0] (n7) {}
        node[pos=0.28, inner sep=0, outer sep=0] (n8) {}
        node[pos=0.32, inner sep=0, outer sep=0] (n9) {}
        node[pos=0.36, inner sep=0, outer sep=0] (n10) {}
        node[pos=0.40, inner sep=0, outer sep=0] (n11) {}
        node[pos=0.44, inner sep=0, outer sep=0] (n12) {}
        node[pos=0.48, inner sep=0, outer sep=0] (n13) {}
        node[pos=0.56, inner sep=0, outer sep=0] (n14) {}
        node[pos=0.60, inner sep=0, outer sep=0] (n15) {}
        node[pos=0.64, inner sep=0, outer sep=0 ] (n16) {}
        node[pos=0.68, inner sep=0, outer sep=0 ] (n17) {}
        node[pos=0.72, inner sep=0, outer sep=0 ] (n18) {}
        node[pos=0.76, inner sep=0, outer sep=0 ] (n19) {}
        node[pos=0.80, inner sep=0, outer sep=0] (n20) {}
        node[pos=0.84, inner sep=0, outer sep=0] (n21) {}
        node[pos=0.88, inner sep=0, outer sep=0] (n22) {}
        node[pos=0.92, inner sep=0, outer sep=0] (n23) {}
        node[pos=0.96, inner sep=0, outer sep=0] (n24) {}
        node[pos=1.00, inner sep=0, outer sep=0] (n25) {}
        ;

        \draw (n1) -- (n2) -- (n3) -- (n4) -- (n5);
        \draw (n6) -- (n7) -- (n8) -- (n9) -- (n10);
        \draw (n11) -- (n12) -- (n13) -- (n14) -- (n15);
        \draw (n16) -- (n17) -- (n18) -- (n19) -- (n20);
        \draw (n21) -- (n22) -- (n23) -- (n24) -- (n25);

        \draw (n5) -- (n10) -- (n15) -- (n25) -- (n20) -- (n5);
    \end{axis}
\end{tikzpicture}}
        \label{subfig:toy5_embedding2d_2}
    }
    \caption{\footnotesize (a) Toy graph composed by 5 path graphs connected each other with a cyclic graph passing through nodes $\lbrace 5,10,15,20,25 \rbrace$; (b) Spectrum of the first 9 eigenvalues of the Laplacian applied to graph in Fig.~\ref{subfig:toy5_data}; (c) 2D embedding space reconstructed using components $\lbrace \mathbf{u}^1,\mathbf{u}^2 \rbrace$; (d) 2D embedding space reconstructed using components $\lbrace \mathbf{u}^3,\mathbf{u}^4 \rbrace$.}
    \label{fig:ex_toy5}
\end{figure}

We illustrate this property in two examples using 4-path and 5-path graphs, respectively, see Fig.~\ref{fig:ex_toy4} and \ref{fig:ex_toy5}. As in our previous example with a 3-path graph, the path graph are connected through one cycle graph. For the 4-path graph, we see that the spectrum of the Laplacian entails two eigenvalues of equal magnitude. The third eigenvalue has magnitude comparable but larger than the previous two, Fig.~\ref{subfig:toy4_eigval}. The graph can be linearized using the two eigenvectors with equal eigenvalues, see Fig.~\ref{subfig:toy4_embedding2d}. Quasi-linearization in the $3D$ embedding space can be achieved by using as coordinates the first three components, see Fig.~\ref{subfig:toy4_embedding3d}.

Using the 5-path graphs, we obtain two pairs of eigenvalues with equal magnitude, visible in the spectrum, Fig.~\ref{subfig:toy5_eigval}. We can hence reconstruct two distinct $2D$ embedding spaces where the DS is perfectly linear. While the embedding space constructed using eigenvectors $\lbrace \mathbf{u}^2, \mathbf{u}^3 \rbrace$ preserves the radial ordering of the path graphs, Fig.~\ref{subfig:toy5_embedding2d_1}, the embedding constructed with components $\lbrace \mathbf{u}^4, \mathbf{u}^5 \rbrace$ does not, as shown Fig.~\ref{subfig:toy5_embedding2d_2}.

\section{Generating the Graph with Real Data}
\label{sec:graph_generate}
To exploit the results stated in the previous section, we must first embed the data in a graph. To achieve this, we propose a kernel that takes advantage of positions and velocities.

\subsection{Velocity-Augmented Kernel}
The kernel provides a distance measure across nodes $\nu_i, \nu_j$:
\begin{equation}
    \scalebox{.9}{$
            k(\nu_i,\nu_j) = \text{exp}{\left( -\frac{1}{2\sigma^2} \left( \underbrace{\norm{\mathbf{x}_i-\mathbf{x}_j}^2}_\text{locality} +  \overbrace{\gamma \left( \abs{g(\mathbf{x}_i, \mathbf{x}_j, \dot{\mathbf{x}}_i)} \right)^2 + \gamma \left( \abs{g(\mathbf{x}_j, \mathbf{x}_i, \dot{\mathbf{x}}_j)} \right)^2}^\text{directionality} \right) \right)}
        $},
    \label{eqn:dir_kernel}
\end{equation}
where \(\dot{\mathbf{x}}_i = \mathbf{f}(\mathbf{x}_i)\) is the DS map that relates positions to velocities.

We introduce $g$ a function to measure the angular distance across the velocity vectors for a pair of data points $\mathbf{x}_i$ and $\mathbf{x}_j$. We relate the distance vector $\mathbf{x}_j - \mathbf{x}_i$ and the velocity vector $\dot{\mathbf{x}}_i$ through a cosine kernel
\begin{equation}
    \scalebox{.9}{$
            g(\mathbf{x}_i, \mathbf{x}_j, \dot{\mathbf{x}}_i) = \underbrace{\delta_{\sigma_f} \left(\norm{\dot{\mathbf{x}}_i}\norm{\dot{\mathbf{x}}_j}\right)}_\text{Filter} + \underbrace{\dprod{\frac{\mathbf{x}_j - \mathbf{x}_i}{\norm{\mathbf{x}_j - \mathbf{x}_i}},\frac{\dot{\mathbf{x}}_i}{\norm{\dot{\mathbf{x}}_i}}}}_\text{Cosine similarity kernel}
        $},
    \label{eqn:cos_kernel}
\end{equation}
where $\delta_{\sigma_f}$ is a normal Gaussian distribution \( \delta_{\sigma_f}(x)=\exp \left( -\frac{x^2}{2\sigma_f^2} \right)\) centered in \( \norm*{\boldsymbol{\dot{\mathbf{x}}}}=0 \), with $\sigma_f$ set approximately to $0$. The presence of the filtering part in Eq.~\ref{eqn:cos_kernel} is necessary to withhold the "directionality" penalization among zero velocity points potentially close to an attractor. The filter $\delta_{\sigma_f}$ takes care of this situation outputting $1$ whenever the velocity is zero. The parameter $\sigma_f$ can be regulated to take into account known noise, e.g., when one knows that the velocity at an attractor point is not zero for numerical reasons.

The linear function $\gamma : \mathbb{R} \rightarrow \mathbb{R}$ in Eq.~\ref{eqn:dir_kernel} is defined as
\begin{equation}
    \gamma(g) = \frac{3}{2}\theta_r\left( 1 - g\right)\sigma.
    \label{eqn:gamma_fun}
\end{equation}
The co-domain the function $g$, in Eq.~\ref{eqn:cos_kernel}, is in between $-1$ and $1$. The function $\gamma$, in Eq.~\ref{eqn:gamma_fun}, maps the part of the co-domain of $g$ in between $cos(\theta_{r})$ and $1$ to the range between $3 \sigma$ and $0$. Consequently, the range of values of $g$ in between $-1$ and $cos(\theta_{r})$ will be mapped to values greater than $3 \sigma$. When the angle between two velocity vectors is $\theta_{r}$ the linear map yields a value equal to $3\sigma$ causing a penalization of the weight about $99\%$. When the output of the cosine is proximal to $1$, due to the almost complete co-linearity, the linear function will yield $0$ causing no penalization over the weight produced by the standard RBF (Radial Basis Functions) kernel. The reference angle, $\theta_{r}$, is a parameter that can be set depending on the sampling frequency. High frequencies allow for a more strict ($\theta_{r} \approx 0$) selection of such angle and vice-versa.

\begin{figure}[ht]
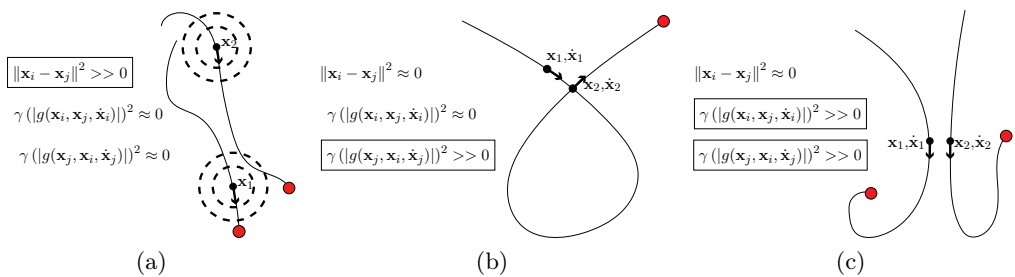

    \centering
    \subfloat[][]{
        \scalebox{0.55}{\subimport{../figures/}{kernel_loc}}
        \label{subfig:kernel_loc}
    }
    \subfloat[][]{
        \scalebox{0.55}{\subimport{../figures/}{kernel_cont}}
        \label{subfig:kernel_cont}
    }
    \subfloat[][]{
        \scalebox{0.55}{\subimport{../figures/}{kernel_dir}}
        \label{subfig:kernel_dir}
    }
    \caption{\footnotesize Illustration of the effect of the spatial distribution of points and velocity vectors on the kernel Eq.~\ref{eqn:dir_kernel}: (a) the first term in the exponential generates low values for points that are far apart; (b) the third term in the exponential generates low values when consecutive velocity vectors are not aligned; (c) the second and third term in the exponential generate low values whenever the distance vector connecting two points is not aligned with the velocity vector of one of them.}
    \label{fig:kernel}
\end{figure}

In order to understand the kernel better, we illustrate the effect of each term in three scenarios depicted in Fig~\ref{fig:kernel}. The directed kernel encompasses two components: "locality" and "directionality".
The locality term gives a measure of the spatial distance between pairs of points based on the Euclidean distance, similar to the standard RBF kernel. The two directionality terms measure the co-linearity of the velocity vectors. Two points far apart with co-linear velocity vectors or two points close with distinct velocity vector will have a small value on the kernel and hence a loose connectivity in the similarity matrix used to generate the graph. Sole points that are both close and with co-directed velocity will reach the maximum kernel value, 1.

Points $\mathbf{x}_1$ and $\mathbf{x}_2$ in Fig.~\ref{subfig:kernel_loc} will have negligible connection weights due to the large distance between them even though the "directionality" terms would yield values closed to $1$ (about $0$ after mapping $\gamma$) since $\mathbf{x}_2 - \mathbf{x}_1 \parallel \dot{\mathbf{x}}_1$ and $\mathbf{x}_1 - \mathbf{x}_2 \parallel \dot{\mathbf{x}}_2$. The second example in Fig.~\ref{subfig:kernel_cont} considers the possibility of having intersecting trajectories perhaps due to second order dynamics or imprecise user demonstration. Since the figure might be misleading we clarify that point $\mathbf{x}_1$ belongs to the beginning of the trajectory while point $\mathbf{x}_2$ lies at the end, close to the attractor. In this situation we would like to "separate" the two points since they are not close from a dynamical perspective. Due to the proximity of the points, the "Locality" part contributes to generate a strong connection between the two points. The second component of the "directionality" part will take care of drastically decreasing the weight connection. Indeed, if the first term yields a high value, $\mathbf{x}_2 - \mathbf{x}_1 \parallel \dot{\mathbf{x}}_1$, the second term will be decisive in cutting of the connection between the points since $\mathbf{x}_1 - \mathbf{x}_2 \approx \dot{\mathbf{x}}_1 \perp \dot{\mathbf{x}}_2$. As last scenario in Fig.~\ref{subfig:kernel_dir}, we consider the case of two proximal trajectories belonging to different sub-dynamics. As in the previous example the "Locality" cannot account for this situation; the "directionality" part will take care of decreasing the weight connection given that $\mathbf{x}_2 - \mathbf{x}_1 \perp \dot{\mathbf{x}}_1$ and $\mathbf{x}_1 - \mathbf{x}_2 \perp \dot{\mathbf{x}}_2$.

\begin{figure}[ht]
    \centering
    \subfloat[][]{
        \includegraphics[scale=0.35]{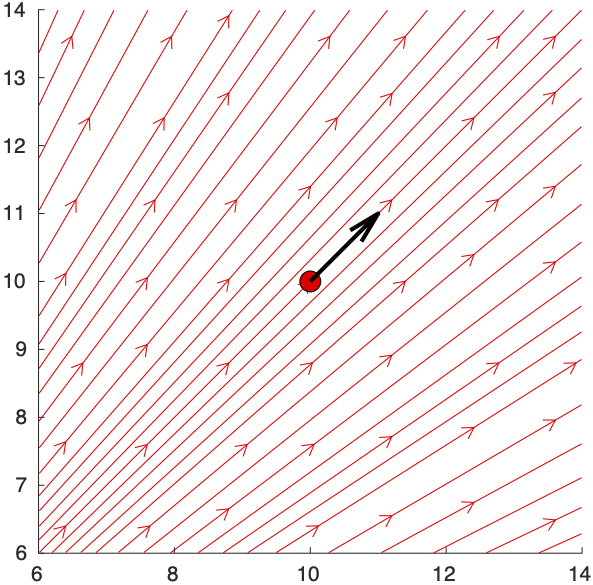}
        \label{subfig:reference_field}
    }
    \qquad
    \subfloat[][]{
        \includegraphics[scale=0.35]{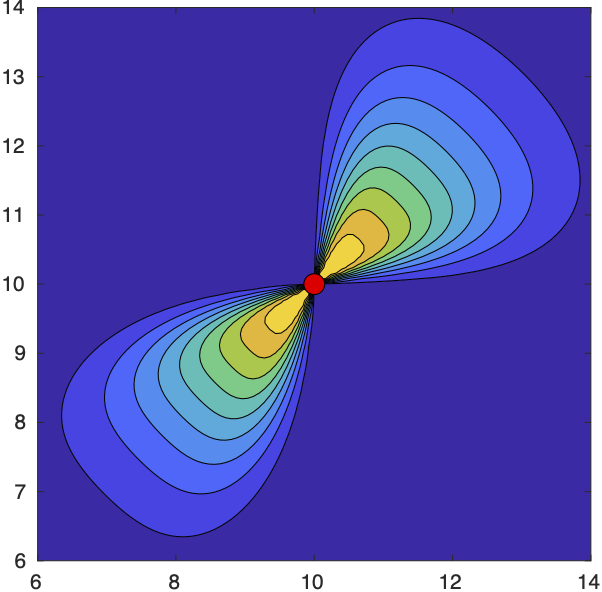}
        \label{subfig:kernel_eval}
    }
    \caption{\footnotesize (a) Background vector field and reference evaluation point/velocity; (b) Contour of the velocity-augmented kernel.}
    \label{fig:kernel_plot}
\end{figure}

Fig.~\ref{subfig:reference_field} shows streamlines of an expanding vector field $\dot{\mathbf{x}} = \mathbf{x}$ and a reference point with its velocity vector.
In Fig.~\ref{subfig:kernel_eval} it is possible to see how the proposed kernel generates, with respect to the considered vector field, high value regions symmetrically placed with respect to the plane orthogonal to reference velocity in the proximity of the evaluation point.

\subsection{Selecting the Hyperparameters}
The choice of the hyperparameter $\sigma$ is important as it modulates the granularity of the distance measure in Cartesian space. To inform the choice of $\sigma$, we can use  the sampling frequency $f_{sampling}$ of the acquisition system, when recording trajectories. The higher the frequency, the closer the two consecutive data points.
The maximum distance between two consecutive point is $d_{max} = \frac{\text{argmax}(\mathcal{D})}{f_{sampling}} \quad \text{with} \quad \mathcal{D}:=\lbrace {\dot{\mathbf{x}}}_i, i=1,\dots,m \rbrace$. If sampling is perfect (no frame loss), we can set \( \sigma = d_{max} \). Otherwise, a margin of error may be warranted.

The adjacency matrix is computed as follows: $A_{ij} = 1$ if $k(\nu_i,\nu_j) \ge \epsilon$ otherwise $A_{ij} = 0$. The tolerance $\epsilon$ must be chosen in relation to $\sigma$. For instance, if $\sigma=d_{max}$, the kernel will be equal to at maximum $0.6$ for the two most distant pair of points.

\section{Clustering Dynamics \& Attractor Search}
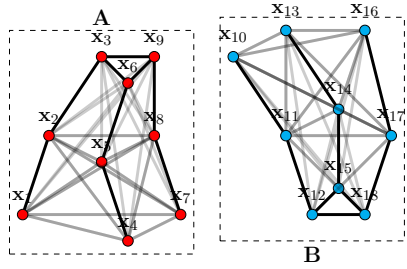
\begin{figure}[ht]
    \centering
    \scalebox{0.7}{\begin{tikzpicture}[
                every node/.style={draw,
                                circle,
                                black,
                                inner sep=0,
                                outer sep=0,
                                minimum size=2mm,
                        }
        ]
        \node[label={$\mathbf{x}_1$},fill=red] (x1) at (0.5,1) {};
        \node[label={$\mathbf{x}_2$},fill=red] (x2) at (1,2.5) {};
        \node[label={$\mathbf{x}_3$},fill=red] (x3) at (2,4) {};
        \node[label={$\mathbf{x}_4$},fill=red] (x4) at (2.5,0.5) {};
        \node[label={$\mathbf{x}_5$},fill=red] (x5) at (2,2) {};
        \node[label={$\mathbf{x}_6$},fill=red] (x6) at (2.5,3.5) {};
        \node[label={$\mathbf{x}_7$},fill=red] (x7) at (3.5,1) {};
        \node[label={$\mathbf{x}_8$},fill=red] (x8) at (3,2.5) {};
        \node[label={$\mathbf{x}_9$},fill=red] (x9) at (3,4) {};
        \node[label={$\mathbf{x}_{10}$},fill=cyan] (x10) at (4.5,4) {};
        \node[label={$\mathbf{x}_{11}$},fill=cyan] (x11) at (5.5,2.5) {};
        \node[label={$\mathbf{x}_{12}$},fill=cyan] (x12) at (6,1) {};
        \node[label={$\mathbf{x}_{13}$},fill=cyan] (x13) at (5.5,4.5) {};
        \node[label={$\mathbf{x}_{14}$},fill=cyan] (x14) at (6.5,3) {};
        \node[label={$\mathbf{x}_{15}$},fill=cyan] (x15) at (6.5,1.5) {};
        \node[label={$\mathbf{x}_{16}$},fill=cyan] (x16) at (7,4.5) {};
        \node[label={$\mathbf{x}_{17}$},fill=cyan] (x17) at (7.5,2.5) {};
        \node[label={$\mathbf{x}_{18}$},fill=cyan] (x18) at (7,1) {};
        \draw[dashed] (0.25,0.25) rectangle (3.75,4.5);
        \draw[dashed] (4.25,0.5) rectangle (7.75,4.75);
        \node[draw=none, fill=none] at (2,4.75) {\large $\mathbf{A}$};
        \node[draw=none, fill=none] at (6,0.25) {\large $\mathbf{B}$};
        \draw[ultra thick] (x1) -- (x2);
        \draw[ultra thick] (x2) -- (x3);
        \draw[ultra thick] (x4) -- (x5);
        \draw[ultra thick] (x5) -- (x6);
        \draw[ultra thick] (x7) -- (x8);
        \draw[ultra thick] (x8) -- (x9);
        \draw[ultra thick] (x10) -- (x11);
        \draw[ultra thick] (x11) -- (x12);
        \draw[ultra thick] (x13) -- (x14);
        \draw[ultra thick] (x14) -- (x15);
        \draw[ultra thick] (x16) -- (x17);
        \draw[ultra thick] (x17) -- (x18);
        \draw[ultra thick] (x3) -- (x6);
        \draw[ultra thick] (x6) -- (x9);
        \draw[ultra thick] (x3) -- (x9);
        \draw[ultra thick] (x12) -- (x15);
        \draw[ultra thick] (x12) -- (x18);
        \draw[ultra thick] (x15) -- (x18);
        \draw[ultra thick,opacity=0.2] (x1) -- (x4);
        \draw[ultra thick,opacity=0.2] (x1) -- (x5);
        \draw[ultra thick,opacity=0.2] (x1) -- (x6);
        \draw[ultra thick,opacity=0.2] (x1) -- (x7);
        \draw[ultra thick,opacity=0.2] (x1) -- (x8);
        \draw[ultra thick,opacity=0.2] (x1) -- (x9);
        \draw[ultra thick,opacity=0.2] (x2) -- (x4);
        \draw[ultra thick,opacity=0.2] (x2) -- (x5);
        \draw[ultra thick,opacity=0.2] (x2) -- (x6);
        \draw[ultra thick,opacity=0.2] (x2) -- (x7);
        \draw[ultra thick,opacity=0.2] (x2) -- (x8);
        \draw[ultra thick,opacity=0.2] (x2) -- (x9);
        \draw[ultra thick,opacity=0.2] (x4) -- (x1);
        \draw[ultra thick,opacity=0.2] (x4) -- (x2);
        \draw[ultra thick,opacity=0.2] (x4) -- (x3);
        \draw[ultra thick,opacity=0.2] (x4) -- (x7);
        \draw[ultra thick,opacity=0.2] (x4) -- (x8);
        \draw[ultra thick,opacity=0.2] (x4) -- (x9);
        \draw[ultra thick,opacity=0.2] (x5) -- (x1);
        \draw[ultra thick,opacity=0.2] (x5) -- (x2);
        \draw[ultra thick,opacity=0.2] (x5) -- (x3);
        \draw[ultra thick,opacity=0.2] (x5) -- (x7);
        \draw[ultra thick,opacity=0.2] (x5) -- (x8);
        \draw[ultra thick,opacity=0.2] (x5) -- (x9);
        \draw[ultra thick,opacity=0.2] (x7) -- (x1);
        \draw[ultra thick,opacity=0.2] (x7) -- (x2);
        \draw[ultra thick,opacity=0.2] (x7) -- (x3);
        \draw[ultra thick,opacity=0.2] (x7) -- (x4);
        \draw[ultra thick,opacity=0.2] (x7) -- (x5);
        \draw[ultra thick,opacity=0.2] (x7) -- (x6);
        \draw[ultra thick,opacity=0.2] (x8) -- (x1);
        \draw[ultra thick,opacity=0.2] (x8) -- (x2);
        \draw[ultra thick,opacity=0.2] (x8) -- (x3);
        \draw[ultra thick,opacity=0.2] (x8) -- (x4);
        \draw[ultra thick,opacity=0.2] (x8) -- (x5);
        \draw[ultra thick,opacity=0.2] (x8) -- (x6);
        \draw[ultra thick,opacity=0.2] (x10) -- (x13);
        \draw[ultra thick,opacity=0.2] (x10) -- (x14);
        \draw[ultra thick,opacity=0.2] (x10) -- (x15);
        \draw[ultra thick,opacity=0.2] (x10) -- (x16);
        \draw[ultra thick,opacity=0.2] (x10) -- (x17);
        \draw[ultra thick,opacity=0.2] (x10) -- (x18);
        \draw[ultra thick,opacity=0.2] (x11) -- (x13);
        \draw[ultra thick,opacity=0.2] (x11) -- (x14);
        \draw[ultra thick,opacity=0.2] (x11) -- (x15);
        \draw[ultra thick,opacity=0.2] (x11) -- (x16);
        \draw[ultra thick,opacity=0.2] (x11) -- (x17);
        \draw[ultra thick,opacity=0.2] (x11) -- (x18);
        \draw[ultra thick,opacity=0.2] (x13) -- (x10);
        \draw[ultra thick,opacity=0.2] (x13) -- (x11);
        \draw[ultra thick,opacity=0.2] (x13) -- (x12);
        \draw[ultra thick,opacity=0.2] (x13) -- (x16);
        \draw[ultra thick,opacity=0.2] (x13) -- (x17);
        \draw[ultra thick,opacity=0.2] (x13) -- (x18);
        \draw[ultra thick,opacity=0.2] (x14) -- (x10);
        \draw[ultra thick,opacity=0.2] (x14) -- (x11);
        \draw[ultra thick,opacity=0.2] (x14) -- (x12);
        \draw[ultra thick,opacity=0.2] (x14) -- (x16);
        \draw[ultra thick,opacity=0.2] (x14) -- (x17);
        \draw[ultra thick,opacity=0.2] (x14) -- (x18);
        \draw[ultra thick,opacity=0.2] (x16) -- (x13);
        \draw[ultra thick,opacity=0.2] (x16) -- (x14);
        \draw[ultra thick,opacity=0.2] (x16) -- (x15);
        \draw[ultra thick,opacity=0.2] (x16) -- (x10);
        \draw[ultra thick,opacity=0.2] (x16) -- (x11);
        \draw[ultra thick,opacity=0.2] (x16) -- (x12);
        \draw[ultra thick,opacity=0.2] (x17) -- (x13);
        \draw[ultra thick,opacity=0.2] (x17) -- (x14);
        \draw[ultra thick,opacity=0.2] (x17) -- (x15);
        \draw[ultra thick,opacity=0.2] (x17) -- (x10);
        \draw[ultra thick,opacity=0.2] (x17) -- (x11);
        \draw[ultra thick,opacity=0.2] (x17) -- (x12);
\end{tikzpicture}}
    \caption{\footnotesize Toy data set of 2-attractor samples DS.}
    \label{fig:toy_graph}
\end{figure}

We consider as toy example a 2-attractor DS where each sub-dynamics is constituted by $3$ sampled trajectories. Fig.~\ref{fig:toy_graph} shows the connection strength generated by the velocity-augmented kernel in Eq.~\ref{eqn:dir_kernel}. Connection between nodes belonging to different trajectories will have very low weight. Only points at the end of each trajectory, close to an attractor, will have high weight even if they belong to different trajectories.

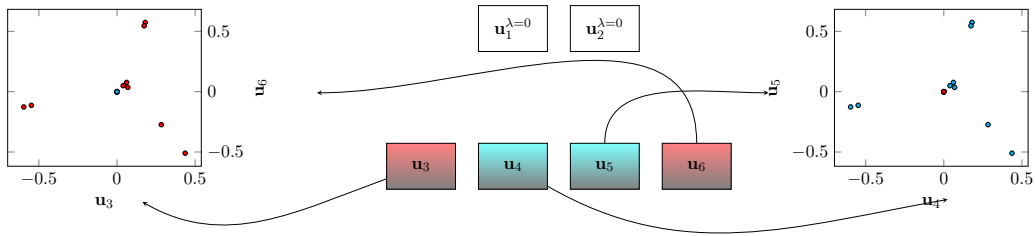
\begin{figure}[h!]
    \centering
    \resizebox{.9\textwidth}{!}{\begin{tikzpicture}[
        every node/.style={
                cyan/.style={bottom color=white!50!black, top color=cyan!50},
                red/.style={bottom color=white!50!black, top color=red!50}
            }
    ]

    \begin{scope}[shift={(-9,-3)}]
        \begin{axis}[
                xlabel=$\mathbf{u}_3$,
                ylabel=$\mathbf{u}_6$,
                height = 5cm,
                scatter/classes={%
                        1={mark=*,black,fill=red,scale=0.7},%
                        2={mark=*,black,fill=cyan,scale=0.7}},
                ylabel near ticks, yticklabel pos=right
            ]
            \addplot[scatter, only marks, scatter src=explicit symbolic] table [x=e3, y=e5, meta=labels]{\relativepath ../figures/data/toy_eigvec.dat};
        \end{axis}
    \end{scope}

    \begin{scope}
        \node[draw,minimum height=1cm,minimum width=1.5cm] at (2,0) {$\mathbf{u}_1^{\lambda=0}$};
        \node[draw,minimum height=1cm,minimum width=1.5cm] at (4,0) {$\mathbf{u}_2^{\lambda=0}$};

        \node[draw,minimum height=1cm,minimum width=1.5cm,red] at (0,-3) (u3) {$\mathbf{u}_3$};
        \node[draw,minimum height=1cm,minimum width=1.5cm,cyan] at (2,-3) (u4) {$\mathbf{u}_4$};
        \node[draw,minimum height=1cm,minimum width=1.5cm,cyan] at (4,-3) (u5) {$\mathbf{u}_5$};
        \node[draw,minimum height=1cm,minimum width=1.5cm,red] at (6,-3) (u6) {$\mathbf{u}_6$};
    \end{scope}

    \begin{scope}[shift={(9,-3)}]
        \begin{axis}[
                xlabel=$\mathbf{u}_4$,
                ylabel=$\mathbf{u}_5$,
                height = 5cm,
                scatter/classes={%
                        1={mark=*,black,fill=red,scale=0.7},%
                        2={mark=*,black,fill=cyan,scale=0.7}}
            ]
            \addplot[scatter, only marks, scatter src=explicit symbolic] table [x=e4, y=e6, meta=labels]{\relativepath ../figures/data/toy_eigvec.dat};
        \end{axis}
    \end{scope}

    \node[] at (-6.2,-3.7) (a) {};
    \node[] at (-2.4,-1.4) (b) {};

    \node[] at (11.6,-3.7) (c) {};
    \node[] at (7.7,-1.4) (d) {};

    \draw[-stealth] (u3) to[out=-160,in=-30] (a);
    \draw[-stealth] (u6) to[out=90,in=0] (b);
    \draw[-stealth] (u4) to[out=-30,in=-170] (c);
    \draw[-stealth] (u5) to[out=90,in=180] (d);
\end{tikzpicture}}
    \vspace{-5mm}
    \caption{\footnotesize (left) $\lbrace \mathbf{u}_3, \mathbf{u}_6 \rbrace$ embedding space; (center) Scatter matrix of the embedding space for the DS in \ref{fig:toy_graph}; (right) $\lbrace \mathbf{u}_4, \mathbf{u}_5 \rbrace$ embedding space.}
    \label{fig:eigvect_static}
\end{figure}

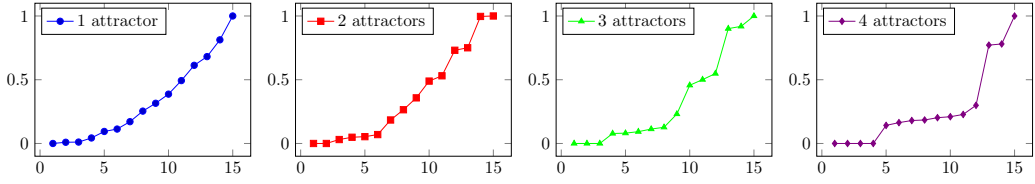
\begin{figure}[ht]
    \centering
    \resizebox{.9\textwidth}{!}{\begin{tikzpicture}
    \begin{groupplot}[group style={
                    group name=myplot,
                    group size= 4 by 2},height=5cm,width=6.4cm]
        \nextgroupplot[legend pos=north west]
        \addplot table [x=x, y=l1]{\relativepath ../figures/data/spectrum_attracts.dat};
        \addlegendentry{$1$ attractor}
        \nextgroupplot[legend pos=north west]
        \addplot[mark=square*, red] table [x=x, y=l2]{\relativepath ../figures/data/spectrum_attracts.dat};
        \addlegendentry{$2$ attractors}
        \nextgroupplot[legend pos=north west]
        \addplot[mark=triangle*, green] table [x=x, y=l3]{\relativepath ../figures/data/spectrum_attracts.dat};
        \addlegendentry{$3$ attractors}
        \nextgroupplot[legend pos=north west]
        \addplot[mark=diamond*, violet] table [x=x, y=l4]{\relativepath ../figures/data/spectrum_attracts.dat};
        \addlegendentry{$4$ attractors}
    \end{groupplot}
\end{tikzpicture}}
    \caption{\footnotesize Spectrum analysis of the multiple-attractor DS for three demonstrated trajectories and increasing number of attractors.}
    \label{fig:eigvec_analysis}
\end{figure}

The labeling of points, for clustering dynamics, is done by taking advantage of the first right eigenvectors, $\lbrace \mathbf{u}^1, \dots, \mathbf{u}^Q \rbrace$, connected to the eigenvalues equal to zero. Assuming $n$ data points for each of the $Q$ demonstrated dynamics the k-th eigenvectors extracted will have non-zero entries between $\mathbf{u}_{n(k-1)+1}$ and $\mathbf{u}_{nk}$, see Fig.~\ref{fig:toy_eigvec}. For each eigenvector the coefficients corresponding to the points belonging to a particular DS (sub-graph) are equal to a constant value $c$ while all the others are $0$.
\begin{figure}[ht]
    \centering
    \scalebox{0.75}{\begin{tikzpicture}
    \begin{scope}[shift={(0,0)}]
        \node[] at (.5,3.5) {\large{$\mathbf{u}^1$}};
        \draw[] (.25,0) -- (0,0) -- (0,3) -- (.25,3);
        \draw[] (1-.25,0) -- (1,0) -- (1,3) -- (1-.25,3);
        \node[] at (-1.5,2.85) (a) {$u = u_1^1$};
        \node[] at (0,2.85) (b) {};
        \draw[->] (a) -- (b);
        \node[] at (.5,2.85) {$c$};
        \node[] at (.5,2.5) {$c$};
        \node[draw, circle, outer sep=0, inner sep=0, minimum size=.5mm, fill] at (.5,2.2) {};
        \node[draw, circle, outer sep=0, inner sep=0, minimum size=.5mm, fill] at (.5,2.1) {};
        \node[draw, circle, outer sep=0, inner sep=0, minimum size=.5mm, fill] at (.5,2) {};
        \node[] at (-1.5,1.7) (c) {$u_n = u_n^1$};
        \node[] at (0,1.7) (d) {};
        \draw[->] (c) -- (d);
        \node[] at (.5,1.7) {c};
        \node[] at (.5,1.35) {0};
        \node[] at (.5,1) {0};
        \node[draw, circle, outer sep=0, inner sep=0, minimum size=.5mm, fill] at (.5,0.7) {};
        \node[draw, circle, outer sep=0, inner sep=0, minimum size=.5mm, fill] at (.5,0.6) {};
        \node[draw, circle, outer sep=0, inner sep=0, minimum size=.5mm, fill] at (.5,0.5) {};
        \node[] at (.5,0.2) {0};
    \end{scope}
    \begin{scope}[shift={(2,0)}]
        \node[] at (.5,3.5) {\large{$\mathbf{u}^2$}};
        \draw[] (.25,0) -- (0,0) -- (0,3) -- (.25,3);
        \draw[] (1-.25,0) -- (1,0) -- (1,3) -- (1-.25,3);
        \node[] at (.5,2.85) {0};
        \node[draw, circle, outer sep=0, inner sep=0, minimum size=.5mm, fill] at (.5,2.6) {};
        \node[draw, circle, outer sep=0, inner sep=0, minimum size=.5mm, fill] at (.5,2.5) {};
        \node[draw, circle, outer sep=0, inner sep=0, minimum size=.5mm, fill] at (.5,2.4) {};
        \node[] at (.5,2.15) {0};
        \node[] at (2.6,1.8) (a) {$u_1^2 = u_{n+1}$};
        \node[] at (1,1.8) (b) {};
        \draw[->] (a) -- (b);
        \node[] at (.5,1.8) {$c$};
        \node[draw, circle, outer sep=0, inner sep=0, minimum size=.5mm, fill] at (.5,1.55) {};
        \node[draw, circle, outer sep=0, inner sep=0, minimum size=.5mm, fill] at (.5,1.45) {};
        \node[draw, circle, outer sep=0, inner sep=0, minimum size=.5mm, fill] at (.5,1.35) {};
        \node[] at (2.5,1.1) (c) {$u_n^2 = u_{2n}$};
        \node[] at (1,1.1) (d) {};
        \draw[->] (c) -- (d);
        \node[] at (.5,1.1) {$c$};
        \node[] at (.5,0.8) {0};
        \node[draw, circle, outer sep=0, inner sep=0, minimum size=.5mm, fill] at (.5,0.6) {};
        \node[draw, circle, outer sep=0, inner sep=0, minimum size=.5mm, fill] at (.5,0.5) {};
        \node[draw, circle, outer sep=0, inner sep=0, minimum size=.5mm, fill] at (.5,0.4) {};
        \node[] at (.5,0.2) {0};
    \end{scope}
    \node[draw, circle, outer sep=0, inner sep=0, minimum size=1mm, fill] at (4,1.5) {};
    \node[draw, circle, outer sep=0, inner sep=0, minimum size=1mm, fill] at (4.5,1.5) {};
    \node[draw, circle, outer sep=0, inner sep=0, minimum size=1mm, fill] at (5,1.5) {};
    \begin{scope}[shift={(6,0)}]
        \node[] at (.5,3.5) {\large{$\mathbf{u}^Q$}};
        \draw[] (.25,0) -- (0,0) -- (0,3) -- (.25,3);
        \draw[] (1-.25,0) -- (1,0) -- (1,3) -- (1-.25,3);
        \node[] at (.5,2.85) {0};
        \node[] at (.5,2.5) {0};
        \node[draw, circle, outer sep=0, inner sep=0, minimum size=.5mm, fill] at (.5,2.2) {};
        \node[draw, circle, outer sep=0, inner sep=0, minimum size=.5mm, fill] at (.5,2.1) {};
        \node[draw, circle, outer sep=0, inner sep=0, minimum size=.5mm, fill] at (.5,2) {};
        \node[] at (.5,1.7) {0};
        \node[] at (3,1.35) (a) {$u_1^Q = u_{n(Q-1)+1}$};
        \node[] at (1,1.35) (b) {};
        \draw[->] (a) -- (b);
        \node[] at (.5,1.35) {$c$};
        \node[] at (.5,1) {$c$};
        \node[draw, circle, outer sep=0, inner sep=0, minimum size=.5mm, fill] at (.5,0.7) {};
        \node[draw, circle, outer sep=0, inner sep=0, minimum size=.5mm, fill] at (.5,0.6) {};
        \node[draw, circle, outer sep=0, inner sep=0, minimum size=.5mm, fill] at (.5,0.5) {};
        \node[] at (3,0.2) (c) {$u_n^Q= u_{nQ} = u_m$};
        \node[] at (1,0.2) (d) {};
        \draw[->] (c) -- (d);
        \node[] at (.5,0.2) {c};
    \end{scope}
\end{tikzpicture}}
    \caption{\footnotesize Structure of the constant right eigenvectors associated to eigenvalues equal to $0$.}
    \label{fig:toy_eigvec}
\end{figure}
In summary, the number of $0$ eigenvalues is used to determine the number of attractors present in the DS. The corresponding eigenvectors are then used to cluster the data points, belonging to each sub-dynamics. Plotting the first four eigenvectors, Fig.~\ref{fig:eigvect_static}, it is possible to observe that the spectral decomposition of the Laplacian produces a set of eigenvectors that are "expanding" one dynamics while "compressing", in zero, the other ones for each projected space.
In this example $\mathbf{u}^3$ and $\mathbf{u}^6$ focused on sub-dynamics \textbf{A} while $\mathbf{u}^4$ and $\mathbf{u}^5$ on sub-dynamics \textbf{B}. The selection of the "interesting" components can be easily achieved by checking the change of the slope in the spectrum of the Laplacian matrix. Fig.~\ref{fig:eigvec_analysis} gives the spectrum analysis for varying the number of attractors, keeping the number of demonstrated trajectories to three.
The number of relevant components grows proportionally to the number of sub-dynamics present in the dataset, \( n_{DS} = \sum_{q=1}^K K_q - 1 \) with $Q$ being the total number of sub-dynamics and $K_q$ the number of sampled trajectories for the $q$-th sub-dynamics. As shown in Alg.~\ref{alg:clustering}, taking advantage of the stationary right eigenvectors used to label points, each eigenvector is assigned to either one dynamics or the other.
\begin{center}
    \scalebox{0.6}{

\begin{minipage}{.9\textwidth}
    \begin{algorithm}[H]
        \DontPrintSemicolon

        \KwIn{$\Lambda = \lbrace \lambda_1, \dots, \lambda_m \rbrace$, $U = \lbrace \mathbf{u}_1, \dots, \mathbf{u}_m \rbrace$}
        \KwOut{$\mathbf{x}_q^*$, $Q$}

        $n_{\text{sub-DS}}$ = ($\lambda_k$ == 0)  \tcp*{Compute the number of attractors}
        $U_{clusters} = \lbrace \mathbf{u}_{\lambda=0}^1,\dots,\mathbf{u}_{\lambda=0}^{n_{\text{sub-DS}}} \rbrace$ \tcp*{Right "stationary" eigenvectors}
        \While{stop = false}
        {
        stop = false \;
        \tcc{From the first non-stationary eigenvector till the last one}
        \For{$i=n_{\text{attractors}} + 1$ {\bfseries to} $m$}
        {
        \tcc{Iterate over the number of attractors}
        \For{$k=1$ {\bfseries to} $n_{\text{attractors}}$}
        {
            \tcc{Assign eigenvector to sub-dynamics/attractor $k$}
            \If{$\mathbf{1}^T(u_{\lambda=0}^k \circ |u_i|) \neq 0$}
            {$U^k \leftarrow u_i$}
        }
        \tcc{Stop loop at the spectral gap}
        \If{$|\lambda_{i+1} - \lambda_i| > tol$}
        {stop = true}
        }
        }

        \caption{Clustering Dynamics \& Embedding Reconstruction.}
        \label{alg:clustering}
    \end{algorithm}
\end{minipage}

}
\end{center}
For the toy case in exam we have two eigenvectors associated to zero eigenvalues. The eigenvectors associated the least two eigenvalues, with algebraic multiplicity equal to two, are used for the reconstruction of the embedding spaces as shown in Fig.~\ref{fig:eigvect_static}.

The search of the attractor positions is carried out by exploiting the particular structure of the embedding spaces.
\begin{figure}[ht]
    \centering
    \subfloat[]{
        \includestandalone[scale=0.55]{../algorithms/search_attractor}
        \label{alg:attractor_search}
    }
    \quad
    \subfloat[]{
        \scalebox{0.85}{\subimport{../figures/}{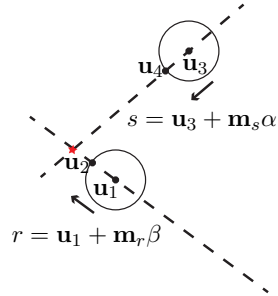}}
        \label{subfig:lines_intersect}
    }
    \caption{\footnotesize Finding the attractor in the embedding space.}
\end{figure}
In each embedding space the related dynamics is "linearized", while the other ones, as said before, are "compressed" in zero. In this space the position of the attractor, $\mathbf{u}^*$, is easily found at the intersection of the lines whose slopes can be calculated using diffusion distances. With reference to Fig.~\ref{subfig:lines_intersect}, the algorithm randomly selects a point in the embedding space, $\mathbf{u}_1$ representing the vector containing the embedding coordinates of the chosen point. For a clear visualization of the problem, the process is illustrated in 2-dimensional space but, as shown before, the embedding space can have a higher dimension depending on the number of demonstrated trajectories or the input space dimension. $\mathbf{u}_2$ is selected searching among the nearest neighbors of $\mathbf{u}_1$. The line direction $\boldsymbol{m}_r = \mathbf{u}_2 - \mathbf{u}_1$ is stored. The process is repeated until all the line directions, $\boldsymbol{m}_s = \mathbf{u}_4 - \mathbf{u}_3$, have been found. Pairwise intersections of the of the found lines can be simply calculated by solving the overdetermined linear system $A\boldsymbol{x}=\boldsymbol{b}$, where \(A=[\boldsymbol{m}_s, \boldsymbol{m}_r]\), \(\boldsymbol{x}=[\alpha, \beta]^T\) and \(\boldsymbol{b}=\mathbf{u}_3-\mathbf{u}_1\) for the two parametric lines, $r = \mathbf{u}_1 + \boldsymbol{m}_r \alpha$ and $s = \mathbf{u}_3 + \boldsymbol{m}_s \beta$. The mean of the intersection points calculated is used to established the position of the attractor, $\mathbf{u}^*$, in the embedding space.

\section{Results}
\label{sec:results}

We validate our approach first at correctly identifying the underlying dynamics of well-known DS. We choose three nonlinear DS known to embed two separate sub-dynamics which are asymptotically stable at two separate attractors. Second, to test the sensitivity of the approach to real and noisy data, we validate the approach to decompose DS generated from handwritten data. For each of the systems, we generate a set of three example trajectories so as to obtain an embedding of the same dimension as the original system, namely $D=2$.

To quantify the clustering results, we compare our approach, to three clustering techniques: Kernel K-means, Spectral Clustering and Gaussian Mixture Model (GMM). GMM adopts full covariance matrix. Kernel K-means and GMM require the number of clusters. For those, we provide them with the correct number of dynamics. For Kernel K-means we adopt, as similarity metric, a Radial Basis Functions (RBF) kernel with the hyper-parameter $\sigma$ set as previously explained. Spectral Clustering adopts \emph{k-nearest} neighborhoods technique for building the adjacency matrix and the identification of the sub-dynamics clusters is achieved through spectral decomposition of the symmetric normalized Laplacian. In order to provide the same information to all the algorithms, the feature state for the compared approaches has been augmented with the velocities.

As none of the clustering techniques we compare can extract the attractors, we compute the reconstruction error for the attractors only for our approach., the {\em Orthogonal Expansion}. The error is calculated as the squared error between the actual and the extracted location, normalized by the standard deviation of the position data set
\begin{equation}
    e(\mathbf{x}^*) = \left( (\mathbf{x}^*_{real} - \mathbf{x}^*_{estimated})^T \Sigma^{-1} (\mathbf{x}^*_{real} - \mathbf{x}^*_{estimated}) \right)^{\frac{1}{2}},
\end{equation}
where $\Sigma$ is the diagonal matrix containing the standard deviation of the dataset.

Note that K-means and GMM are iterative approaches sensitive to initialization. For each of them we provide average performance over $10$ trials as the number of iterations increases.

\begin{example}
    As a first example we consider the following 2-dimensional DS
    \begin{align}
        \dot{x}_1 & = 2x_1 - x_1x_2 \notag \\
        \dot{x}_2 & = 2x_1^2 -x_2,
        \label{eqn:heart}
    \end{align}
    which present two stable equilibrium points in $(-1,2)$ and $(1,2)$. The vector field generated from Eq.~\ref{eqn:heart} and the sampled trajectories are shown in Fig.~\ref{fig:ex_heart}, case (a) and (b) respectively.
    \begin{figure}[ht]
        \vspace{-4mm}
        \centering
        \subfloat[]{
            \scalebox{.25}{\subimport{../figures/}{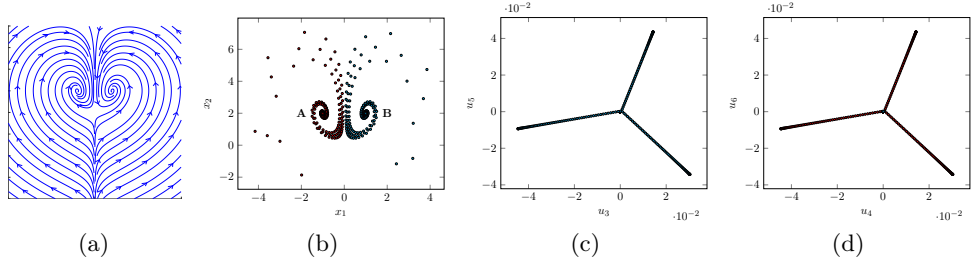}}
            \label{subfig:heart_field}
        }
        \subfloat[]{
            \scalebox{.4}{\begin{tikzpicture}
    \begin{axis}[
            xlabel={$x_1$},
            ylabel={$x_2$},
            scatter/classes={%
                    1={mark=*,black,fill=red,scale=0.5},%
                    2={mark=*,black,fill=cyan,scale=0.5}
                }
        ]
        \addplot[scatter, only marks, scatter src=explicit symbolic, each nth point=5] table[x=x, y=y, meta=labels] {\relativepath ../figures/data/heart_data.csv};
        \node at (axis cs:-2,2) {$\mathbf{A}$};
        \node at (axis cs:2,2) {$\mathbf{B}$};
    \end{axis}
\end{tikzpicture}}
            \label{subfig:heart_data}
        }
        \subfloat[]{
            \scalebox{.4}{\begin{tikzpicture}
    \begin{axis}[
            xlabel={$u_3$},
            ylabel={$u_5$},
            scatter/classes={%
                    1={mark=*,black,fill=red,scale=0.5},%
                    2={mark=*,black,fill=cyan,scale=0.5}
                }
        ]
        \addplot[scatter, only marks, scatter src=explicit symbolic, each nth point=10] table[x=e3, y=e5, meta=labels] {\relativepath ../figures/data/heart_data.csv};
    \end{axis}
\end{tikzpicture}}
            \label{subfig:heart_embed1}
        }
        \subfloat[]{
            \scalebox{.4}{\begin{tikzpicture}
    \begin{axis}[
            xlabel={$u_4$},
            ylabel={$u_6$},
            scatter/classes={%
                    1={mark=*,black,fill=red,scale=0.5},%
                    2={mark=*,black,fill=cyan,scale=0.5}
                }
        ]
        \addplot[scatter, only marks, scatter src=explicit symbolic, each nth point=10] table[x=e4, y=e6, meta=labels] {\relativepath ../figures/data/heart_data.csv};
    \end{axis}
\end{tikzpicture}}
            \label{subfig:heart_embed2}
        }
        \caption{\footnotesize (a) Vector field generated by Eq.~\ref{eqn:heart}. (b) Sampled trajectories from the DS in Fig.\ref{subfig:heart_field}. (c) Embedding space of the sub-dynamics with local attractor in $(1,-2)$. (d) Embedding space of the sub-dynamics with local attractor in $(1,2)$.}
        \label{fig:ex_heart}
        \vspace{-1mm}
    \end{figure}
    Such multiple-attractor DS represents a challenging case for our algorithm due to the proximity of the trajectories belonging to different sub-dynamics, symmetrically displaced with respect to the vertical axis passing through zero, Fig.~\ref{subfig:heart_data}.
    \begin{table}[ht]
        \vspace{-4mm}
        \centering
        \subfloat[]{
            
\begin{tabular}{l|cc|c}
    \toprule
    \# CLUSTER                   & $\mathbf{A}$               & $\mathbf{B}$              & $e(\mathbf{x}^*)$ \\

    \hline
    \multicolumn{4}{c}{Our Approach}                                                                          \\
    \hline

    \cellcolor{red!25}Cluster 1  & \cellcolor{red!25}100\%    & \cellcolor{cyan!25}0\%    & 0.061             \\
    \cellcolor{cyan!25}Cluster 2 & \cellcolor{red!25}0\%      & \cellcolor{cyan!25}100\%  & 0.061             \\

    \hline
    \multicolumn{4}{c}{Kernel K-Means}                                                                        \\
    \hline

    \cellcolor{red!25}Cluster 1  & \cellcolor{cyan!25}0\%     & \cellcolor{red!25}100\%   & -                 \\
    \cellcolor{cyan!25}Cluster 2 & \cellcolor{cyan!25}58.42\% & \cellcolor{red!25}41.58\% & -                 \\

    \hline
    \multicolumn{4}{c}{Spectral Clustering}                                                                   \\
    \hline

    \cellcolor{red!25}Cluster 1  & \cellcolor{cyan!25}0\%     & \cellcolor{red!25}100\%   & -                 \\
    \cellcolor{cyan!25}Cluster 2 & \cellcolor{cyan!25}79.04\% & \cellcolor{red!25}20.96\% & -                 \\

    \hline
    \multicolumn{4}{c}{Gaussian Mixture Models}                                                               \\
    \hline

    \cellcolor{red!25}Cluster 1  & \cellcolor{cyan!25}12.71\% & \cellcolor{red!25}87.29\% & -                 \\
    \cellcolor{cyan!25}Cluster 2 & \cellcolor{cyan!25}12.71\% & \cellcolor{red!25}87.29\% & -                 \\

    \bottomrule
\end{tabular}

            \label{tab:ex1_results}
        }
        \quad
        \subfloat[]{
            \includegraphics[width=.25\textwidth]{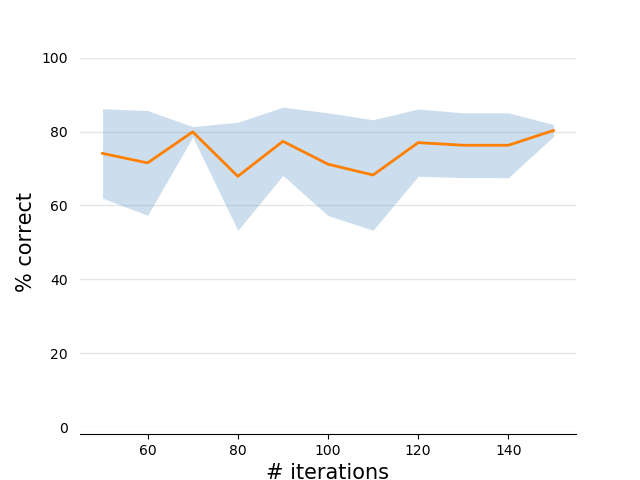}
            \label{subfig:heart_kmeans}
        }
        \quad
        \subfloat[]{
            \includegraphics[width=.25\textwidth]{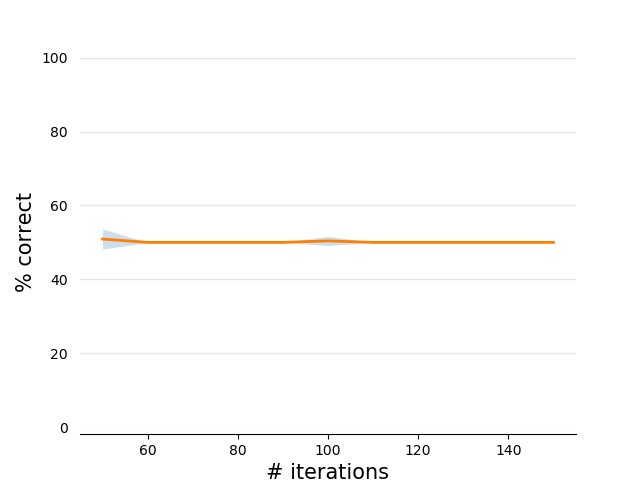}
            \label{subfig:heart_gmm}
        }
        \caption{\footnotesize For sampled points in Fig.~\ref{subfig:heart_data}: (a) Clustering labeling and attractor location error results, (b) Kernel K-Means clustering error over iteration, (c) Gaussian Mixture Model clustering error over iteration.}
    \end{table}
    In order to have a successful reconstruction of the graph we sampled from the DS for $10s$ at a frequency of $100Hz$ reaching a total of $6006$ points sampled.
    \begin{figure}[ht]
        \centering
        \resizebox{.95\textwidth}{!}{\begin{tikzpicture}
    \begin{scope}[shift={(0,0)}]
        \node at (3.5,6.5) {\Large Our Approach};
        \begin{axis}[
                xlabel={$x_1$},
                ylabel={$x_2$},
                scatter/classes={%
                        1={mark=*,red,fill=red,scale=0.5},%
                        2={mark=*,cyan,fill=cyan,scale=0.5}
                    }
            ]
            \addplot[scatter, only marks, scatter src=explicit symbolic, each nth point=5] table[x=x, y=y, meta=our] {\relativepath ../figures/data/heart_results.csv};
        \end{axis}
    \end{scope}

    \begin{scope}[shift={(9,0)}]
        \node at (3.5,6.5) {\Large Kernel K-Means};
        \begin{axis}[
                xlabel={$x_1$},
                ylabel={$x_2$},
                scatter/classes={%
                        1={mark=*,cyan,fill=cyan,scale=0.5},%
                        2={mark=*,red,fill=red,scale=0.5}
                    }
            ]
            \addplot[scatter, only marks, scatter src=explicit symbolic, each nth point=5] table[x=x, y=y, meta=kmeans] {\relativepath ../figures/data/heart_results.csv};
        \end{axis}
    \end{scope}

    \begin{scope}[shift={(18,0)}]
        \node at (3.5,6.5) {\Large Spectral Clustering};
        \begin{axis}[
                xlabel={$x_1$},
                ylabel={$x_2$},
                scatter/classes={%
                        1={mark=*,cyan,fill=cyan,scale=0.5},%
                        2={mark=*,red,fill=red,scale=0.5}
                    }
            ]
            \addplot[scatter, only marks, scatter src=explicit symbolic, each nth point=5] table[x=x, y=y, meta=spectral] {\relativepath ../figures/data/heart_results.csv};
        \end{axis}
    \end{scope}

    \begin{scope}[shift={(27,0)}]
        \node at (3.5,6.5) {\Large Gaussian Mixture Models};
        \begin{axis}[
                xlabel={$x_1$},
                ylabel={$x_2$},
                scatter/classes={%
                        1={mark=*,cyan,fill=cyan,scale=0.5},%
                        2={mark=*,red,fill=red,scale=0.5}
                    }
            ]
            \addplot[scatter, only marks, scatter src=explicit symbolic, each nth point=5] table[x=x, y=y, meta=gmm] {\relativepath ../figures/data/heart_results.csv};
        \end{axis}
    \end{scope}

\end{tikzpicture}}
        \caption{\footnotesize Clustering results of Tab.~\ref{tab:ex1_results}.}
        \label{fig:heart_clustering}
    \end{figure}
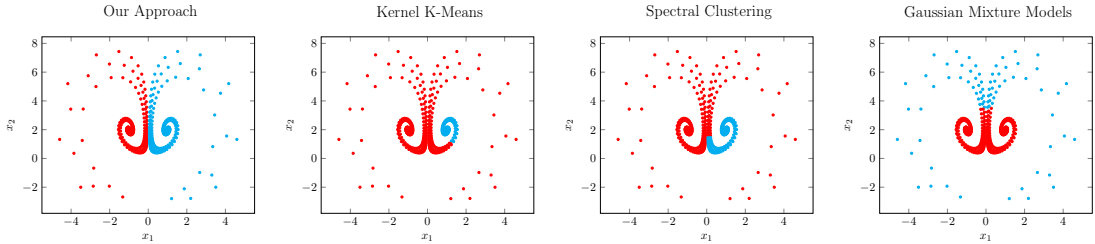
    The correct reconstruction of the graph can be assessed by looking at Fig.~\ref{subfig:heart_embed1} and \ref{subfig:heart_embed2}, where the two sub-dynamics are correctly linearized. In particular using components $\mathbf{u}_3$ and $\mathbf{u}_5$ the embedding space linearizes sub-dynamics $\mathbf{B}$ while sub-dynamics $\mathbf{A}$ is compressed in zero; vice-versa for the embedding space constructed adopting $\mathbf{u}_4$ and $\mathbf{u}_6$ components. Results of clustering are reported in Tab.~\ref{tab:ex1_results} and Fig.\ref{fig:heart_clustering}. Kernel K-means does not yield good performance and, as shown in Fig.~\ref{subfig:heart_kmeans}, it remains fairly sensitive to centroids initialization. Spectral Clustering shows good performance being able to cluster correctly the two high-density region areas. Although it fails in clustering correctly regions with sparse data. GMM present stable solution with respect to parameters initialization, see Fig.~\ref{subfig:heart_gmm}, but poor performance failing in placing the mixtures in a consistent way with respect to the two sub-dynamics. In particular the two high-density regions around the attractors are described by a single component yielding incorrect clustering.
\end{example}

\begin{example}
    We consider in this example the pendulum equation with friction.
    \begin{figure}[ht]
        \vspace{-4mm}
        \centering
        \subfloat[]{
            \scalebox{.25}{\subimport{../figures/}{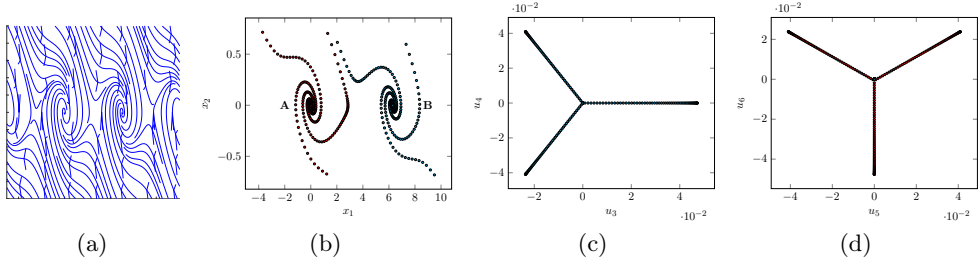}}
            \label{subfig:pendulum_field}
        }
        \subfloat[]{
            \scalebox{.4}{\begin{tikzpicture}
    \begin{axis}[
            xlabel={$x_1$},
            ylabel={$x_2$},
            scatter/classes={%
                    1={mark=*,black,fill=red,scale=0.5},%
                    2={mark=*,black,fill=cyan,scale=0.5}
                }
        ]
        \addplot[scatter, only marks, scatter src=explicit symbolic, each nth point=5] table[x=x, y=y, meta=labels] {\relativepath ../figures/data/pendulum_data.csv};
        \node at (axis cs:-2,0) {$\mathbf{A}$};
        \node at (axis cs:9,0) {$\mathbf{B}$};
    \end{axis}
\end{tikzpicture}}
            \label{subfig:pendulum_data}
        }
        \subfloat[]{
            \scalebox{.4}{\begin{tikzpicture}
    \begin{axis}[
            xlabel={$u_3$},
            ylabel={$u_4$},
            scatter/classes={%
                    1={mark=*,black,fill=red,scale=0.5},%
                    2={mark=*,black,fill=cyan,scale=0.5}
                }
        ]
        \addplot[scatter, only marks, scatter src=explicit symbolic, each nth point=10] table[x=e3, y=e4, meta=labels] {\relativepath ../figures/data/pendulum_data.csv};
    \end{axis}
\end{tikzpicture}}
            \label{subfig:pendulum_embed1}
        }
        \subfloat[]{
            \scalebox{.4}{\begin{tikzpicture}
    \begin{axis}[
            xlabel={$u_5$},
            ylabel={$u_6$},
            scatter/classes={%
                    1={mark=*,black,fill=red,scale=0.5},%
                    2={mark=*,black,fill=cyan,scale=0.5}
                }
        ]
        \addplot[scatter, only marks, scatter src=explicit symbolic, each nth point=10] table[x=e5, y=e6, meta=labels] {\relativepath ../figures/data/pendulum_data.csv};
    \end{axis}
\end{tikzpicture}}
            \label{subfig:pendulum_embed2}
        }
        \caption{\footnotesize (a) Vector field generated by Eq.~\ref{eqn:pendulum}. (b) Sampled trajectories from the DS in Fig.\ref{subfig:pendulum_field}. (c) Embedding space of the sub-dynamics with local attractor in $(0,0)$. (d) Embedding space of the sub-dynamics with local attractor in $(0,2\pi)$.}
        \label{fig:ex_pendulum}
        \vspace{-1mm}
    \end{figure}
    The second order differential equation describing the dynamics of the pendulum is $\ddot{\theta} = -\frac{l}{g}sin(\theta) - kl\dot{\theta}$, where $l$ is the length of the pendulum, $g$ the gravity constant and $k$ the friction coefficient. The phase space of such system yields a multiple-attractor vector fields with periodic stable equilibria at $k 2\pi$, with $k \in \mathbb{Z}$. Let $x_1 = \theta$ and $x_2 = \dot{\theta}$. The pendulum dynamics can be rewritten as a system of first order differential equations
    \begin{align}
        \dot{x}_1 & = x_2                         \notag \\
        \dot{x}_2 & = -\frac{l}{g}sin(x_1) - klx_2
        \label{eqn:pendulum}
    \end{align}
    \begin{table}[ht]
        \vspace{-5mm}
        \centering
        \subfloat[]{
            
\begin{tabular}{l|cc|c}
    \toprule
    \# CLUSTER                   & $\mathbf{A}$               & $\mathbf{B}$              & $e(\mathbf{x}^*)$ \\

    \hline
    \multicolumn{4}{c}{Our Approach}                                                                          \\
    \hline

    \cellcolor{red!25}Cluster 1  & \cellcolor{red!25}100\%    & \cellcolor{cyan!25}0\%    & 1.34\%            \\
    \cellcolor{cyan!25}Cluster 2 & \cellcolor{red!25}0\%      & \cellcolor{cyan!25}100\%  & 1.34\%            \\

    \hline
    \multicolumn{4}{c}{Kernel K-Means}                                                                        \\
    \hline

    \cellcolor{red!25}Cluster 1  & \cellcolor{cyan!25}0\%     & \cellcolor{red!25}100\%   & -                 \\
    \cellcolor{cyan!25}Cluster 2 & \cellcolor{cyan!25}97.67\% & \cellcolor{red!25}2.33\%  & -                 \\

    \hline
    \multicolumn{4}{c}{Spectral Clustering}                                                                   \\
    \hline

    \cellcolor{red!25}Cluster 1  & \cellcolor{cyan!25}3.63\%  & \cellcolor{red!25}96.37\% & -                 \\
    \cellcolor{cyan!25}Cluster 2 & \cellcolor{cyan!25}100\%   & \cellcolor{red!25}0\%     & -                 \\

    \hline
    \multicolumn{4}{c}{Gaussian Mixture Models}                                                               \\
    \hline

    \cellcolor{red!25}Cluster 1  & \cellcolor{cyan!25}0\%     & \cellcolor{red!25}100\%   & -                 \\
    \cellcolor{cyan!25}Cluster 2 & \cellcolor{cyan!25}97.00\% & \cellcolor{red!25}3.00\%  & -                 \\

    \bottomrule
\end{tabular}

            \label{tab:ex2_results}
        }
        \quad
        \subfloat[]{
            \includegraphics[width=.25\textwidth]{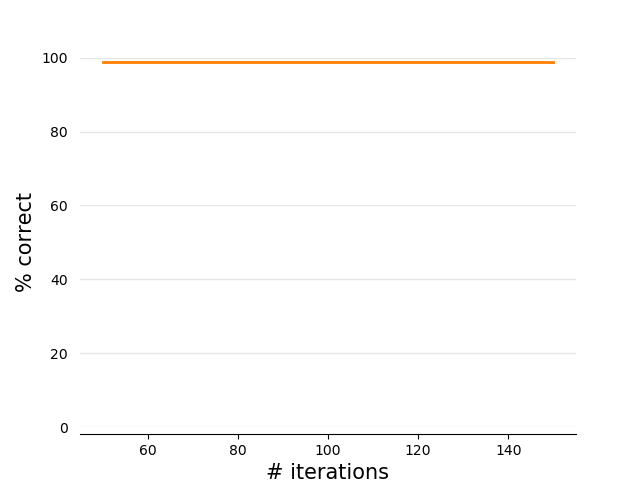}
            \label{subfig:pendulum_kmeans}
        }
        \quad
        \subfloat[]{
            \includegraphics[width=.25\textwidth]{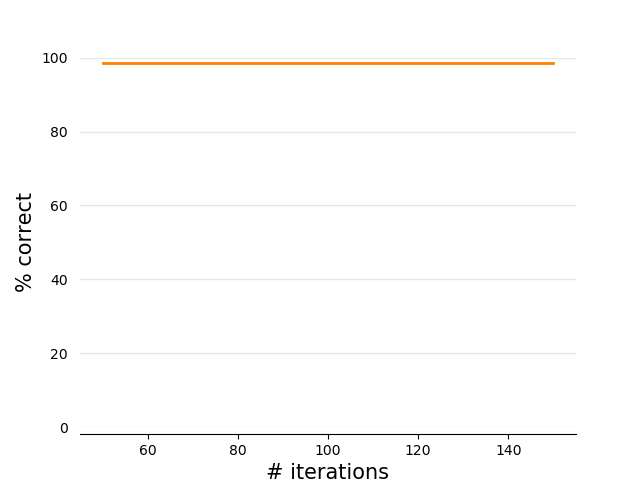}
            \label{subfig:pendulum_gmm}
        }
        \caption{\footnotesize For sampled points in Fig.~\ref{subfig:pendulum_data}: (a) Clustering labeling and attractor location error results, (b) Kernel K-Means clustering error over iteration, (c) Gaussian Mixture Model clustering error over iteration.}
        \vspace{-2mm}
    \end{table}
    We consider a domain $x_1 \in [0, 2\pi]$ where such DS presents 2 attractors at the boundary of the domain and one unstable point at $\pi$. In order to asses the ability of our kernel to generate the desire graph structure to build the Laplacian matrix we appositely sampled proximal trajectories pointing to two different attractors (central region in Fig.~\ref{subfig:pendulum_data}). Each trajectories has been sampled for $60s$ at a frequency of $10Hz$ for a total $3592$ points. As it is possible to see from the embedding spaces extracted, Fig.~\ref{subfig:pendulum_embed1} and \ref{subfig:pendulum_embed2}, the kernel is able to approximate the theoretical graph proposed leading to linearization of the sub-dynamics.
    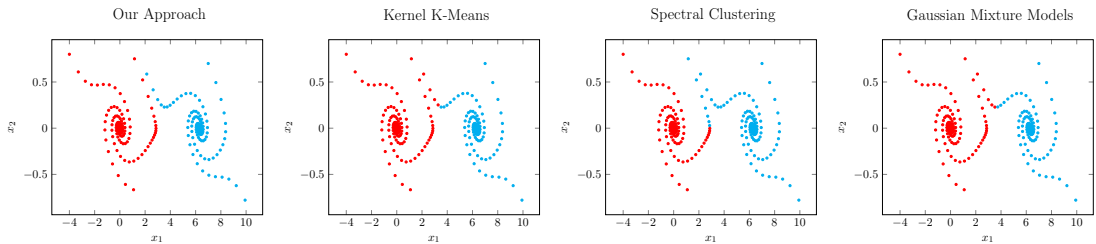
\begin{figure}[ht]
        \centering
        \resizebox{.95\textwidth}{!}{\begin{tikzpicture}
    \begin{scope}[shift={(0,0)}]
        \node at (3.5,6.5) {\Large Our Approach};
        \begin{axis}[
                xlabel={$x_1$},
                ylabel={$x_2$},
                scatter/classes={%
                        1={mark=*,red,fill=red,scale=0.5},%
                        2={mark=*,cyan,fill=cyan,scale=0.5}
                    }
            ]
            \addplot[scatter, only marks, scatter src=explicit symbolic, each nth point=10] table[x=x, y=y, meta=our] {\relativepath ../figures/data/pendulum_results.csv};
        \end{axis}
    \end{scope}

    \begin{scope}[shift={(9,0)}]
        \node at (3.5,6.5) {\Large Kernel K-Means};
        \begin{axis}[
                xlabel={$x_1$},
                ylabel={$x_2$},
                scatter/classes={%
                        1={mark=*,red,fill=red,scale=0.5},%
                        2={mark=*,cyan,fill=cyan,scale=0.5}
                    }
            ]
            \addplot[scatter, only marks, scatter src=explicit symbolic, each nth point=10] table[x=x, y=y, meta=kmeans] {\relativepath ../figures/data/pendulum_results.csv};
        \end{axis}
    \end{scope}

    \begin{scope}[shift={(18,0)}]
        \node at (3.5,6.5) {\Large Spectral Clustering};
        \begin{axis}[
                xlabel={$x_1$},
                ylabel={$x_2$},
                scatter/classes={%
                        1={mark=*,red,fill=red,scale=0.5},%
                        2={mark=*,cyan,fill=cyan,scale=0.5}
                    }
            ]
            \addplot[scatter, only marks, scatter src=explicit symbolic, each nth point=10] table[x=x, y=y, meta=spectral] {\relativepath ../figures/data/pendulum_results.csv};
        \end{axis}
    \end{scope}

    \begin{scope}[shift={(27,0)}]
        \node at (3.5,6.5) {\Large Gaussian Mixture Models};
        \begin{axis}[
                xlabel={$x_1$},
                ylabel={$x_2$},
                scatter/classes={%
                        1={mark=*,cyan,fill=cyan,scale=0.5},%
                        2={mark=*,red,fill=red,scale=0.5}
                    }
            ]
            \addplot[scatter, only marks, scatter src=explicit symbolic, each nth point=10] table[x=x, y=y, meta=gmm] {\relativepath ../figures/data/pendulum_results.csv};
        \end{axis}
    \end{scope}

\end{tikzpicture}}
        \caption{\footnotesize Clustering results of Tab.~\ref{tab:ex2_results}.}
        \label{fig:pendulum_clustering}
    \end{figure}
    In Tab.~\ref{tab:ex2_results} and Fig.~\ref{fig:pendulum_clustering} the results of clustering are shown. All the algorithms yields good performance in this case correctly assigning points belonging to high-density regions around attractors. Nevertheless, with exception of our approach, they all fail to achieve perfect clustering due to mis-clustering in regions with sparse data.
\end{example}

\begin{example}
    As a third example we consider the Duffing equation (or oscillator),
    \begin{figure}[ht]
        \vspace{-4mm}
        \centering
        \subfloat[]{
            \scalebox{.25}{\subimport{../figures/}{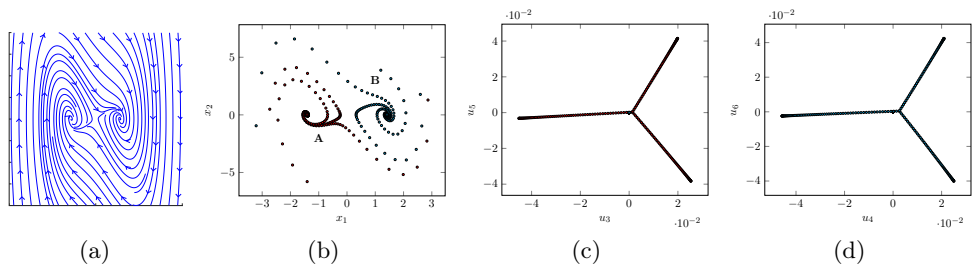}}
            \label{subfig:duffing_field}
        }
        \subfloat[]{
            \scalebox{.4}{\begin{tikzpicture}
    \begin{axis}[
            xlabel={$x_1$},
            ylabel={$x_2$},
            scatter/classes={%
                    1={mark=*,black,fill=red,scale=0.5},%
                    2={mark=*,black,fill=cyan,scale=0.5}
                }
        ]
        \addplot[scatter, only marks, scatter src=explicit symbolic, each nth point=5] table[x=x, y=y, meta=labels] {\relativepath ../figures/data/duffing_data.csv};
        \node at (axis cs:-1,-2) {$\mathbf{A}$};
        \node at (axis cs:1,3) {$\mathbf{B}$};
    \end{axis}
\end{tikzpicture}}
            \label{subfig:duffing_data}
        }
        \subfloat[]{
            \scalebox{.4}{\begin{tikzpicture}
    \begin{axis}[
            xlabel={$u_3$},
            ylabel={$u_5$},
            scatter/classes={%
                    1={mark=*,black,fill=red,scale=0.5},%
                    2={mark=*,black,fill=cyan,scale=0.5}
                }
        ]
        \addplot[scatter, only marks, scatter src=explicit symbolic, each nth point=10] table[x=e3, y=e5, meta=labels] {\relativepath ../figures/data/duffing_data.csv};
    \end{axis}
\end{tikzpicture}}
            \label{subfig:duffing_embed1}
        }
        \subfloat[]{
            \scalebox{.4}{\begin{tikzpicture}
    \begin{axis}[
            xlabel={$u_4$},
            ylabel={$u_6$},
            scatter/classes={%
                    1={mark=*,black,fill=red,scale=0.5},%
                    2={mark=*,black,fill=cyan,scale=0.5}
                }
        ]
        \addplot[scatter, only marks, scatter src=explicit symbolic, each nth point=10] table[x=e4, y=e6, meta=labels] {\relativepath ../figures/data/duffing_data.csv};
    \end{axis}
\end{tikzpicture}}
            \label{subfig:duffing_embed2}
        }
        \caption{\footnotesize (a) Vector field generated by Eq.~\ref{eqn:duffing}. (b) Sampled trajectories from the DS in Fig.\ref{subfig:duffing_field}. (c) Embedding space of the sub-dynamics with local attractor in $(0,\sqrt{-\alpha / \beta})$. (d) Embedding space of the sub-dynamics with local attractor in $(0,-\sqrt{-\alpha / \beta})$.}
        \label{fig:ex_duffing}
    \end{figure}
    a non-linear second-order differential equation used for modeling damped oscillator
    \begin{equation}
        \ddot{x} + \delta \dot{x} + \alpha x + \beta x^3 = 0.
    \end{equation}
    Differently from a simple harmonic oscillator such equation models more complex potential by including a non-linear spring with restoring force $\alpha x + \beta x^3$. Such DS exhibits chaotic behavior. We consider positive damping case, $\delta = 0.3$, with $\alpha = -1.2 < 0$ and $\beta = 0.3 > 0$ which present two stable at $+ \sqrt{-\alpha / \beta}$ and $- \sqrt{-\alpha / \beta}$. The phase space of the duffing equation can be achieved by the following two dimension DS
    \begin{align}
        \dot{x}_1 & = x_2                    \notag \\
        \dot{x}_2 & = 0.3(4x_1 - x_1^3 -x_2).
        \label{eqn:duffing}
    \end{align}
    As for example 1, due the presence of highly compact and non-linear regions, high frequency sampling is required. Each trajectory is sample at $200Hz$ for $5s$ yielding a dataset of $6006$ samples.
    \begin{table}[ht]
        \vspace{-4mm}
        \centering
        \subfloat[]{
            
\begin{tabular}{l|cc|c}
    \toprule
    \# CLUSTER                   & $\mathbf{A}$               & $\mathbf{B}$               & $e(\mathbf{x}^*)$ \\

    \hline
    \multicolumn{4}{c}{Our Approach}                                                                           \\
    \hline

    \cellcolor{red!25}Cluster 1  & \cellcolor{red!25}100\%    & \cellcolor{cyan!25}0\%     & 1.34\%            \\
    \cellcolor{cyan!25}Cluster 2 & \cellcolor{red!25}0\%      & \cellcolor{cyan!25}100\%   & 1.34\%            \\

    \hline
    \multicolumn{4}{c}{Kernel K-Means}                                                                         \\
    \hline

    \cellcolor{red!25}Cluster 1  & \cellcolor{red!25}100\%    & \cellcolor{cyan!25}0\%     & -                 \\
    \cellcolor{cyan!25}Cluster 2 & \cellcolor{red!25}27.81\%  & \cellcolor{cyan!25}72.19\% & -                 \\

    \hline
    \multicolumn{4}{c}{Spectral Clustering}                                                                    \\
    \hline

    \cellcolor{red!25}Cluster 1  & \cellcolor{red!25}94.61\%  & \cellcolor{cyan!25}5.39\%  & -                 \\
    \cellcolor{cyan!25}Cluster 2 & \cellcolor{red!25}5.65\%   & \cellcolor{cyan!25}94.35\% & -                 \\

    \hline
    \multicolumn{4}{c}{Gaussian Mixture Models}                                                                \\
    \hline

    \cellcolor{red!25}Cluster 1  & \cellcolor{cyan!25}4.12\%  & \cellcolor{red!25}95.88\%  & -                 \\
    \cellcolor{cyan!25}Cluster 2 & \cellcolor{cyan!25}94.88\% & \cellcolor{red!25}5.12\%   & -                 \\

    \bottomrule
\end{tabular}

            \label{tab:ex3_results}
        }
        \quad
        \subfloat[]{
            \includegraphics[width=.25\textwidth]{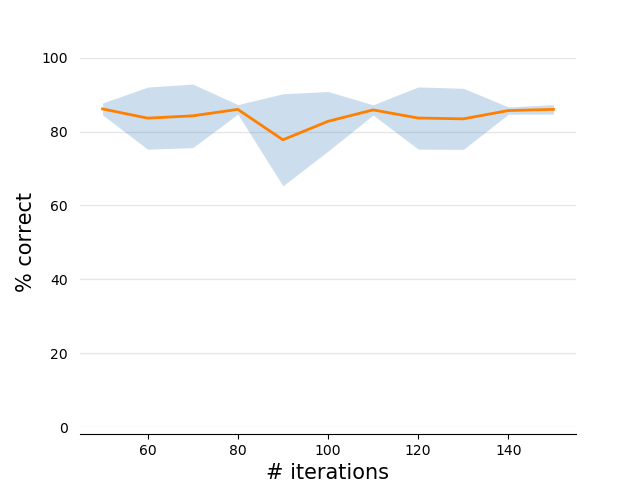}
            \label{subfig:duffing_kmeans}
        }
        \quad
        \subfloat[]{
            \includegraphics[width=.25\textwidth]{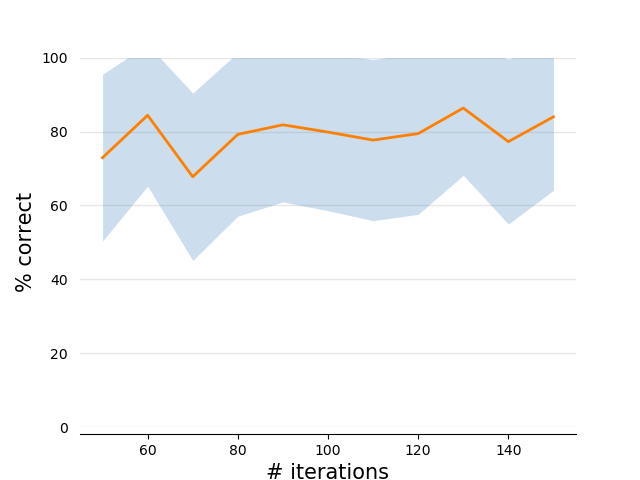}
            \label{subfig:duffing_gmm}
        }
        \caption{\footnotesize For sampled points in Fig.~\ref{subfig:duffing_data}: (a) Clustering labeling and attractor location error results, (b) Kernel K-Means clustering error over iteration, (c) Gaussian Mixture Model clustering error over iteration.}
    \end{table}
    For this case eigenvectors $\mathbf{u}_3$ and $\mathbf{u}_5$ yield an embedding space where sub-dynamics $\mathbf{A}$ is linearized while the space generate by $\mathbf{u}_4$ and $\mathbf{u}_6$ achieve linearization of sub-dynamics $\mathbf{B}$.
    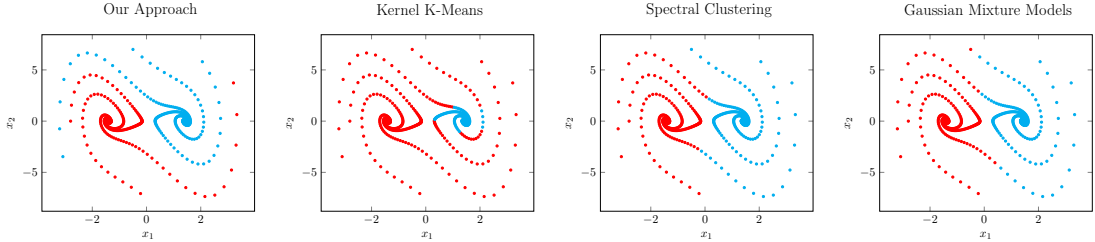
\begin{figure}[ht]
        \centering
        \resizebox{.95\textwidth}{!}{\begin{tikzpicture}
    \begin{scope}[shift={(0,0)}]
        \node at (3.5,6.5) {\Large Our Approach};
        \begin{axis}[
                xlabel={$x_1$},
                ylabel={$x_2$},
                scatter/classes={%
                        1={mark=*,red,fill=red,scale=0.5},%
                        2={mark=*,cyan,fill=cyan,scale=0.5}
                    }
            ]
            \addplot[scatter, only marks, scatter src=explicit symbolic, each nth point=5] table[x=x, y=y, meta=our] {\relativepath ../figures/data/duffing_results.csv};
        \end{axis}
    \end{scope}

    \begin{scope}[shift={(9,0)}]
        \node at (3.5,6.5) {\Large Kernel K-Means};
        \begin{axis}[
                xlabel={$x_1$},
                ylabel={$x_2$},
                scatter/classes={%
                        1={mark=*,red,fill=red,scale=0.5},%
                        2={mark=*,cyan,fill=cyan,scale=0.5}
                    }
            ]
            \addplot[scatter, only marks, scatter src=explicit symbolic, each nth point=5] table[x=x, y=y, meta=kmeans] {\relativepath ../figures/data/duffing_results.csv};
        \end{axis}
    \end{scope}

    \begin{scope}[shift={(18,0)}]
        \node at (3.5,6.5) {\Large Spectral Clustering};
        \begin{axis}[
                xlabel={$x_1$},
                ylabel={$x_2$},
                scatter/classes={%
                        1={mark=*,red,fill=red,scale=0.5},%
                        2={mark=*,cyan,fill=cyan,scale=0.5}
                    }
            ]
            \addplot[scatter, only marks, scatter src=explicit symbolic, each nth point=5] table[x=x, y=y, meta=spectral] {\relativepath ../figures/data/duffing_results.csv};
        \end{axis}
    \end{scope}

    \begin{scope}[shift={(27,0)}]
        \node at (3.5,6.5) {\Large Gaussian Mixture Models};
        \begin{axis}[
                xlabel={$x_1$},
                ylabel={$x_2$},
                scatter/classes={%
                        1={mark=*,cyan,fill=cyan,scale=0.5},%
                        2={mark=*,red,fill=red,scale=0.5}
                    }
            ]
            \addplot[scatter, only marks, scatter src=explicit symbolic, each nth point=5] table[x=x, y=y, meta=gmm] {\relativepath ../figures/data/duffing_results.csv};
        \end{axis}
    \end{scope}

\end{tikzpicture}}
        \caption{\footnotesize Clustering results of Tab.~\ref{tab:ex3_results}.}
        \label{fig:duffing_clustering}
    \end{figure}
    The results of clustering in Tab.~\ref{tab:ex3_results} and Fig.~\ref{fig:duffing_clustering} show a similar behavior for Spectral Clustering and GMM. Both algorithms yields quantitative good results. Although, due to the chaotic behavior of the DS they fail in clustering correctly trajectories close to the attractor belonging to another sub-dynamics. Kernel K-means is able to cluster correctly points lying close to the attractors but it is incapable of clustering the majority of the trajectories.
\end{example}

\begin{example}
    As last examples
    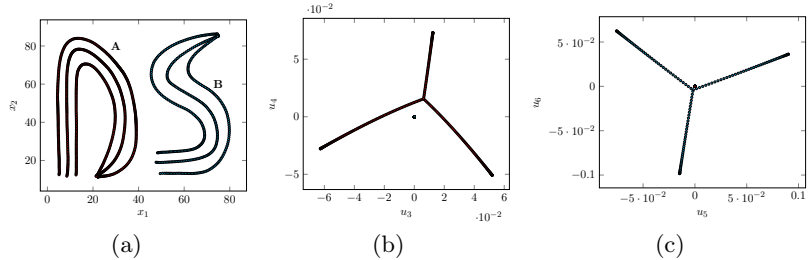
\begin{figure}[ht]
        \vspace{-4mm}
        \centering
        \subfloat[]{
            \scalebox{.4}{\begin{tikzpicture}
    \begin{axis}[
            xlabel={$x_1$},
            ylabel={$x_2$},
            scatter/classes={%
                    1={mark=*,black,fill=red,scale=0.5},%
                    2={mark=*,black,fill=cyan,scale=0.5}
                }
        ]
        \addplot[scatter, only marks, scatter src=explicit symbolic] table[x=x, y=y, meta=labels] {\relativepath ../figures/data/drawn_data.csv};
        \node at (axis cs:30,80) {$\mathbf{A}$};
        \node at (axis cs:75,60) {$\mathbf{B}$};
    \end{axis}
\end{tikzpicture}}
            \label{subfig:handdrawn_data}
        }
        \subfloat[]{
            \scalebox{.4}{\begin{tikzpicture}
    \begin{axis}[
            xlabel={$u_3$},
            ylabel={$u_4$},
            scatter/classes={%
                    1={mark=*,black,fill=red,scale=0.5},%
                    2={mark=*,black,fill=cyan,scale=0.5}
                }
        ]
        \addplot[scatter, only marks, scatter src=explicit symbolic] table[x=e3, y=e4, meta=labels, each nth point=3] {\relativepath ../figures/data/drawn_data.csv};
    \end{axis}
\end{tikzpicture}}
            \label{subfig:handdrawn_embed1}
        }
        \subfloat[]{
            \scalebox{.4}{\begin{tikzpicture}
    \begin{axis}[
            xlabel={$u_5$},
            ylabel={$u_6$},
            scatter/classes={%
                    1={mark=*,black,fill=red,scale=0.5},%
                    2={mark=*,black,fill=cyan,scale=0.5}
                }
        ]
        \addplot[scatter, only marks, scatter src=explicit symbolic] table[x=e5, y=e6, meta=labels,  each nth point=3] {\relativepath ../figures/data/drawn_data.csv};
    \end{axis}
\end{tikzpicture}}
            \label{subfig:handdrawn_embed2}
        }
        \caption{\footnotesize (a) Demonstrated trajectories. (b) Embedding space of the red sub-dynamics. (d) Embedding space of the cyan sub-dynamics.}
        \label{fig:ex_handdrawn}
    \end{figure}
    we show applicability to hand-drawn multiple-attractor DS where our algorithm can be used to cluster the sub-dynamics and locate the various attractors for then providing such information to stable learning algorithms present in the literature. Fig.~\ref{subfig:handdrawn_data} shows the demonstrated trajectories for a 2-attractor DS in 2D. For drawing such trajectories we took advantage of Wacom tablet and the software provided by ML\_toolbox\footnote{https://github.com/epfl-lasa/ML\_toolbox}.
    \begin{table}[ht]
        \centering
        \subfloat[]{
            
\begin{tabular}{l|cc|c}
    \toprule
    \# CLUSTER                   & $\mathbf{A}$               & $\mathbf{B}$               & $e(\mathbf{x}^*)$ \\

    \hline
    \multicolumn{4}{c}{Our Approach}                                                                           \\
    \hline

    \cellcolor{red!25}Cluster 1  & \cellcolor{red!25}100\%    & \cellcolor{cyan!25}0\%     & 1.56\%            \\
    \cellcolor{cyan!25}Cluster 2 & \cellcolor{red!25}0\%      & \cellcolor{cyan!25}100\%   & 1.56\%            \\

    \hline
    \multicolumn{4}{c}{Kernel K-Means}                                                                         \\
    \hline

    \cellcolor{red!25}Cluster 1  & \cellcolor{cyan!25}48.85\% & \cellcolor{red!25}51.15\%  & -                 \\
    \cellcolor{cyan!25}Cluster 2 & \cellcolor{cyan!25}48.67\% & \cellcolor{red!25}51.33\%  & -                 \\

    \hline
    \multicolumn{4}{c}{Spectral Clustering}                                                                    \\
    \hline

    \cellcolor{red!25}Cluster 1  & \cellcolor{red!25}100\%    & \cellcolor{cyan!25}0\%     & -                 \\
    \cellcolor{cyan!25}Cluster 2 & \cellcolor{red!25}0\%      & \cellcolor{cyan!25}100\%   & -                 \\

    \hline
    \multicolumn{4}{c}{Gaussian Mixture Models}                                                                \\
    \hline

    \cellcolor{red!25}Cluster 1  & \cellcolor{red!25}55.23\%  & \cellcolor{cyan!25}44.77\% & -                 \\
    \cellcolor{cyan!25}Cluster 2 & \cellcolor{red!25}47.71\%  & \cellcolor{cyan!25}52.29\% & -                 \\

    \bottomrule
\end{tabular}

            \label{tab:ex4_results}
        }
        \quad
        \subfloat[]{
            \includegraphics[width=.25\textwidth]{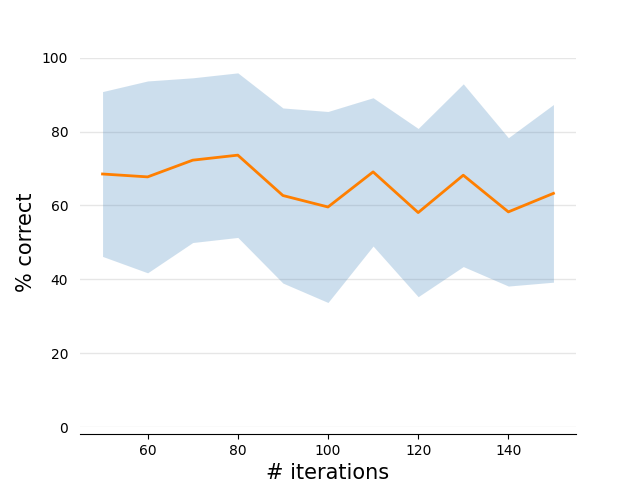}
            \label{subfig:drawn_kmeans}
        }
        \quad
        \subfloat[]{
            \includegraphics[width=.25\textwidth]{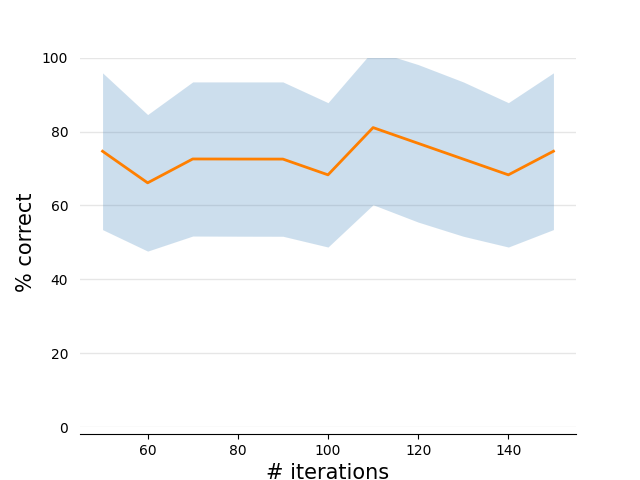}
            \label{subfig:drawn_gmm}
        }
        \caption{\footnotesize For sampled points in Fig.~\ref{subfig:handdrawn_data}: (a) Clustering labeling and attractor location error results, (b) Kernel K-Means clustering error over iteration, (c) Gaussian Mixture Model clustering error over iteration.}
    \end{table}
    Spectral Clustering is able to achieve perfect clustering while GMM misplaces the Gaussian components yielding poor results.
    \begin{figure}[h!]
        \centering
        \resizebox{.95\textwidth}{!}{\begin{tikzpicture}
    \begin{scope}[shift={(0,0)}]
        \node at (3.5,6.5) {\Large Our Approach};
        \begin{axis}[
                xlabel={$x_1$},
                ylabel={$x_2$},
                scatter/classes={%
                        1={mark=*,red,fill=red,scale=0.5},%
                        2={mark=*,cyan,fill=cyan,scale=0.5}
                    }
            ]
            \addplot[scatter, only marks, scatter src=explicit symbolic, each nth point=2] table[x=x, y=y, meta=our] {\relativepath ../figures/data/drawn_results.csv};
        \end{axis}
    \end{scope}

    \begin{scope}[shift={(9,0)}]
        \node at (3.5,6.5) {\Large Kernel K-Means};
        \begin{axis}[
                xlabel={$x_1$},
                ylabel={$x_2$},
                scatter/classes={%
                        1={mark=*,red,fill=red,scale=0.5},%
                        2={mark=*,cyan,fill=cyan,scale=0.5}
                    }
            ]
            \addplot[scatter, only marks, scatter src=explicit symbolic, each nth point=2] table[x=x, y=y, meta=kmeans] {\relativepath ../figures/data/drawn_results.csv};
        \end{axis}
    \end{scope}

    \begin{scope}[shift={(18,0)}]
        \node at (3.5,6.5) {\Large Spectral Clustering};
        \begin{axis}[
                xlabel={$x_1$},
                ylabel={$x_2$},
                scatter/classes={%
                        1={mark=*,red,fill=red,scale=0.5},%
                        2={mark=*,cyan,fill=cyan,scale=0.5}
                    }
            ]
            \addplot[scatter, only marks, scatter src=explicit symbolic, each nth point=2] table[x=x, y=y, meta=spectral] {\relativepath ../figures/data/drawn_results.csv};
        \end{axis}
    \end{scope}

    \begin{scope}[shift={(27,0)}]
        \node at (3.5,6.5) {\Large Gaussian Mixture Models};
        \begin{axis}[
                xlabel={$x_1$},
                ylabel={$x_2$},
                scatter/classes={%
                        1={mark=*,cyan,fill=cyan,scale=0.5},%
                        2={mark=*,red,fill=red,scale=0.5}
                    }
            ]
            \addplot[scatter, only marks, scatter src=explicit symbolic, each nth point=2] table[x=x, y=y, meta=gmm] {\relativepath ../figures/data/drawn_results.csv};
        \end{axis}
    \end{scope}

\end{tikzpicture}}
        \caption{\footnotesize Clustering results of Tab.~\ref{tab:ex4_results}.}
        \label{fig:drawn_clustering}
    \end{figure}
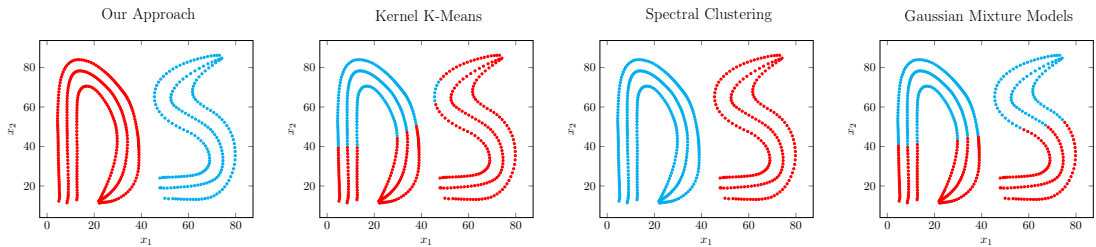
    Kernel K-means yields particular poor results in this case when a small kernel width such the one adopted for our algorithm is used. In order to increase the performance of Kernel K-means we set the kernel width equal to the standard deviation of the dataset.
    \begin{figure}[ht]
        \centering
        \subfloat[]{\includegraphics[width=.25\textwidth]{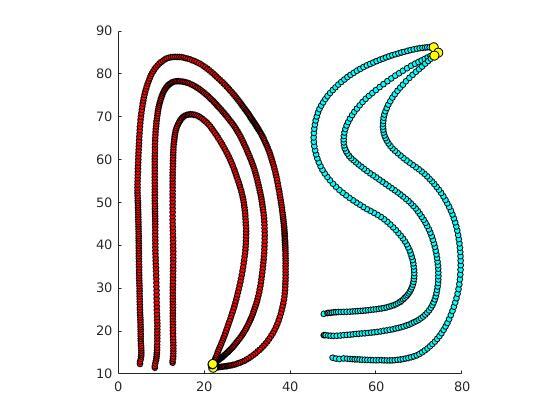}}
        \subfloat[]{\includegraphics[width=.25\textwidth]{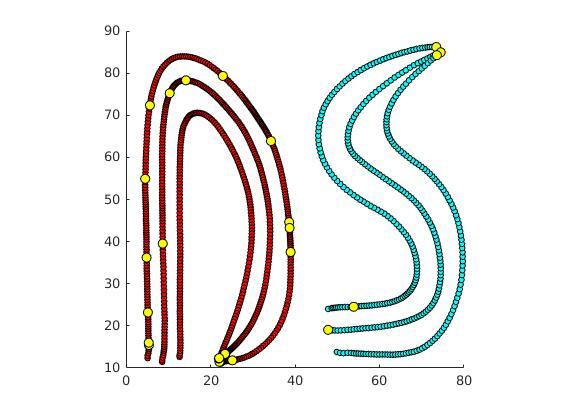}}
        \subfloat[]{\includegraphics[width=.25\textwidth]{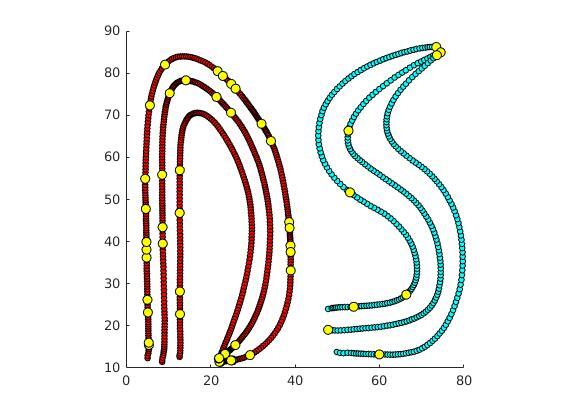}}
        \caption{\footnotesize Quasi-zero velocity heuristic for attractor location. Threshold at (a) 10\%, (b) 5\% and (c) 1\% of the average velocity.}
        \label{fig:zero_velocity}
    \end{figure}
    Nevertheless Kernel K-means is not able to consistently cluster the two sub-dynamics as shown in Fig.~\ref{subfig:drawn_kmeans}. For such case the real attractor location is assumed to be the mean average of the position of the last point of the demonstrated trajectories. For this specific case we show results for the location of the attractor based on quasi-zero velocity heuristic. Fig.\ref{fig:zero_velocity} shows the identification of potential locations for the attractors based on different velocity thresholds. Notice that zero-velocity crossing ends up with identifying multiple and incorrect attractor locations. This happens often at the beginning of a trajectory or in region of high curvature where the velocity can be proximal to zero.
\end{example}

\section{Dynamical System Learning via NVP Transformations}
\begin{figure}[ht]
    \centering
    \scalebox{2}{\subimport{../figures/}{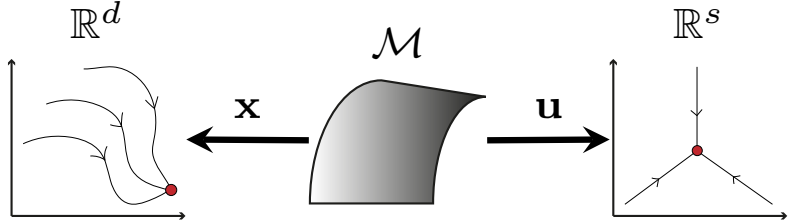}}
    \caption{\footnotesize (left) Original Euclidean Space where the dataset is sampled from; (center) manifold where the DS is taking place; (right) extracted Euclidean embedding space.}
    \label{fig:diffeomorphism}
\end{figure}

From a differential geometry perspective Manifold Learning has a simple and straightforward interpretation. Consider the \emph{differentiable} manifold $\mathcal{M}$ in Fig.~\ref{fig:diffeomorphism}. The maps $\mathbf{x}$ and $\mathbf{u}$, generally called \emph{chart maps}, represent two different (local) representation of the manifold in the Euclidean spaces $\mathbb{R}^d$ and $\mathbb{R}^s$.

Going back to the problem of learning stable DS, we view trajectories as motions over a differentiable manifold $\mathcal{M}$. Starting from an Euclidean representation $\mathbb{R}^d$ of such motions (the original dataset), our approach is capable of reconstructing an alternative Euclidean representation $\mathbb{R}^s$ of the manifold in which the DS looks linear. If the number of selected eigenvectors to reconstruct the embedding space is equal to the dimension of the original space, namely $s=d$, such ensemble represents the discrete counterpart of the continuous diffeomorphic map that links the two different representation of the manifold $\mathbf{u} \circ \mathbf{x}^{-1} : \mathbb{R}^d \rightarrow \mathbb{R}^s$. A differentiable manifold ensures that any two charts (representations of the same manifold) are differentiable-compatible, namely $\mathbf{u} \circ \mathbf{x}^{-1}$ and its inverse $\mathbf{x} \circ \mathbf{u}^{-1}$ are continuous and differentiable. This properties allows to relate velocity components in the two different Euclidean spaces by the means of the so called Jacobian. For easier notation let the diffeomorphism be $\psi = \mathbf{u} \circ \mathbf{x}^{-1}$, and the Jacobian $\mathbf{J}_{\psi} = \partial_{\mathbf{x}} \psi$ then
\begin{equation}
    \dot{\mathbf{u}} = \mathbf{J}_{\psi} \dot{\mathbf{x}}.
    \label{eqn:y_to_x}
\end{equation}
It is easy to show that if Lyapunov stability is guaranteed in one Euclidean representation of $\mathcal{M}$ the diffeomorphism \emph{induces} the same property in the other one. Consider the existence of a Lyapunov function $V : \mathbb{R}^d \rightarrow \mathbb{R}$ radially unbounded and positive in all the space with $\dot{V}(\mathbf{x}) < 0$ except in $\mathbf{x}^*$, equilibrium point, where $V(\mathbf{x}) = \dot{V}(\mathbf{x}) = 0$. The diffeomorphism $\psi$ generates a Lyapunov function $\tilde{V} : \mathbb{R}^s \rightarrow \mathbb{R}$ and a related attractor $\mathbf{u}^* = \psi(\mathbf{x}^*)$
\begin{equation}
    \dot{\tilde{V}}(\mathbf{u}) = \frac{\partial V}{\partial \mathbf{x}} \frac{\partial \psi^{-1}}{\partial \mathbf{u}} \frac{\partial \mathbf{u}}{\partial t} = \frac{\partial V}{\partial \mathbf{x}} \mathbf{J}_{\psi}^{-1} \mathbf{J}_{\psi} \dot{\mathbf{x}} = \frac{\partial V}{\partial \mathbf{x}} \dot{\mathbf{x}} = \dot{V}(\mathbf{x}) < 0.
\end{equation}
Given the bijectivity of the diffeomorphism if $\tilde{V}$ is a Lyapunov function defining a stable equilibrium point at $\mathbf{u}^*$, $\mathbf{x}^*$ is a globally asymptotically stable equilibrium point. We apply this principle to reconstruct the dynamics in the original space.

We start by constructing a linear DS in embedding space that follows the linearized trajectories of our DS: $\dot{\mathbf{u}} = \mathbf{u}^* - \mathbf{u}$. This system is globally asymptotically stable at $\mathbf{u}^*$. Stability can be proved easily by considering the quadratic Lyapunov function $\tilde{V}(\mathbf{u}) = \frac{1}{2}(\mathbf{u}^* - \mathbf{u})^T(\mathbf{u}^* - \mathbf{u})$. Observe that $\tilde{V}$ is also the potential of the vector field (namely $\dot{\mathbf{u}} = \nabla \tilde{V}$). Following from Eq.~\ref{eqn:y_to_x}, the original dynamics can be recovered through
\begin{equation}
    \dot{\mathbf{x}} = \mathbf{J}_{\psi}^{-1} (\mathbf{u}^* - \mathbf{u}) = \mathbf{J}_{\psi}^{-1} (\psi(\mathbf{x}^*) - \psi(\mathbf{x})).
    \label{eqn:diff_ds}
\end{equation}
The corresponding (deformed) Lyapunov function in original space can be recovered as $V = \tilde{V} \circ \psi = \frac{1}{2}(\psi(\mathbf{x}^*) - \psi(\mathbf{x}))^T(\psi(\mathbf{x}^*) - \psi(\mathbf{x}))$.

In order to reconstruct the diffeomorphic map using the embedding space as a ground truth, we adopt Non-Volume Preserving (NVP) transformations introduced by \cite{dinh2016density}. The diffeomorphism is obtained through a sequence of $k$ \emph{coupling layers} each of which is given by:
\begin{equation}
    \begin{cases}
        \mathbf{u}_{1:n}   & = \mathbf{x}_{1:n}                                                               \\
        \mathbf{u}_{n+1:d} & = \mathbf{x}_{n+1:d} \odot \text{exp}(s(\mathbf{x}_{1:n})) + t(\mathbf{x}_{1:n})
    \end{cases}
\end{equation}
with $n<d$, $d$ the dimension of the original space. $s(\cdot)$ and $t(\cdot)$ are scaling and translating functions, respectively. Each of these functions is approximated through Random Fourier feature approximation (\cite{rahimi2007random}) of a vector-valued isotropic Radial Basis Functions kernel as shown by \cite{rana_euclideanizing_2020}.
\begin{figure}[ht]
    \centering
    \subfloat[]{
        \includegraphics[scale=.24]{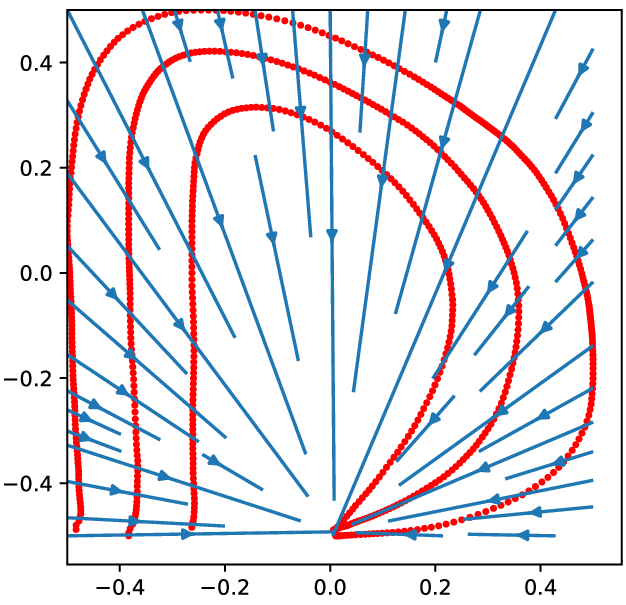}
        \label{subfig:ds1_dynamics_original}
    }
    \quad
    \subfloat[]{
        \includegraphics[scale=.24]{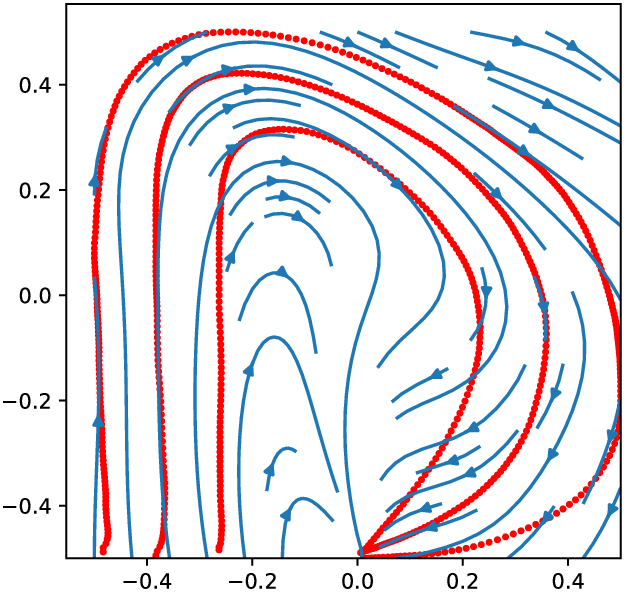}
        \label{subfig:ds1_dynamics_deformed}
    }
    \quad
    \subfloat[]{
        \scalebox{1.2}{\subimport{../figures/}{ds1_potential_original}}
        \label{subfig:ds1_potential_original}
    }
    \quad
    \subfloat[]{
        \scalebox{1.2}{\subimport{../figures/}{ds1_potential_deformed}}
        \label{subfig:ds1_potential_deformed}
    }
    \caption{\footnotesize (a) Linear DS generated when using identity diffeomorphism; (b) reconstructed dynamics through learned diffeomorphism; (c) initial quadratic potential function generating a linear DS in Fig.~\ref{subfig:ds1_dynamics_original}; (d) deformed potential under the action of the learned diffeomorphism generating the nonlinear DS in Fig.~\ref{subfig:ds1_dynamics_deformed}.}
    \label{fig:ds1_learn}
\end{figure}
As loss function, we use the standard Mean Squared Error, $\text{MSE} = \sum_{i=1}^{N_D} \norm{\mathbf{u}_i - \psi(\mathbf{x}_i)}^2$, adopting the embedding space coordinates as ground truth.

\begin{figure}[ht]
    \centering
    \subfloat[]{
        \includegraphics[scale=.24]{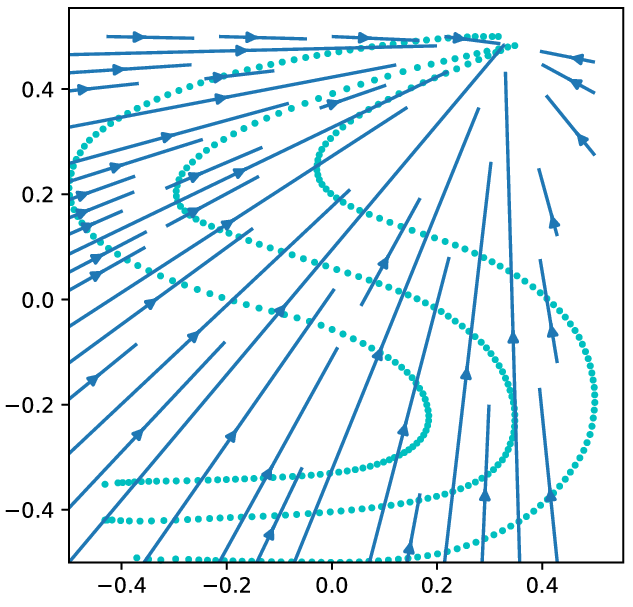}
        \label{subfig:ds2_dynamics_original}
    }
    \quad
    \subfloat[]{
        \includegraphics[scale=.24]{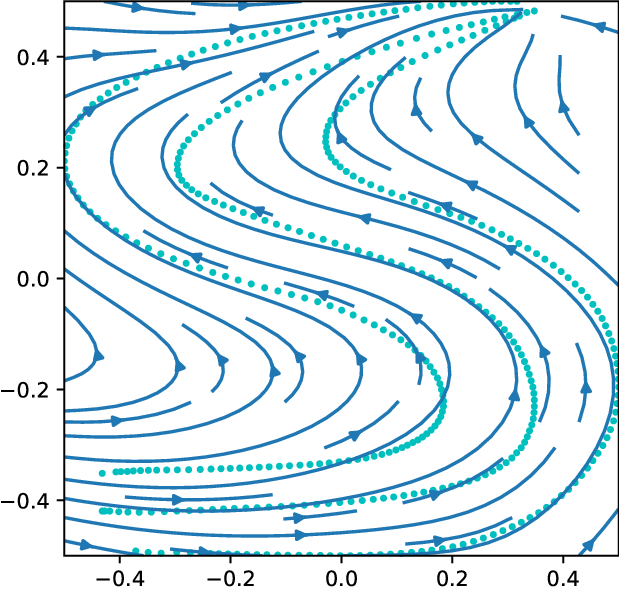}
        \label{subfig:ds2_dynamics_deformed}
    }
    \quad
    \subfloat[]{
        \scalebox{1.2}{\subimport{../figures/}{ds2_potential_original}}
        \label{subfig:ds2_potential_original}
    }
    \quad
    \subfloat[]{
        \scalebox{1.2}{\subimport{../figures/}{ds2_potential_deformed}}
        \label{subfig:ds2_potential_deformed}
    }
    \caption{\footnotesize (a) Linear DS generated via identity diffeomorphism; (b) deformed DS under the action of the learned diffeomorphism; (c) initial quadratic potential function generating the linear DS in Fig.~\ref{subfig:ds2_dynamics_original}; (d) deformed potential under the action of the learned diffeomorphism generating the nonlinear DS in Fig.~\ref{subfig:ds2_dynamics_deformed}.}
    \label{fig:ds2_learn}
\end{figure}

We applied the diffeomorphism to learn a mapping from our original space to the embedding spaces for the hand-drawn multiple-attractor unknown DS, presented in Fig.~\ref{subfig:handdrawn_data}. Fig.~\ref{fig:ds1_learn} focuses on the red sub-dynamics. Fig.~\ref{subfig:ds1_dynamics_original} shows the original linear dynamics produced by Eq.~\ref{eqn:diff_ds} when the diffeomorphism $\psi$ is the identity. After having learned the diffeomorphism, the deformed nonlinear DS is shown in Fig.~\ref{subfig:ds1_dynamics_deformed}. Fig.~\ref{subfig:ds1_potential_original} and \ref{subfig:ds1_potential_deformed} show the potential function $V = \tilde{V} \circ \psi$ with identity and learned diffeomorphism, respectively. Results for the second extracted sub-dynamics are shown in Fig.~\ref{fig:ds2_learn}.

Our approach to learning DS through diffeomorphic mapping is highly inspired by the works of \cite{perrin_fast_2016} and \cite{rana_euclideanizing_2020}. Different from the approach of \cite{perrin_fast_2016}, our learned diffeomorphism generates a map from the original space to the embedding space, rather than the opposite.
This leads to a different formulation of the deformed non-linear DS. The advantage of our approach is that it does not require to construct explicitly the inverse diffeomorphism. This comes at the cost of having to invert the Jacobian when reconstructing the DS. Compared to the approach proposed in \cite{rana_euclideanizing_2020}, we rely solely on the geometrical deformation of the space, induced by the embedding space information, without taking into account dynamics information.
\begin{figure}[ht]
    \centering
    \subfloat[]{\includegraphics[scale=.3]{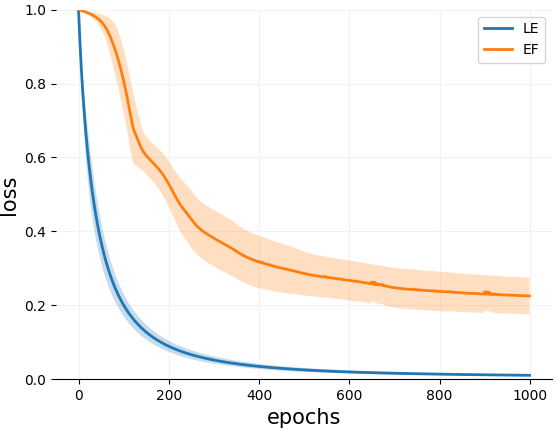}} \quad
    \subfloat[]{\includegraphics[scale=.3]{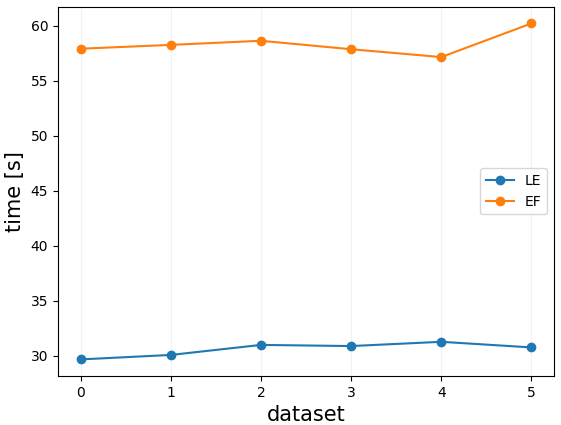}}
    \caption{\footnotesize For Laplacian Embedding (LE) and Euclideanizing Flows (EF) approaches: (a) loss decay over $1000$ epochs; (b) training time for $1000$ epochs.}
    \label{fig:comparison_performance}
\end{figure}
As consequence our loss function considers only position information (original vs. embedding space) discarding the velocities. This allows to achieve faster convergence, due to the simpler learning problem, and sensible lower learning time due to the advantage of not having to calculate and invert the Jacobian in the process. Fig.\ref{fig:comparison_performance} shows a comparison between our approach, termed as Laplacian Embedding (LE), and Euclideanizing Flows (EF) presented in \cite{rana_euclideanizing_2020}. All the tests have been performed on a machine endowed with an Intel i9-10900K CPU and a NVIDIA GeForce RTX 2080Ti GPU\footnote{Repository available at: https://github.com/nash169/learn-diffeomorphism}. On the left it is possible to see how the easier formulation proposed in LE leads to a fast, exponential decaying of the loss while EF generally requires more epochs to properly converge. On the right the training time over $1000$ epochs is shown. As LE does not require to invert the Jacobian, it requires approximately half training time compared to EF.

\begin{figure}[ht]
    \centering
    \subfloat{\includegraphics[width=.14\textwidth]{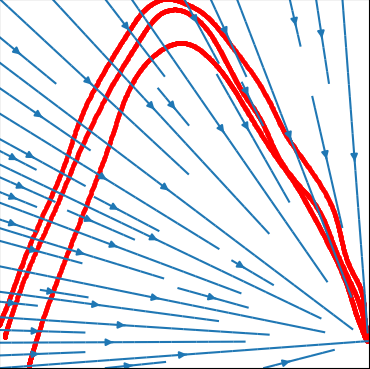}}
    \hspace{0.5mm}
    \subfloat{\includegraphics[width=.14\textwidth]{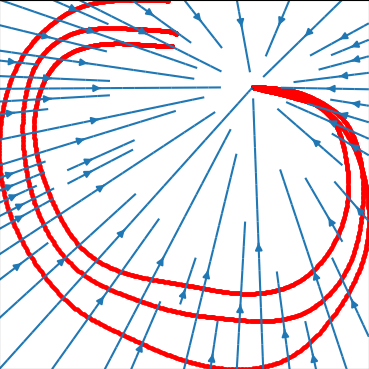}}
    \hspace{0.5mm}
    \subfloat{\includegraphics[width=.14\textwidth]{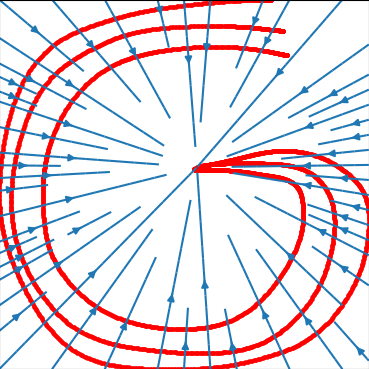}}
    \hspace{0.5mm}
    \subfloat{\includegraphics[width=.14\textwidth]{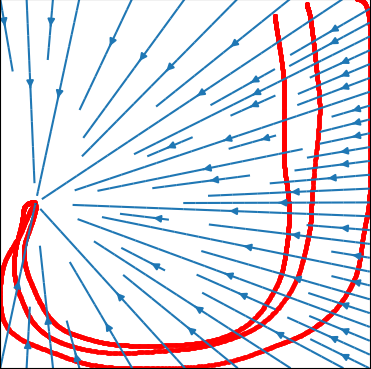}}
    \hspace{0.5mm}
    \subfloat{\includegraphics[width=.14\textwidth]{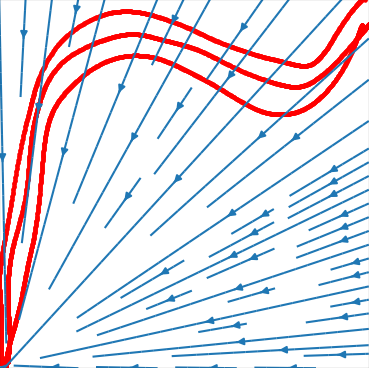}}
    \hspace{0.5mm}
    \subfloat{\includegraphics[width=.14\textwidth]{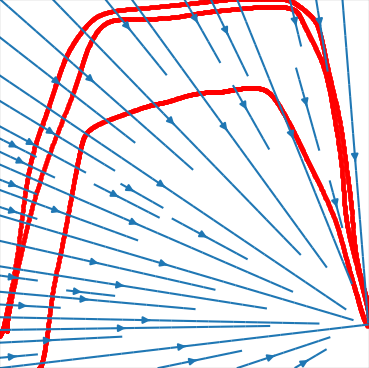}}
    \vspace{0.5mm}
    \subfloat{\includegraphics[width=.14\textwidth]{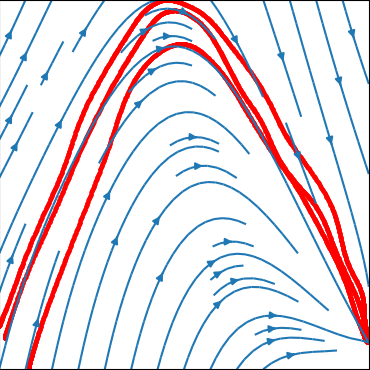}}
    \hspace{0.5mm}
    \subfloat{\includegraphics[width=.14\textwidth]{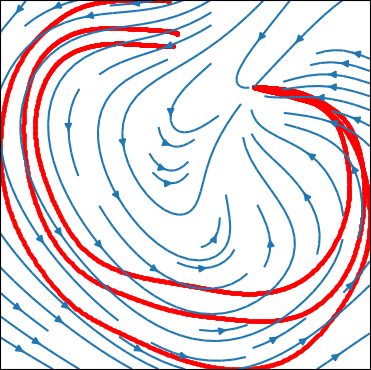}}
    \hspace{0.5mm}
    \subfloat{\includegraphics[width=.14\textwidth]{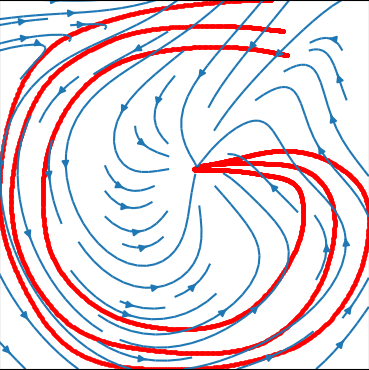}}
    \hspace{0.5mm}
    \subfloat{\includegraphics[width=.14\textwidth]{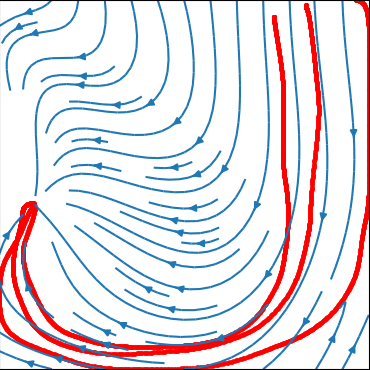}}
    \hspace{0.5mm}
    \subfloat{\includegraphics[width=.14\textwidth]{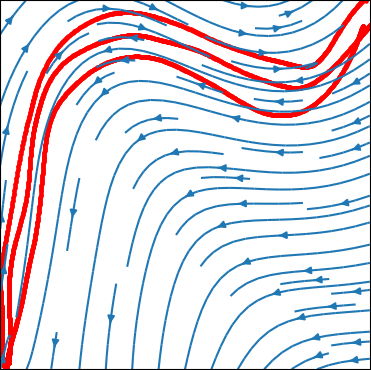}}
    \hspace{0.5mm}
    \subfloat{\includegraphics[width=.14\textwidth]{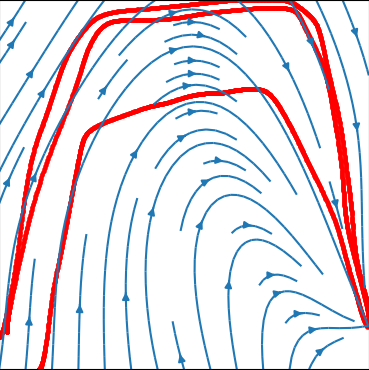}}
    \vspace{0.5mm}
    \renewcommand{\thesubfigure}{a}
    \subfloat[Angle]{\includegraphics[width=.14\textwidth]{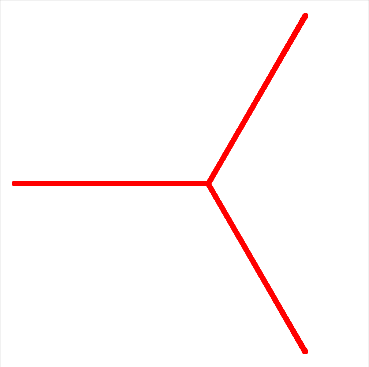}}
    \hspace{0.5mm}
    \renewcommand{\thesubfigure}{b}
    \subfloat[CShape]{\includegraphics[width=.14\textwidth]{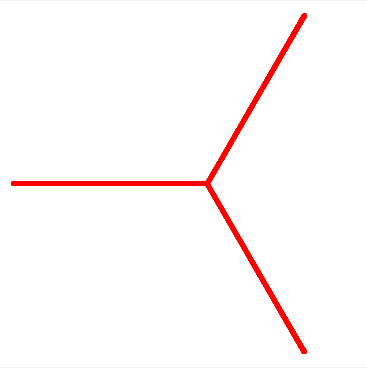}}
    \hspace{0.5mm}
    \renewcommand{\thesubfigure}{c}
    \subfloat[GShape]{\includegraphics[width=.14\textwidth]{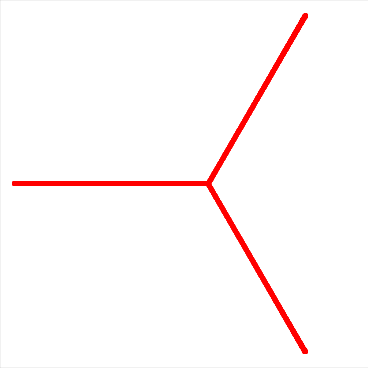}}
    \hspace{0.5mm}
    \renewcommand{\thesubfigure}{d}
    \subfloat[JShape]{\includegraphics[width=.14\textwidth]{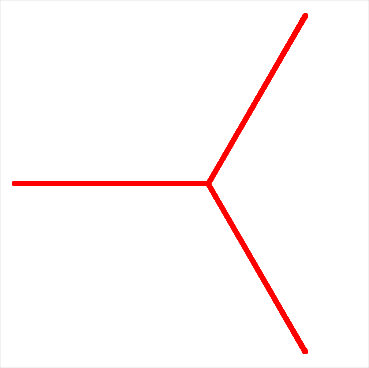}}
    \hspace{0.5mm}
    \renewcommand{\thesubfigure}{e}
    \subfloat[RShape]{\includegraphics[width=.14\textwidth]{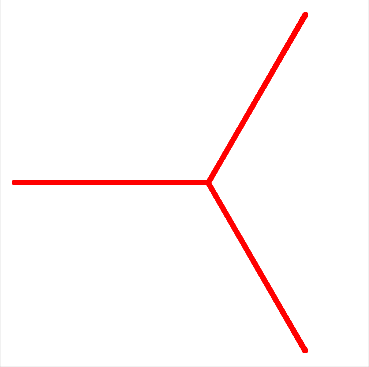}}
    \hspace{0.5mm}
    \renewcommand{\thesubfigure}{f}
    \subfloat[Trapezoid]{\includegraphics[width=.14\textwidth]{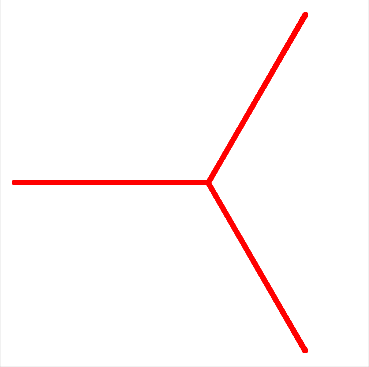}}
    \caption{\footnotesize For each demonstrated DS (column wise), the first row shows the DS generated by the identity diffeomorphism (no train), the second row shows the DS generated after having learned the diffeomorphism between the original space and the embedding space, the third row shows the embedding space reconstructed using the eigenvectors extracted from the eigen-decomposition of the Laplacian matrix.}
    \label{fig:lasaset_ds}
\end{figure}
We evaluate the ability of the proposed non-volume preserving transformation to reconstruct the diffeomorphism between the original and the embedding space on the LASA dataset\footnote{https://bitbucket.org/khansari/lasahandwritingdataset}, a dataset composed of hand-drawn letters, that has been often used to compare performance of learned non-linear DS. Each demonstration is composed by $7$ trajectories ($1000$ samples of position-velocity pairs). Since, in two dimensions, the reconstruction of the embedding does not require more than three trajectories, for each demonstration, we discard 4 trajectories that we use at testing time to evaluate performance.

For the reconstruction of the embedding to be successful, it is necessary that the demonstrated trajectories are instances of a first order DS. While noise is tolerated, the trajectories should not intersect, as this would violate a fundamental property of first order DS and core assumption of the DS graph representation.
\begin{figure}[ht]
    \centering
    \subfloat[Angle]{\includegraphics[width=.15\textwidth]{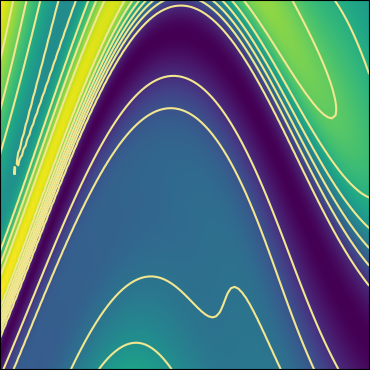}}
    \hspace{0.5mm}
    \subfloat[CShape]{\includegraphics[width=.15\textwidth]{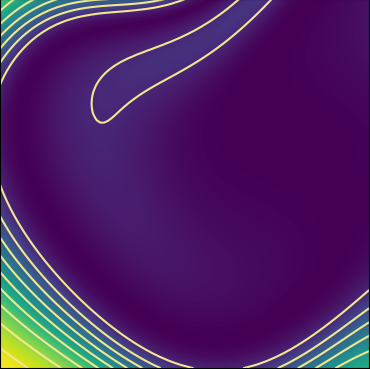}}
    \hspace{0.5mm}
    \subfloat[GShape]{\includegraphics[width=.15\textwidth]{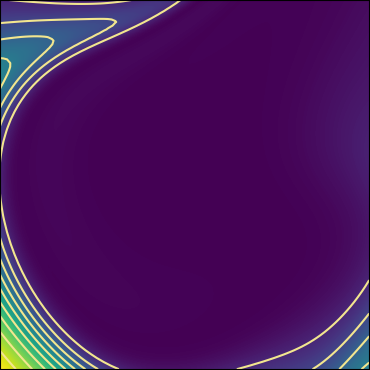}}
    \hspace{0.5mm}
    \subfloat[JShape]{\includegraphics[width=.15\textwidth]{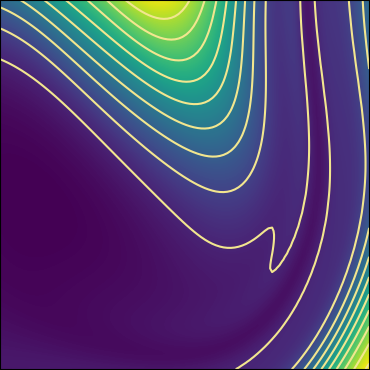}}
    \hspace{0.5mm}
    \subfloat[RShape]{\includegraphics[width=.15\textwidth]{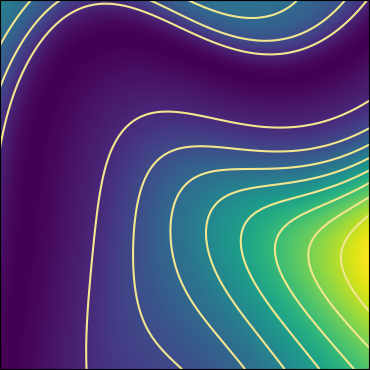}}
    \hspace{0.5mm}
    \subfloat[Trapezoid]{\includegraphics[width=.15\textwidth]{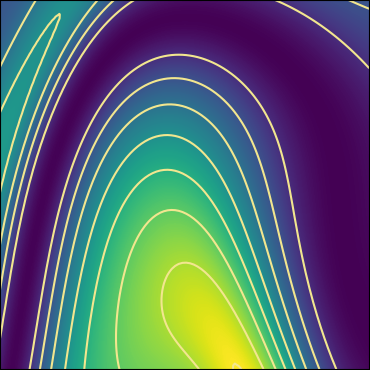}}
    \caption{\footnotesize Deformed potential under the action of the learned diffeomorphism generating the nonlinear DS for the LASA dataset.}
    \label{fig:lasaset_pot}
\end{figure}
Therefore, whenever it was possible, from each demonstration, we selected trajectories in line with this requirement. In those cases where it was not possible, in order to retain a good statistical analysis on the diffeomorphism learning part, we reconstructed the graph structure manually.

Fig.~\ref{fig:lasaset_pot} shows the deformed quadratic potential function under the action of the learned diffeomorphism for a subset of demonstrations, while Fig.~\ref{fig:lasaset_pot} shows the relative learned DS. In order to evaluate the performance of the algorithm we employ three metrics: ($1$) prediction cosine similarity error $\dot{e} = \frac{1}{N_D}\sum_{i=1}^{N_D} \abs{1 - \frac{f(\mathbf{x}_i^{ref})^T\dot{\mathbf{x}}_i^{ref}}{\norm{f(\mathbf{x}_i^{ref})}\norm{\dot{\mathbf{x}}_i^{ref}}}}$, ($2$) \emph{average} Dynamic Time Warping Distance (DTWD) \cite{salvador2007toward} and ($3$) prediction Root Mean Square Error $\text{RMSE} = \frac{1}{N_D}\sum_{i=1}^{N_D} \norm{\dot{\mathbf{x}}_i^{ref} - f(\mathbf{x}_i^{ref})}$.
Table \ref{tab:metrics} shows the performance at reconstructing each of the demonstrations for each of the 12 dynamics shown in Figure \ref{fig:lasaset_ds}.
\begin{table}[ht]
    \centering
    
\begin{tabular}{l|cccccc}
    \toprule
         & Angle & CShape & GShape & JShape & RShape & Trapezoid \\
    RSME & 23.92 & 16.55  & 18.50  & 16.11  & 14.36  & 15.13     \\
    DTWD & 0.015 & 0.066  & 0.098  & 0.060  & 0.046  & 0.073     \\
    CS   & 0.679 & 0.992  & 0.979  & 0.861  & 0.504  & 0.749     \\
    \bottomrule
\end{tabular}

    \caption{\footnotesize Performance evaluation at reconstructing the demonstrations for each of the 12 handdrawn examples of Figure \ref{fig:lasaset_ds}. Performance is measured according to three metrics: root mean square error (RSMR), dynamic time warping distance (DTWD) and cosine similarity error (CS).}
    \label{tab:metrics}
\end{table}

\begin{figure}[ht]
    \centering
    \includegraphics[scale=0.4]{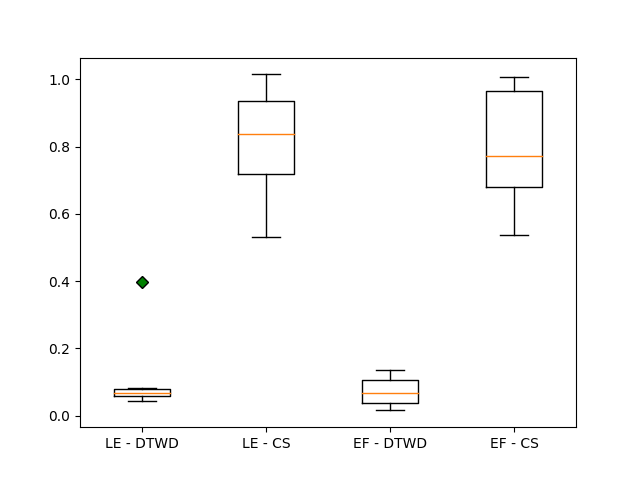}
    \caption{\footnotesize For Laplacian Embedding (LE) and Euclideanizing Flows (EF) approaches box plots comparison of DTWD and CS metrics.}
    \label{fig:comparison_dynamics}
\end{figure}

The proposed implementation of the diffeomorphism appears more capable at shaping the streamlines of the vector field according to the demonstrated trajectories. It also has good accuracy, with values of prediction cosine similarity error and average DTWD about $0.5$ and $0.2$, respectively, all of which are comparable to the current state-of-the-art (e.g. \cite{figueroa_physicallyconsistent_2018, rana_euclideanizing_2020}) as shown in Fig.\ref{fig:comparison_dynamics}.
This comes at the cost of a poor RMSE, over the velocities, given that the velocity information has been discarded in the learning process. Reconstruction of the desired velocity profile can be achieved through proper re-scaling techniques of the reconstructed DS as shown, for instance, by \cite{perrin_fast_2016}.

\section{Discussion \&  Conclusion}
This paper showed that it is possible to automatically decompose a set of unlabelled data, stemming from a multiple-attractor DS, and to identify the number of underlying dynamics and their associated attractors. We further provided theoretical guarantees for DS linearization based on Manifold Learning reconstruction of multiple embedding spaces. We proved that, for a graph structure of the shape described in Sec.~\ref{sec:laplacian}, the eigenvectors of the Laplacian matrix generate an embedding space where the sampled trajectories from a given DS are linearized. We proposed a novel velocity-augmented kernel to achieve the desired graph structure from real data.

Relying on these theoretical results, we proposed an algorithm to cluster the sub-dynamics and identify the equilibria locations of multiple-attractor DS through eigendecompostion of a graph-based Laplacian matrix. In particular, we utilized the spectral properties of a Laplacian matrix applied to the particular graph proposed to reconstruct multiple embedding spaces, where each sub-dynamics is linearized while the others are compressed into a single point at zero. We showcased that, in such space, it is possible to identify the position of the attractor while, at the same time, clustering the sub-dynamics.

In comparison to alternative clustering techniques, our approach clusters correctly all examples we tested, even the complicated, chaotic DS generated by the Duffing equation. Our approach also allowed to identify correctly in each cases the attractors' location within an error inferior to $5\%$. Nevertheless, our algorithm relies heavily on the reconstruction of the correct graph. Whenever the velocity-augmented kernel is not able to reconstruct the correct graph, the clustering performance degrades drastically and it is not possible to locate the attractors anymore. This generally happens in extreme cases of considerable high curvature of the sampled trajectories. In addition we highlight that Spectral Clustering taking advantage of the kernel proposed in this work is capable of matching the clustering performance of our algorithm. However Spectral Clustering does not exploit the particular structure of the DS in the embedding space, and therefore, it is incapable of assessing the location of the attractors.

A usual approach to determining the attractors is to search for zero-velocity crossing. Velocity-crossing usually performs poorly around sharp curvatures since, at the junction, the velocity may decrease importantly. By using the velocity-augmented kernel, our approach is robust to sharp curvatures and can embed highly curvy dynamics, while determining the true attractors.

In all examples used in this paper, the DS were 2-dimensional and represented single motion patterns, However, our algorithm is not limited theoretically to 2D and scales well to higher dimensions since both the proposed kernel and the Laplacian matrix are dimension-independent. The dimension of the embedding space is upper bounded by the number of trajectories sampled and it does not depend on the dimension of the original space.

Combined with state-of-the-art techniques for learning diffeomorphic maps, our method provides an algorithm for stable learning of multiple-attractor DS in a complete unsupervised learning scenario.

\newpage
\acks{We would like to acknowledge Aradhana Nayak for her contribution to Prop.~\ref{prop:multiplicity_eig} and \ref{prop:eig_bound}.}

\newpage
\appendix

\section{Proof of Lem.~\ref{lem:recursion} from Sec~\ref{sec:laplacian}}
\label{app:proof_lem2}
For each m-th row $L_m$ of $L(G)$, we have $D+1$ non-zero entries, where $D$ is the degree of the $m$-th node with $L_{m,n}=D$ for $m=n$ and $L_{m,n}=-1$ for $m \neq n$. In the following, for the sake of clarity, we will drop the upper index $k$ for the eigenvector's entries.

The first node, $n=1$  has degree $D=1$, as it is connected only to its direct neighbor. Using the monotonic ordered labelling of the nodes, $L_1 \mathbf{u} = \lambda u_1$ yields $u_1 - u_2 = \lambda u_1$, that simplifies into:
\begin{equation}
    u_2= (1 - \lambda) u_1.
    \label{eqn:u1_to}
\end{equation}

The second node of the path graph is connected to the previous and next node. It has hence degree $D=2$. $L_2 \mathbf{u} = \lambda u_2$ yields $2 u_2 - u_1 - u_3 = \lambda u_2$. The same holds for all the other nodes in the path. Hence, we have $2 u_n - u_{n-1} -u_{n+1} = \lambda u_n$. This recurrence can be rewritten into the recursion:
\begin{equation}
    u_{n+1} =  (2 - \lambda) u_{n} - u_{n-1} \quad \text{for } n=2, \dots, p_k-1,
    \label{eqn:ui_to}
\end{equation}
or, equivalently,
\begin{equation}
    u_n = (2 - \lambda) u_{n-1} - u_{n-2} \quad \text{for } n = 3, \dots, p_k.
\end{equation}

\section{Proof of Lem.~\ref{lem:chebyshev_poly} from Sec~\ref{sec:laplacian}}
\label{app:proof_lem3}
Consider the following Chebyshev polynomial of first kind $T$:
\begin{align}
    T_1(\lambda) & = 1 \notag                                                                                            \\
    T_2(\lambda) & = 1 - \frac{\lambda}{2} \notag                                                                        \\
    T_n(\lambda) & = 2 \left( 1-\frac{\lambda}{2} \right) T_{n-1}(\lambda) - T_{n-2}(\lambda) \quad \text{for } n \ge 3,
\end{align}
equivalent to the following trigonometric expression:
\begin{equation}
    T_n(\lambda) = \cos \left((n - 1) \cos^{-1} \left( 1-\frac{\lambda}{2} \right) \right) \quad \text{for } n \ge 1.
    \label{eqn:first_chebyshev}
\end{equation}
Consider the following Chebyshev polynomial of the second kind $V$:
\begin{align}
    V_1(\lambda) & = 1 \notag                                                                                            \\
    V_2(\lambda) & = 2 \left( 1-\frac{\lambda}{2} \right) \notag                                                         \\
    V_n(\lambda) & = 2 \left( 1-\frac{\lambda}{2} \right) V_{n-1}(\lambda) - V_{n-2}(\lambda) \quad \text{for } n \ge 3,
\end{align}
equivalent to the following trigonometric expression:
\begin{equation}
    V_n(\lambda) = \frac{ \sin \left( n \cos^{-1}\left( 1-\frac{\lambda}{2} \right) \right) }{ \sin \left( \cos^{-1}\left( 1-\frac{\lambda}{2} \right) \right) } \quad \text{for }n \ge 1.
    \label{eqn:second_chebyshev}
\end{equation}
The combination of the Chebyshev polynomials in Eq.~\ref{eqn:first_chebyshev} and \ref{eqn:second_chebyshev}, as indicated in Eq.~\ref{eqn:chebyshev_poly1}, yields the trigonometric expression in Eq.~\ref{eqn:chebyshev_poly2}.

\section{Proof of Prop.~\ref{prop:monotonicity} from Sec~\ref{sec:laplacian}}
\label{app:proof_prop4}
For simplicity, we remove the superscript $k$ in $u^k_n$ and denote it as $u_n$. To preserve monotonicity, we must determine the conditions for which the entries do not change sign along one path. From Lem.~\ref{lem:chebyshev_poly}, we know that $u_n$ follows periodic functions that are expressed as a combination of trigonometric functions given by Eq.~\ref{eqn:chebyshev_poly2} for $n\geq 1$. We study the stationary points of Eq.~\ref{eqn:chebyshev_poly2}  with respect to the index $n$:
\begin{align}
    \frac{\partial}{\partial n} u_n & = u_1 \left[ - \theta \sin((n-1) \theta)  - \theta \frac{\lambda}{2} \frac{\cos((n-1) \theta)}{\sin(\theta)}\right] \notag \\
                                    & = \sin((n-1) \theta)  + \frac{\lambda}{2} \frac{\cos((n-1) \theta)}{\sin(\theta)} = 0.
    \label{eqn:chebyshev_derivative}
\end{align}
Let the eigenvalue $\lambda$ be
\begin{equation}
    \lambda = 2 \left( 1 - \cos(\theta - 2 \pi j) \right), \quad j \in \mathbb{N}.
    \label{eqn:eig_def}
\end{equation}
This expression of $\lambda$ is sufficient to represent all eigenvalues in $[0,4]$. Replacing Eq.~\ref{eqn:eig_def} in Eq.~\ref{eqn:chebyshev_derivative}, we obtain
\begin{align}
    \sin((n-1) \theta) + \frac{2 (1 - \cos(\theta - 2 \pi j))}{2} \frac{\cos((n-1) \theta)}{\sin(\theta - 2 \pi j)} & = 0 \notag \\
    \sin((n-1) \theta) + \tan(\frac{\theta}{2} - \pi j) \cos((n-1) \theta)                                          & = 0 \notag \\
    \tan((n-1) \theta) + \tan(\frac{\theta}{2} - \pi j)                                                             & = 0.
\end{align}
The stationary points correspond to all $n$, such that:
\begin{equation}
    n = \frac{\pi j}{\theta} + \frac{1}{2}.
\end{equation}
Expressing the stationary points in term of $\lambda$, we have:
\begin{equation}
    n = \frac{\pi j}{\cos^{-1}(1-\frac{\lambda}{2})} + \frac{1}{2}.
    \label{eqn:upper_monotonicity}
\end{equation}
The monotonicity of the entries $u_n$ is hence preserved until one reaches a stationary point.

Observe, from Eq.~\ref{eqn:upper_monotonicity}, that the smaller $\lambda$, the larger the number of nodes in the path with monotonicity preserved. If $p_k$ is the number of nodes within each path graph $k$, we can determine an upper bound on $\lambda$ for which all $u_n$ within one path would evolve monotonically :
\begin{equation}\label{lambda_bds}
    \lambda \le 2\left[ 1 - \text{cos}\left( \frac{\pi}{p_k-\frac{1}{2}} \right) \right].
\end{equation}
From \eqref{lambda_bds}, it is clear that for $p_k\geq 3$ one has additionally $\lambda<1$. From Prop.~\ref{lem:recursion}, we have $u_2= (1-\lambda) u_2$, hence $u_2 < u_1$ if $u_1$ positive and $u_2 > u_1$, otherwise.

\section{Proof of Prop.~\ref{prop:multiplicity_eig} from Sec~\ref{sec:laplacian}}
\label{app:proof_prop5}
In the following we will drop the reference to the underlying graph structure $G$. Given the circulant block structure of the matrix $J$ in Eq.~\ref{eqn:matrix_j}, as shown by \cite{tee2007eigenvectors}, the eigenvalues and corresponding eigenvectors of such matrix are determined by the following $K$ equations, each giving us $N$ eigenvalues and eigenvectors
\begin{equation}
    H_j v = \lambda_j v \quad \quad j \in \lbrace 0, \dots, K-1 \rbrace,
\end{equation}
with the matrix $H_j$ defined as
\begin{equation}
    H_j \coloneqq B_0 + B_1 \left( \rho_j + \rho^{K-1}_j \right),
    \label{eqn:matrix_h}
\end{equation}
where $\rho_j = \text{exp}\left( i \frac{2 \pi j}{K} \right)$.
Observe that $\rho^{K-1}_j = \rho^{-1}_j$ for all $j$. Further,
\begin{equation}
    \rho_j + \rho_j^{-1} = 2 \text{cos} \left( \frac{2 \pi j}{K}  \right)= 2 \text{cos} \left( \frac{2 \pi (K-j)}{K} \right)= \rho_{K-j} + \rho_{K-j}^{-1}.
\end{equation}
Therefore the matrices $H_j$ and $H_{K-j}$, for $j \ge 1$, yield the same eigenvalues. This shows how in the matrix $J$ has at least $\frac{(K-1)}{2}$, if $K$ is odd, or $\frac{K}{2} - 1$, if $K$ is even, eigenvalues with algebraic multiplicity equal to $2$.

\section{Proof of Prop~\ref{prop:eig_bound} from Sec~\ref{sec:laplacian}}
\label{app:proof_prop6}
In the following we will drop the reference to the underlying graph structure $G$. The eigenvalues of the Laplacian matrix $L$ are given by
\begin{equation}
    \Lambda_L = 2 I - \Lambda_J,
\end{equation}
where $\Lambda_L$ and $\Lambda_J$ are the diagonal matrices containing the eigenvalues of $L$ and $J$, respectively. Therefore the smallest non-zero eigenvalue of $L$ corresponds to the largest non-zero eigenvalue of $J$. The matrices $H_j$ in Eq.~\ref{eqn:matrix_h} can be expressed as
\begin{equation}
    H_j = I + M_j,
\end{equation}
where $M_j$ is the matrix
\begin{equation}
    M_j = \begin{bmatrix}
        0 & 1      &        &        &              \\
        1 & -1     & \ddots &        &              \\
          & \ddots & \ddots & \ddots &              \\
          &        & \ddots & -1     & 1            \\
          &        &        & 1      & \alpha_j - 2
    \end{bmatrix} = M_{1,j} + M_2,
\end{equation}
wherein
\begin{equation}
    M_{1,j} = \begin{bmatrix}
        M_0 & v            \\
        v^T & \alpha_j - 2
    \end{bmatrix},
    \quad
    M_2 = \begin{bmatrix}
        0 &    &        &    &   \\
          & -1 &        &    &   \\
          &    & \ddots &    &   \\
          &    &        & -1 &   \\
          &    &        &    & 0
    \end{bmatrix}
\end{equation}
with $\alpha_j = 2 \text{cos} \left( \frac{2 \pi j}{K} \right)$ and
\begin{equation}
    M_0 = \begin{bmatrix}
        0 & 1      &        &   \\
        1 & \ddots & \ddots &   \\
          & \ddots & \ddots & 1 \\
          &        & 1      & 0
    \end{bmatrix},
    \quad
    v = \left[0, \dots, 0, 1\right]^T.
\end{equation}
$M_0$ is the adjacency matrix of a single path graph with $N-1$ vertices.
From Thm. 3.7 in \cite{bapat2010graphs} the eigenvalues of $M_0$ are given by
\begin{equation}
    \lambda_n(P) = 2 \text{cos}\left( \frac{\pi n}{N} \right), \quad n = 1, \dots, N-1.
\end{equation}
Let $\lambda_{max}$ and $\lambda_{min}$ the largest and the smallest eigenvalues in the spectrum, respectively. From Thm 4.3.17 in \cite{horn2012matrix}, the largest eigenvalue of each $M_1(i)$ matrices is equal or larger than the largest eigenvalue of $M_0$. The largest eigenvalue of $M_0$ is found for $n=1$. Hence, we have
\begin{equation}
    2 \cos\left( \frac{\pi}{N}\right) \leq \lambda_{n}(M_{1,j}), \quad \forall j.
\end{equation}
From Cor. 4.3.15 in \cite{horn2012matrix} the largest eigenvalue among all $M_j$ matrices is bounded above and below by
\begin{equation}
    2 \text{cos} \left( \frac{\pi}{N} \right) - 1 = \lambda_{max}(M_{1,j}) + \lambda_{min}(M_2) < \lambda_{max}(M_j) < 2 \cos \left( \frac{\pi}{N} \right) \quad \forall j.
\end{equation}
The inequality is strict as $M_{1,j}$ and $M_2$ do not have a common eigenvector. Recalling that $\lambda(L) = 1 - \lambda(M_j)$, this is equivalent to
\begin{equation}
    1-2\cos\left( \frac{\pi}{N}\right) < \lambda_{min}(L) < 2 \left(1 - \text{cos} \left( \frac{\pi}{N} \right) \right),
\end{equation}
being $\lambda_{min}(L)$ the smallest non-zero eigenvalue of the Laplacian matrix. Since $\cos \left( \frac{\pi}{N} \right) > \cos \left( \frac{\pi}{N - \frac{1}{2}} \right)$ for $N \ge 2$, we obtain the inequality in Eq.~\ref{eqn:lambda_bound}.

\vskip 0.2in
\bibliography{references,references_add}

\end{document}